\documentclass[pdflatex,sn-mathphys-num]{sn-jnl}

\usepackage{graphicx}%
\usepackage{multirow}%
\usepackage{amsmath,amssymb,amsfonts}%
\usepackage{amsthm}%
\usepackage{mathrsfs}%
\usepackage[title]{appendix}%
\usepackage{xcolor}%
\usepackage{textcomp}%
\usepackage{manyfoot}%
\usepackage{booktabs}%
\usepackage{algorithm}%
\usepackage{algorithmicx}%
\usepackage{algpseudocode}%
\usepackage{listings}%
\usepackage{array}

\theoremstyle{thmstyleone}%
\newtheorem{theorem}{Theorem}
\theoremstyle{thmstyletwo}%
\newtheorem{example}{Example}%

\theoremstyle{thmstylethree}%

\begin{document}
\title[On the use  of the Wasserstein metric with various discrete probability measures with application to 2D curves]{On the application  of the Wasserstein metric  to 2D curves classification}


\author[1]{\fnm{Agnieszka } \sur{Kaliszewska}}\email{agnieszka.kaliszewska@ibspan.waw.pl}

\author*[2]{\fnm{ Monika} \sur{Syga}}\email{monika.syga@pw.edu.pl}
\equalcont{These authors contributed equally to this work.}

\affil*[1]{\orgdiv{Department of Optimization and Modelling of Dynamical Systems }, \orgname{Systems Research Institute, PAS}, \orgaddress{\street{Newelska 6}, \city{Warsaw}, \postcode{01-447},  \country{Poland}}}

\affil[2]{\orgdiv{Faculty of Mathematics and Information Science}, \orgname{Warsaw University of Technology}, \orgaddress{\street{ Koszykowa 75}, \city{Warsaw}, \postcode{00-662}, \country{Poland}}}



\abstract{In this work we analyse  a number of variants of the Wasserstein distance which allow to focus the classification on the prescribed parts (fragments) of classified 2D curves. These variants are based on the use of a number of discrete probability measures which reflect the importance of given fragments of curves.
 The  performance of this approach is tested through a series of experiments  related  to the clustering analysis of 2D curves  performed on data coming from the field of archaeology.}

\keywords{Wasserstein metric, similarity, clustering 2D curves, signal clustering, penalties, weights, arc length}

\maketitle

\section{Introduction}
\label{sec:intro}


The Wasserstein distance can be used for broadly understood similarity detection, shape fitting, and image color matching (cf. \cite{Bonneel}, \cite{WD_in_Geo}). It is sometimes used in combination with the Procrustes distance (e.g. \cite{PetitJean2004}).

From the computational point of view the problem of determining the Wasserstein distance reduces to solving the linear programming problem.
To increase the efficiency a number of heuristic approaches has been proposed e.g.  \cite{Bonneel}.

In the present work, we exploit the flexibility of the definition of the  Wasserstein distance in order to include the possibility to investigate the influence of different probability distribution on the classification results. In the cases of some distributions this allows to provide a closed form expression of solutions for the Wasserstein distance. 

In the Wasserstein distance the weights $\alpha_j$, $\beta_i$ $j=1,...,m$, $i=1,...,n$, see Equations \eqref{beta},\eqref{alpha} below, represent discrete probability measures which can be interpreted as the indicators of importance (interest) of given components $i$ and $j$ respectively in the classification process. The most common choice are  $\alpha_j=\frac{1}{m}$ and $\beta_i=\frac{1}{n}$ refers to the uniform distribution of importance.


We apply  the Wasserstein distance with various discrete distributions in the clustering analysis and similarity detection of 2D curves. For the use of different measures in similar problems see e.g. \cite{C&C}, \cite{AMCS_A&M}. 
We will discuss the properties of the Wasserstein distance that make it well suited to detect similarity in our application as described in Section \ref{Section_Partial_Optimal_Transport}.  
The Wasserstein distance is designed to compare the distribution  of probability within given sets. It is well suited to handle sets that differ in dimensions (the number of observations).

The Wasserstein distance is largely widespread in  applications \cite{Solomon}. It's popularity stems from the flexibility it can be formulated, and in consequence  the numerous problems it can be applied to. One particular property that contributes to the widespread of the Wasserstein/earth movers distance and the Optimal transport problem is its ability to match partial data to a bigger set such as in the case of fitting two point clouds \cite{Bonneel}.

The feature of the Wasserstein distance which is particularly important in this paper is the possibility to distinguish particular parts of the objects on which the classification is concentrated. This is discussed in Section \ref{Section_Wass_for_different_distributions}. This is the original contribution of the present work, which is not too commonly discussed in the literature.  The problem of concentration of the classification on a prescribed part of object is referred to as {\em Constrained Mass Optimal Transport} (\cite{constrained_transport}).

The organization of the paper is as follows. In Section \ref{Wass_dist} we discuss the Wasserstein distance in the context of similarity measures in 2D objects. Section \ref{Section_Wass_for_different_distributions} contains the main contribution of the present paper, various distributions which allow to concentrate the classification process on a selected section of the objects. Section \ref{Section_Partial_Optimal_Transport} presents the formulation of the Partial Optimal Transport.  In Section \ref{Section_Experiments} we discuss the results of extensive experiments performed using the distributions proposed in Section \ref{Section_Wass_for_different_distributions}. Section \ref{Section_conclusions} concludes the paper.

\section{The Wasserstein distance}
	\label{Wass_dist}
	
The Wasserstein distance is in its essence a formulation of the broader optimal transport problem (OTP). The optimal transport problem can be successfully used in many applications, especially in computer graphics and image recognition. It is well suited to compare histograms or to produce interpolations between probability distributions. In literature OTP is considered to be a computationally difficult problem. A recently commonly applied approach is the formulation of OTP as a linear assignment problem.
Due to the possibility to assign weights to points we are able to highlight the areas of significance in the investigated curves. 

Let $x=(x_i^1,x_i^2)_{i=1}^{n}$ and $y=(y^1_{j},y^2_{j})_{j=1}^{m}$.  The Wasserstein distance, \cite{MAL-073}, \cite{Solomon}, \cite{PetitJean2004} as a special case of OTP can be formulated as follows: 
	\begin{equation}
		\label{WD}
		W(x,y)= \min_{\pi_{ij}\geq 0} \{\sum_{i=1}^{n}\sum_{j=1}^{m}d((x_i^{1},x_{i}^{2}),(y_j^{1},y_{j}^{2}))\pi_{ij}\},                        
	\end{equation}
subject to:	
	\begin{equation}
	\label{beta}
		\sum_{j=1}^{m}\pi_{ij}=\beta_{i}, i=1...n,
	\end{equation}

		\begin{equation}
		\label{alpha}
		\sum_{i=1}^{n}\pi_{ij}=\alpha_{j}, j=1...m,
	\end{equation}

where: 
    \begin{equation}
        \begin{array}{cc}
        x=\left[\begin{array}{c}(x_{1}^{1},x_{1}^{2})\\
        \vdots\\
        (x_{n}^{1},x_{n}^{2})\end{array}\right]&
         y=\left[\begin{array}{c}(y_{1}^{1},y_{1}^{2})\\ 
        \vdots\\
        (y_{m}^{1},y_{m}^{2})\end{array}\right]
        \end{array}
    \end{equation}
    
    \begin{equation}
    \label{mu}
	\mu=\sum_{i=1}^{n}\beta_{i}\delta _{(x{_i}^{1},x{_i}^{2})}, \\ \sum_{i=1}^{n}\beta_{i} =1,
	\end{equation}
	\begin{equation}
	\label{nu}
	\nu=\sum_{j=1}^{m}\alpha_{j}\delta_{(y{_j}^{1},y{_j}^{2})}, \\\sum_{j=1}^{m}\alpha_{j} =1, 
	\end{equation}
	where for any $(z{_k}^{1},z{_k}^{2})\in\mathbb{R}^{2}$ and any $(z^{1},z^{2})\in\mathbb{R}^{2}$
	\begin{equation}
\delta_{(z{_k}^{1},z{_k}^{2})}(z^{1},z^{2})=\left\{\begin{array}{ll}
1&(z{_k}^{1},z{_k}^{2})=(z^{1},z^{2})\\
0&(z{_k}^{1},z{_k}^{2})\neq(z^{1},z^{2})
\end{array}
\right.
	\end{equation}

In the general form the discrete optimal transport problem (OT) can be represented as
\begin{equation}
    \label{eq_ot}
    OT(x,y;C)=\left\{ \begin{array}{rl}
         \min_{\pi \in\mathbb{R}^{n\times m}}&\sum_{ij}\pi_{ij}c_{ij} \\
         s.t.&\pi\ge 0\\
         &\sum_{j}\pi_{ij}=\beta_{i}
         \ \forall i=1,...,n\\
         &\sum_{i}\pi_{ij}=\alpha_{j}\ \forall j=1,...,m\\
    \end{array}\right.,
\end{equation}
where  $C$ is an $n\times m$ matrix.

This is a linear programming problem which is computationally efficiently tractable and it is the most natural way to make the optimal transport operational in the discrete context. 

When $n=m=k$ and $C$ is symmetric, non-negative and satisfies the triangle inequality, it can be checked that
$OT(\cdot,\cdot; C)$ is a distance on 
$$
\Sigma_{k}:=\{u\in\mathbb{R}^{k}_{+}\ |\ \sum_{i=1}^{k}u_{i}=1\}.
$$
Namely, in \cite{Cuturi} the following theorem is proved.

\begin{theorem}{[c.f.Theorem 1,\cite{Cuturi}]}
\\
The optimal transport value $OT(x,y;C)$ given by \eqref{eq_ot} is a distance on $\Sigma_{k}$, {\em i.e.} for $x,y\in \Sigma_{k}$.
\end{theorem}


The problem \eqref{eq_ot} is equivalent to the following linear programming problem, written in the matrix form,
\begin{equation}
    \label{eq_ot2}
    OT(x,y;C)=OT(\alpha,\beta;C)=\begin{array}[t]{rl}
    \text{minimize}
        &c^{T}\pi \\
         s.t.&\pi\ge 0\\
         &A\pi=\left[\begin{array}{l}\beta\\ \alpha\end{array}\right]
    \end{array}
\end{equation}



In the literature, there exists another way to solve this problem for increasing dimension of the problem c.f. \cite{Bonneel},  (https://perso.liris.cnrs.fr/nicolas.bonneel/spot/). 

In \cite{Rubner} the following more general formulation, with \eqref{mu} and \eqref{nu} being relaxed, has been considered.
Let $x=(x_i^1,x_i^2)_{i=1}^{n}$ and $y=(y^1_{j},y^2_{j})_{j=1}^{m}$.  

	\begin{equation}
		\label{WDR}
		WR(x,y)= \min_{\pi_{ij}\geq 0} \{\sum_{i=1}^{n}\sum_{j=1}^{m}d((x_i^{1},x_{i}^{2}),(y_j^{1},y_{j}^{2}))\pi_{ij}\},                        
	\end{equation}
subject to:	
	\begin{equation}
		\pi_{ij}\geq0, i=1...n, j=1...m,
	\end{equation}
	\begin{equation}
		\sum_{j=1}^{m}\pi_{ij}\leq \beta_{i}, i=1...n,
	\end{equation}
		\begin{equation}
		\sum_{i=1}^{n}\pi_{ij}\leq \alpha_{j}, j=1...m,
	\end{equation}
	\begin{equation}
	\label{14}
	\sum_{i=1}^n\sum_{j=1}^m\pi_{ij}=\min{\left(\sum_{i=1}^n\beta_{i}, \sum_{j=m}^m\alpha_{j}\right) },
	\end{equation}
	where $\beta_i\geq0,\ i=1,...,n$, $\alpha_j\geq0,\ j=1,...,m$, $\displaystyle\sum_{i=1}^n\beta_{i}>0$ and $\displaystyle\sum_{j=1}^n\alpha_{j}>0$.
	
	If in the above formulation  we add \eqref{mu} and \eqref{nu}, {e.m.,} we assume that $\displaystyle\sum_{i=1}^n\beta_{i}=1$ and  $\displaystyle\sum_{j=1}^n\alpha_{j}=1$ then the problem \eqref{WDR} is equivalent to the problem \eqref{WD}. According to the classical interpretation, the condition \eqref{14} guarantees that the whole mass is transported.
	
	Relationships between \eqref{WD} and its generalizations (including \eqref{WDR})  have been summarized in \cite{Bonneel}, Table 1.
    

\section{Wasserstein distance for different distributions }
\label{Section_Wass_for_different_distributions}
In this section we present various approaches to the application of the Wasserstein distance in the classification of curves.

Taking advantage of the formulation of the Wasserstein distance, we substitute the commonly used values of $\alpha_j=\frac{1}{m}$ and $\beta_i=\frac{1}{n}$ (\ref{waga_uniform}), with other distributions, allowing us to manipulate parameters of distributions in order to focus the classification on a prescribed regions of the curves.

\subsection{Standard distributions}
We consider the following standard distributions:

\begin{description}
\item{(i)} uniform discrete distribution

\begin{equation}
\label{waga_uniform}
\beta_{i} =\frac{1}{n} \ \ \ i=1,...,n  \ \text{and} \ \ \alpha_{j}=\frac{1}{m}, \  j=1,...,m.
\end{equation}
\item{(ii)} binomial distribution
$$
\beta_{i}= {n \choose i}p^i(1-p)^{n-i} \ \ \textnormal{and} \ \ \alpha_{j} ={m \choose j}p^j(1-p)^{m-j},
$$
where parameter $p\in(0,1)$ is fixed. 

\end{description}

\subsection{Other relevant distributions}
\label{Section_other_distributions}

Moreover, we consider the following  distributions, which we introduce in relation to our application, as described below. 
\begin{description}
\item{(i)} Increasing importance with respect to indices
\begin{equation}
		\label{waga_po_ycale}
		\beta^{0}_{i}= 
	\frac{i}{\sum\limits_{i=1}\limits^{n}i},\ i=1,...,n,\ \ \ 
		\alpha^{0}_{j}= 
		\frac{j}{\sum\limits_{j=1}\limits^{m}j},\ \ j=1,...,m \ \ \
		\end{equation}

\item{(ii)} decreasing importance  with respect to indices, using the above $\beta_{i}^{0}$, $\alpha_{j}^{0}$ from \eqref{waga_po_ycale} 
		\begin{equation}
		\label{waga_wspolrzedne_odwrocona}
		\bar{\beta}^{0}_{i}= 
	\frac{1-\beta_{i}^{0}}{n-1},\ i=1,...,n,\ \ \ \bar{\alpha}^{0}_{j}= 
		\frac{1-\alpha_{j}^{0}}{m-1},\ \ j=1,...,m \ \ \
		\end{equation}

\item{(iii)} importance of first components
\begin{equation}
		\label{normal_distribution}
		\beta^{0}_{i}= 
	\frac{x_{i}^{1}}{\sum\limits_{i=1}\limits^{n}x_{i}^{1}},\ i=1,...,n,\ \ \ 
	\alpha^{0}_{j}= 
		\frac{y_{j}^{1}}{\sum\limits_{j=1}\limits^{m}y_{j}^{1}},\ \ j=1,...,m \ \ \
	\end{equation}	

\item{(iv)} importance of second components
\begin{equation}
		\label{normal_distribution}
		\beta^{0}_{i}= 
	\frac{x_{i}^{2}}{\sum\limits_{i=1}\limits^{n}x_{i}^{2}},\ i=1,...,n,\ \ \ 
	\alpha^{0}_{j}= 
		\frac{y_{j}^{2}}{\sum\limits_{j=1}\limits^{m}y_{j}^{2}},\ \ j=1,...,m \ \ \
	\end{equation}

\item {(v)} reversed importance of second components, using the above $\beta_{i}^{0}$, $\alpha_{j}^{0}$ from \eqref{normal_distribution} 
		\begin{equation}
		\label{waga_po_ycale_odwrocona}
		\bar{\beta}^{0}_{i}= 
	\frac{x_{max}^2-x_{i}^{2}}{\sum_{i=1}^{n}(x_{max}^2-x_{i}^{2})},\ i=1,...,n,\ \ \ \bar{\alpha}^{0}_{j}= 
		\frac{y_{max}^2-y_{i}^{2}}{\sum_{i=1}^{n}(y_{max}^2-y_{i}^{2})},\ \ j=1,...,m, \ \ \
		\end{equation}
		where $x^2_{max}=max\{x_1^2,..., x_n^2\}$ and $y^2_{max}=max\{y_1^2,..., y_n^2\}$.
\end{description}
	
	
Below we present a set of distributions with a preselected support set  $i=k_1,...,k_2$, and $j=l_1,...,l_2$. We consider the following distributions:
\begin{description}

    \item{(i)} uniform distribution of on a preselected support set 
		\begin{equation}
		\label{zerow}
		\beta^{0}_{i}= \left\{ \begin{array}{ll}
		\frac{1}{k_2-k_1}& \ \ \ for \ i=k_1,...,k_2\\
         0& \ \ otherwise
    \end{array}\right.,\ 
		\alpha^{0}_{j}= \left\{ \begin{array}{ll}
		\frac{1}{l_2-l_1}& \ \ \ for \ j=l_1,...,l_2\\
         0& \ \ otherwise
    \end{array}\right.,
		\end{equation}
		
		\item{(ii)} importance of first components on a preselected support set 
		\begin{equation}
		\label{zerowa}
		\beta^{0}_{i}= \left\{ \begin{array}{ll}
	\frac{x_{i}^{1}}{\sum\limits_{i=k_1}\limits^{k_2}x_{i}^{1}}& \ \ \ for \ i=k_1,...,k_2\\
         0& \ \ otherwise
    \end{array}\right.,\
		\alpha^{0}_{j}= \left\{ \begin{array}{ll}
	\frac{y_{j}^{1}}{\sum\limits_{j=l_1}\limits^{l_2}y_{j}^{1}}& \ \ \ for \ j=l_1,...,l_2\\
         0& \ \ otherwise
    \end{array}\right.,
		\end{equation}
	
	\item{(iii)} importance of second components on a preselected support set 
		
		\begin{equation}
		\label{waga_po_y}
		\beta^{0}_{i}= \left\{ \begin{array}{ll}
	\frac{x_{i}^{2}}{\sum\limits_{i=k_1}\limits^{k_2}x_{i}^{2}}& \ \ \ for \ i=k_1,...,k_2\\
         0& \ \ otherwise
    \end{array}\right.,\
		\alpha^{0}_{j}= \left\{ \begin{array}{ll}
		\frac{y_{j}^{2}}{\sum\limits_{j=l_1}\limits^{l_2}y_{j}^{2}}& \ \ \ for \ j=l_1,...,l_2\\
         0& \ \ otherwise
    \end{array}\right..
		\end{equation}	
		
			Depending upon the particular shape, $(z_{i}^{1},z_{i}^{2})_{i=1}^{n}$ the respective  (finite sequence) $\beta_{i}^{0}$  obtained according to formula \eqref{waga_po_y} may or may not be increasing.  However, imprecisely speaking, $\beta_{i}^{0}$ for $i$ small are smaller than
			$\beta_{i}^{0}$ for $i$ large. This means that 
			$\beta_{i}^{0}$ which defines a respective discrete distribution, 'favorizes' (puts greater probability) on the final part of the shape. 
			
			In contrary to this, by using $\beta_{i}^{0}$ and $\alpha_{j}^{0}$ from \eqref{waga_po_y} we define
			a discrete distributions which 'favorizes' (puts greater probability) on the initial part of the shape.
			
				\begin{equation}
		\label{waga_po_y_dlak}
		\bar{\beta}^{0}_{i}=\left\{\begin{array}{ll}\frac{1-\beta_{i}^{0}}{(k_2-k_1)-1}& \ \ \ for \ i=k_1,...,k_2\\
         0& \ \ otherwise
    \end{array}\right.,\
		\bar{\alpha}^{0}_{j}= \left\{ \begin{array}{ll}
	\frac{1-\alpha_{j}^{0}}{(l_2-l_1)-1}& \ \ \ for \ j=l_1,...,l_2\\
         0& \ \ otherwise
    \end{array}\right.,
		\end{equation}	
		
\end{description}

	\section{Partial optimal transport}
	\label{Section_Partial_Optimal_Transport}
	In the general form the discrete optimal transport problem (OT) can be represented as
\begin{equation}
    \label{eq_ot_1}
    OT(x,y;C)=\left\{ \begin{array}{rl}
         \min_{\pi \in\mathbb{R}^{n\times m}}&\sum_{ij}\pi_{ij}c_{ij} \\
         s.t.&\pi\ge 0\\
         &\sum_{j}\pi_{ij}\leq \beta_{i}
         \ \forall i=1,...,n\\
         &\sum_{i}\pi_{ij}\leq\alpha_{j}\ \forall j=1,...,m\\
    \end{array}\right.,
\end{equation}
where  $C$ is an $n\times m$ matrix.

Let $\nu_i, \mu_j\geq 0$, $i=1,...,n$, $j=1,...,m$. Then partial optimal transport takes the form
\begin{equation}
    \label{eq_ot_2}
    POT(x,y;C)=\left\{ \begin{array}{rl}
         \min_{\pi \in\mathbb{R}^{n\times m}}&\sum_{ij}\pi_{ij}c_{ij}-\sum_{j}\mu_j\pi_{ij}-\sum_{i}\nu_i\pi_{ij} \\
         s.t.&\pi\ge 0\\
         &\sum_{j}\pi_{ij}\leq \beta_{i}
         \ \forall i=1,...,n\\
         &\sum_{i}\pi_{ij}\leq\alpha_{j}\ \forall j=1,...,m\\
    \end{array}\right.,
\end{equation}
\begin{equation}
    \label{eq_ot_3}
    POT(x,y;C)=\left\{ \begin{array}{rl}
         \min_{\pi \in\mathbb{R}^{n\times m}}&\sum_{ij}(c_{ij}-\mu_j-\nu_i)\pi_{ij} \\
         s.t.&\pi\ge 0\\
         &\sum_{j}\pi_{ij}\leq \beta_{i}
         \ \forall i=1,...,n\\
         &\sum_{i}\pi_{ij}\leq\alpha_{j}\ \forall j=1,...,m\\
    \end{array}\right.,
\end{equation}
Equivalently, by putting $A= [A_1, A_2 ]^{T}$ 
\begin{equation}
    \label{eq_ot_4}
    POT(x,y;C)=\left\{ \begin{array}{rl}
         \min_{\pi \in\mathbb{R}^{n\times m}}&\sum_{ij}(c_{ij}-\mu_j-\nu_i)\pi_{ij} \\
         s.t.&\pi\ge 0\\
         &A_1\pi\leq \beta\\
         & A_2\pi\leq\alpha\ \\
    \end{array}\right.,
\end{equation}
Let us construct the dual problem to \eqref{eq_ot_4}. Let $p\in \mathbb{R}^n$, $q\in \mathbb{R}^m$
\begin{equation}
    \label{eq_ot_5}
    DPOT(x,y;C)=\left\{ \begin{array}{rl}
         \max_{p \in\mathbb{R}^{n},q\in \mathbb{R}^m }&\sum p^T\mu +q^T\nu\\
         s.t.&p,q \le 0\\
         &[A_1, A_2]^T[p, q]\leq [c_{ij}-\mu_j-\nu_i]^T,\forall i=1,...,n, j=1,...,m \\
        
    \end{array}\right.,
\end{equation}
Optimality conditions for problems \eqref{eq_ot_4} and  \eqref{eq_ot_5} can be formulated via complementary slackness theorem. Let $\pi^*,p^*,q^*$ be optimal solutions of \eqref{eq_ot_4} and  \eqref{eq_ot_5} respectively. Then

$$
\pi^*([A_1, A_2][p^*, q^*]^T- [c_{ij}-\mu_j-\nu_i]^T)=0, \ \ \ \forall i=1,...,n, j=1,...,m.
$$
If for some $i,j$ we have  
\begin{equation}
\label{hind}
[c_{ij}-\mu_j-\nu_i]^T>0
\end{equation}
then $[A_1, A_2][p^*, q^*]^T- [c_{ij}-\mu_j-\nu_i]^T\neq 0$, in consequence $\pi^*_{ij}=0$.

\begin{example}
Let us consider two vectors 
$V=[(x_1^{V},y_1^{V}),  (x_2^{V},y_2^{V}),...,(x_n^{V},y_n^{V})]$ and
$W=[(x_1^{W},y_1^{W}),  (x_2^{W},y_2^{W}),...,(x_m^{W},y_m^{W})]$.
Let us define $c_{ij}$ as  the Euclidean distance between any two points from $V$ and $W$ i.e.
$$
c_{ij}:=\sqrt{(x_{i}^{V}-x_{j}^{W})^2+(y_{i}^{V}-y_{j}^{W})^2}
$$
We would like to guarantee that the inequality \eqref{hind} i.e.  $c_{i,j}>\nu_{i}+\mu_{j}$ holds for some given parts of vectors $V$ and $W$. 

Let $t\in \{1,...,n\}$ and $k\in\{1,...,m\}$. Let $A=\{(i,j) \ :\ i\leq t \wedge j\leq k,i=1,...,n, j=1,...,m \}$.  We are looking for numbers $\nu_{i}, \mu_{j}\geq0$, $i=1,...,n$, $j=1,...,m$  such that for every $(i,j)\notin A$ we have $c_{ij}>\nu_{i}+\mu_{j}$.
    
    Let us consider the following example: $n=3$, $m=4$,
    $V=[(1,0.1), (2,0.2), (3,0.3) ]$ and $W=[(1,0.2), (3,0.8),(5,0.6),(6,0.7)]$.
    Then the matrix $c_{ij}$ is as follows 
    \begin{lstlisting}
    c =
    0.1000    2.1190    4.0311    5.0359
    1.0000    1.1662    3.0265    4.0311
    2.0025    0.5000    2.0224    3.0265
    \end{lstlisting}
    In this example we assume that $t=2$ and $k=2$. Then $A=\{(1,1),(1,2),(2,1),(2,2)\}$.
    We choose
    
    $\nu=$
    \begin{lstlisting}
    2.4839
    1.4893
    0.4950
    \end{lstlisting}
    and 
    $\mu=$
    \begin{lstlisting}
    1.4875
         0
    1.5070
    2.5014
     \end{lstlisting}
    Then, the new cost matrix $d_{ij}=c_{ij}-\nu_{i}-\mu{j}$ takes the form 
    
    \begin{lstlisting}
    d =
   -3.8714   -0.3650    0.0402    0.0505
   -1.9768   -0.3231    0.0302    0.0404
    0.0200    0.0050    0.0204    0.0301
     \end{lstlisting}.
     
     Now we solve the problem $POT$ with $\nu$, $\mu$ .... for our example. We get the following optimal solution

      \begin{lstlisting}
     x =
     0.2500         0         0         0
         0          0         0         0
         0          0         0         0
     \end{lstlisting}.
     
     \end{example}
     
         


 \section{Experiments}
 \label{Section_Experiments}

 \subsection{The data used in experiments}

 In the present work the data are 2D curves which represent the contour of a rationally symmetric object. In our specific case, the data comes from the field of archaeology, and specifically from the study of shapes of ancient ceramic vessels.
We assume that the information about the shape of the curve is stored in the form of a finite-dimensional vector $x\in\mathbb{R}^{n}$ with  coordinates $(x_{i}^{1},x_{i}^{2})$, $i=1,...,n$. This vector  is a discretization of a smooth curve, as an accurate representation of the original boundary of the shape.

The specificity of our research comes from the application, which requires the classification of shapes according to subtle differences between them. To this aim we propose a series of similarity measures based on the Wasserstein distance, that can be used for the clustering of given data, as it was done in \cite{C&C} and \cite{AMCS_A&M}. Moreover, we propose tools which allow to focus the attention of the similarity measure on certain parts of our objects (curves), see Section \ref{Section_other_distributions} and \ref{Section_Partial_Optimal_Transport}. The clustering step itself is, however, is conducted according to standard methods available through MatLab Toolbox. 

\begin{figure}[h]
 	\centering
 	\includegraphics[width=0.8\linewidth, height=0.17\textheight]{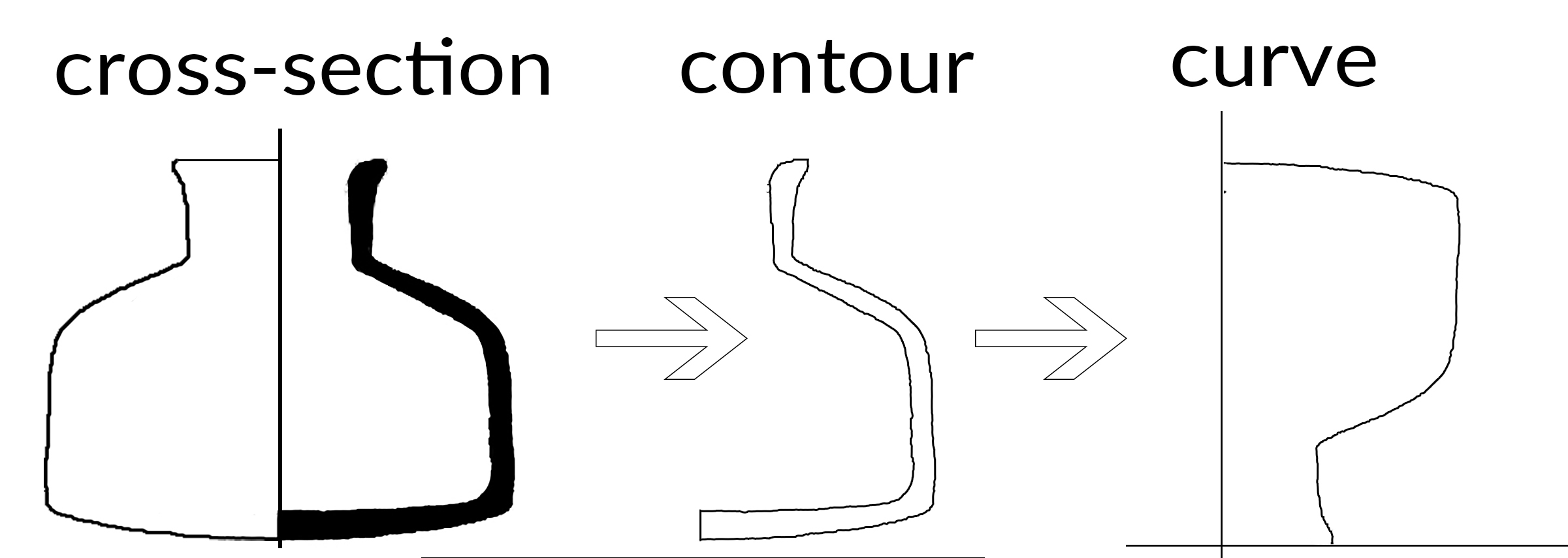}
	\caption{The process of generating of a curve from a cross section.}
 	\label{profil_na_krzywa}
 \end{figure}

 The process of generating of a histogram from a given curve is presented in Figure \ref{profil_na_krzywa}.  

We have conducted all the experiments on five  datasets, labeled from 1 to 5. The datasets are composed in such a way to differ in terms of homogeneity within groups as well as general similarity between groups. Taking into account the sizes of real-life  datasets, and in order to preserve readability of the results, our datasets were limited in number of objects, with the maximum of 45 objects in the set.  
The Data sets were constructed as follows:
\begin{itemize}
    \item Set 1: 19 objects, five shape groups. Consistent size throughout the set, with the exception of object 13. Objects within groups vary in size, but not significantly . 

    \item Set 2: 45 objects, 7 shape groups. Size between groups somewhat consistent throughout the set, slight size variation withing groups.  

    \item Set 3: 36 objects, 5 groups. Large variation in size between groups, and some size variation within the groups.

    \item Set 4: 40 elements, 8 groups. Large variation in size between groups, little variation in size within groups.

    \item Set 5: 45 objects, 9 groups. Large variation in size between groups, little variation in size within groups. 
\end{itemize}

The performed experiments are based on the measures described in Sections \ref{Wass_dist}, \ref{Section_Wass_for_different_distributions}, \ref{Section_Partial_Optimal_Transport} and presented in Sections  \ref{Section_experiments_full}, \ref{Section_full_external}, and \ref{Section_experiments_constrained} and summarized in Table \ref{table_results}. In order to quantify the obtained results, all of the dendrograms were evaluated by an expert. For every experiment discussed below we present a percentage of the elements that are, in the expert’s opinion, correctly classified. As the presented scheme is meant to be of assistance to experts, we choose an expert-based evaluation for our results. An automated or semi-automated assessment of results based on the dendrograms, and the choice of a correct similarity threshold is itself a matter of ongoing research (e.g., \cite{MLCut}, \cite{AMCS2}).

\subsection{Experiments: The Wasserstein distance-basic formulation}
\label{Section_experiments_full}

In the presented experiments the Wasserstein distance is calculated according to Formula \eqref{WD}, with weights chosen according to \eqref{waga_uniform}. This is the most general formulation of the Wasserstein distance. The experiments were conducted in order to establish a baseline in the performance of the method.

\begin{figure}[h]
	\centering
	\includegraphics[width=0.75\linewidth, height=0.3\textheight]{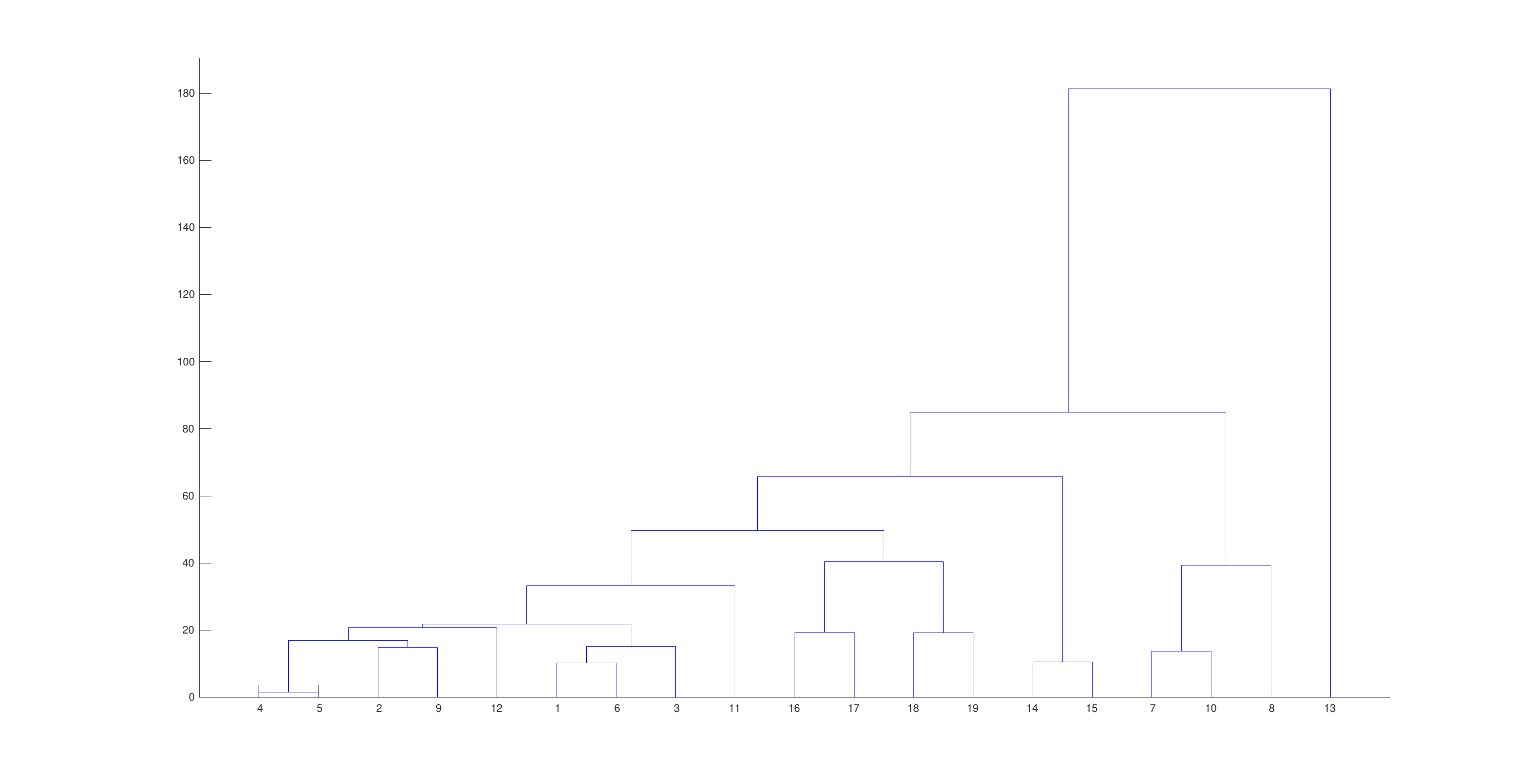}
	\caption{Results of clustering of Set 1 according to Formula \eqref{WD}, with weights chosen according to \eqref{waga_uniform}.}
	\label{Wass_set1}
\end{figure}

\begin{figure}[H]
	\centering
	\includegraphics[height=0.3\textheight]{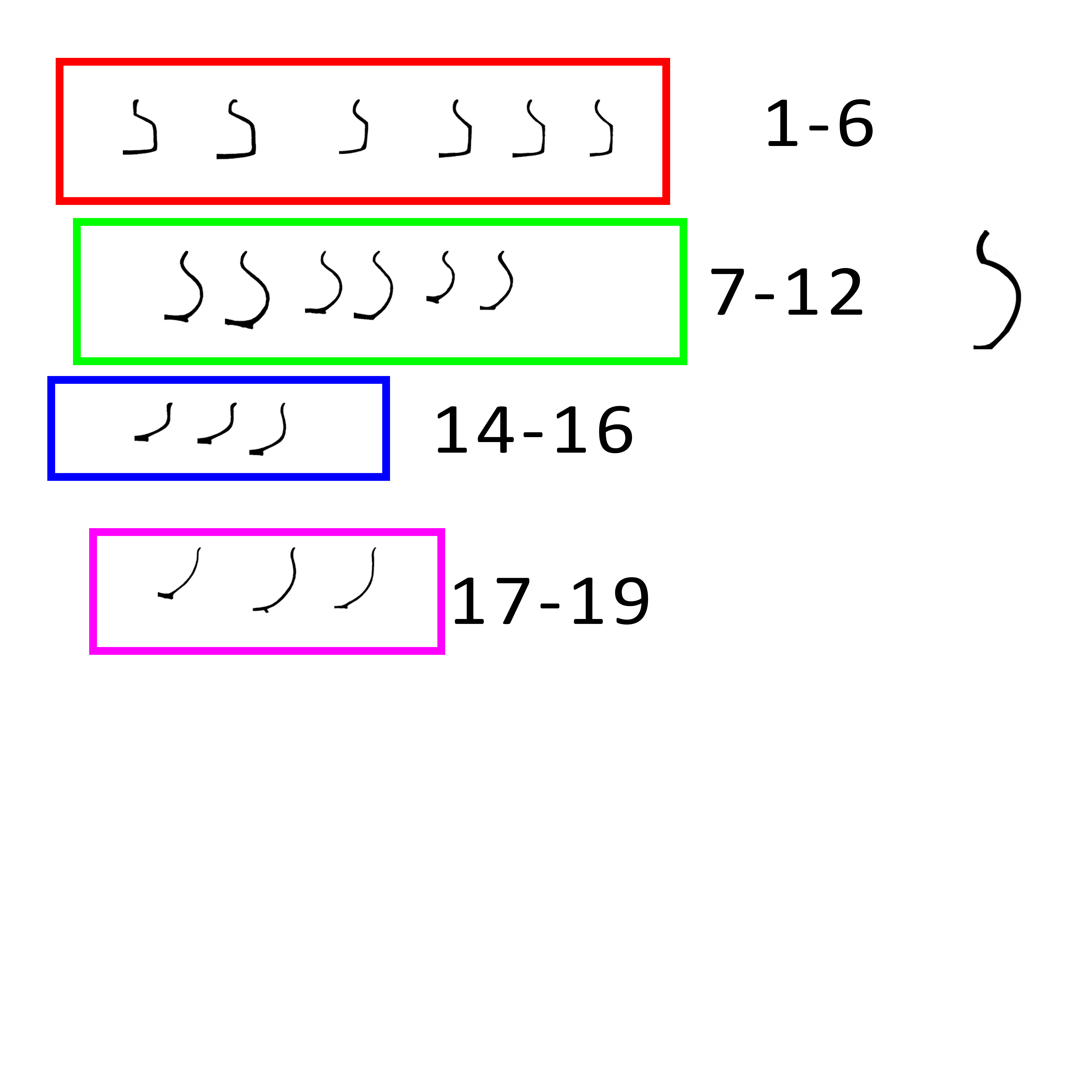}
	\caption{Set 1.}
	\label{Set1}
\end{figure}

After the clustering of the values obtained in the similarity matrix we obtain the classification, which is visualized in the form of a dendrogram. For the purpose of this discussion, this experiment was conducted for Set 1 (Figure \ref{Set1}). Results obtained for Set 1 are presented in Figure \ref{Wass_set1}. 

\begin{figure}[h!]
   	\centering
	\includegraphics[width=0.9\linewidth, height=0.40\textheight]{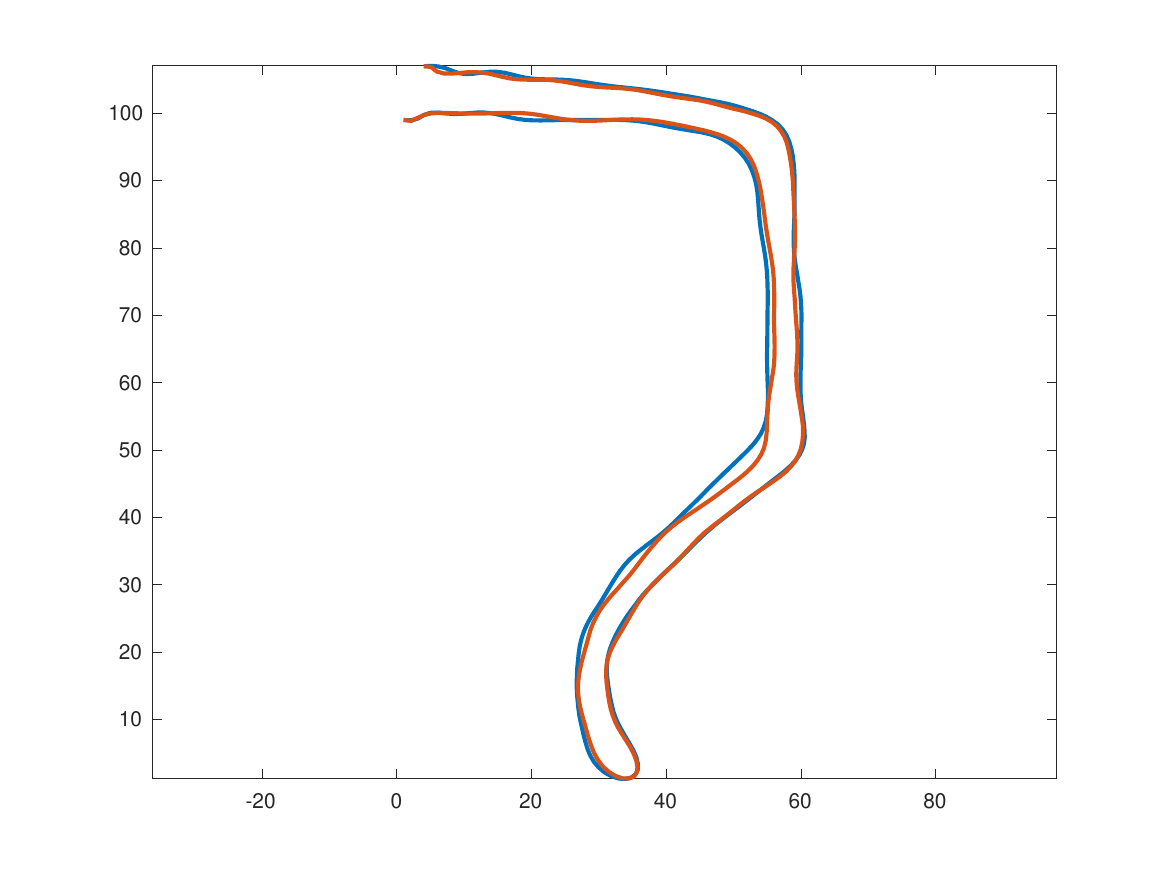}
	\caption{Curves, representing contours of objects 4 and 5 plotted in $x,y$ coordinates: 4 -blue; 5 -orange (Note: the curves are plotted upside down to facilitate calculation).}
    \label{4and5}
\end{figure}

We note that objects 4 and 5 share a very close resemblance (Figure \ref{4and5}), and as a result are grouped together (Figure \ref{Wass_set1}). These are not only very similar in therms of overall shape but are also almost of the exactly same size. 

\begin{figure}[h]
    
	\centering
	\includegraphics[width=0.9\linewidth, height=0.5\textheight]{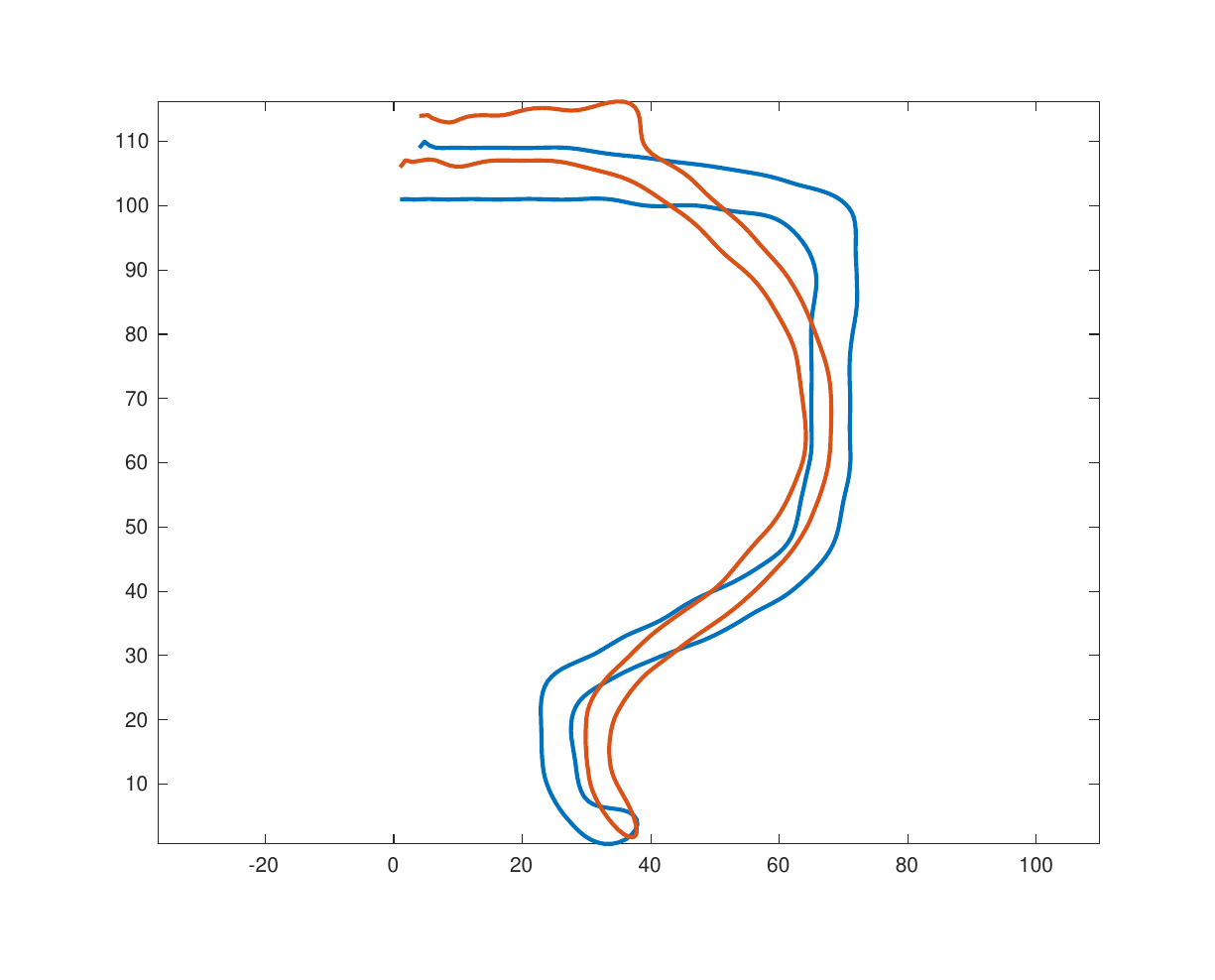}
	\caption{Curves, representing contours of objects 2 and 9 plotted in $x,y$ coordinates: 2 -blue; 9 -orange (Note: the curves are plotted upside down to facilitate calculation).}
	\label{2and9}
\end{figure}

A curious case occurs between objects 2 and 9. Although they share a very general similarity, the details of the shape do not align (different curvature of the body, different type of base). However, by coincidence, these two curves aside for being very close in size (2- 366 points, 9- 305 points), overlap in large sections as demonstrated in Figure \ref{2and9}. Such a relation means that the cost of "transporting" curve 2 into curve 9 is relatively low, lower than "transporting" e.g. curve 9 to 10.

An important remark is that the significantly larger 13 will be considered an outlier. It is composed of 534 points, while the next large object has only 306 (the shortest curve has 206 points). This large difference in the "size" decides of its position as an outlier- a tendency that will be visible throughout all experiments. 
For Set 1 we have obtained a score of 73,7\%.

\begin{figure}[h]
	\centering
	\includegraphics[width=0.8\linewidth, height=0.3\textheight]{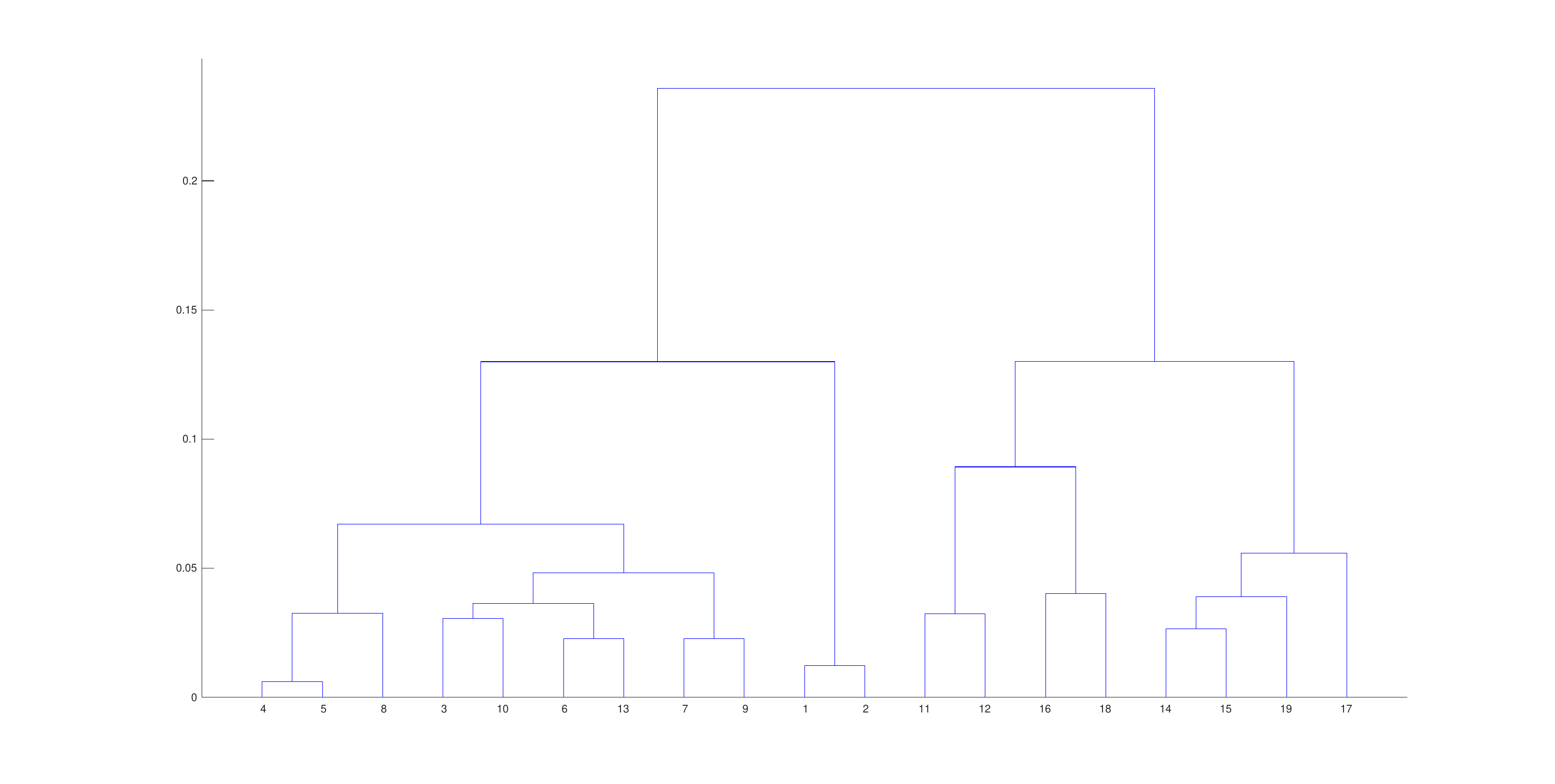}
	\caption{Clustering result using the Procrustes analysis.}
	\label{Procrustes}
\end{figure}

As a reference for the performance of the Wasserstein methods, we have chosen to perform a classification using Procrustes Analysis (\cite{AMCS_A&M}). Although well suited to compare shapes, Procrustes Analysis does not involve size comparison. This can be clearly seen by the incorporation of 13 into the group 6, 3, 10, despite its large size. This is also the only of the performed experiments, where 2 and 9 do not appear paired. This hints to the Wasserstein distance emphasising their closeness in size over the shape similarity. The results of the classification are shown in Figure \ref{Procrustes}.

In our application the difficulty of applying the Wasserstein distance to the clustering of shapes lies in it's very particular way of handling size/distance between compared objects/set. Originally formulated to compare statistical distributions, the Wasserstein distance, generally, does not depend on the $x$ values assumed by the distributions. However, as will be shown in the experiments below, when comparing several objects their physical size, represented by the distribution of points, is relevant to the resulting measure, even if the number of points is equal for all objects. 

\subsection{Processing of the data for the experiments}
\label{Section_full_external}
In our initial experiments we have employed the basic scheme of Wasserstein distance (Formula \eqref{WD}) to compare curves representing full cross-sections of the investigated objects. In order to reduce the computation time we have conducted an experiment where we only used the external half of the section defined as
$$
(\bar{x}_i^1, \bar{x}_i^2)= (x_i^1, x_i^2), \ \text{for} \  i=\hat{I},..., n,\ \text{where} \  x^2_{\hat{I}}=\min{x^2_{i}}.
$$
The results obtained through this scheme were identical or very close to those conducted with the full sections, including both the internal and external sections. Hence we have decided to choose this approach, using only the external part of the section, as our basic scheme.

Additionally, we have found that moving the (discrete) curves to a common center of the mass $(u,w)$ i.e.
$$
(u_{x},w_{x})=(\frac{\sum_{i=1}^n(x_{i}^1)}{n},\frac{\sum_{i=1}^n(x_{i}^2)}{n} ), \ \ \  (u_{y},w_{y})=(\frac{\sum_{j=1}^m(y_{j}^1)}{m},\frac{\sum_{j=1}^m(y_{j}^2)}{m} )
$$
$$
(u,w)=(\frac{u_{x}+u_{y}}{2}, \frac{w_{x}+w_{y}}{2} )
$$
allows us to obtain more refined results. This can be explained by a better correspondence of the shapes and it reduces the intersecting of sections, which can result in problems for the Wasserstein metric. This approach allowed us to avoid the problems described in Section \ref{Section_experiments_full}.

\begin{figure}[H]
	\centering
	\includegraphics[width=0.9\linewidth]{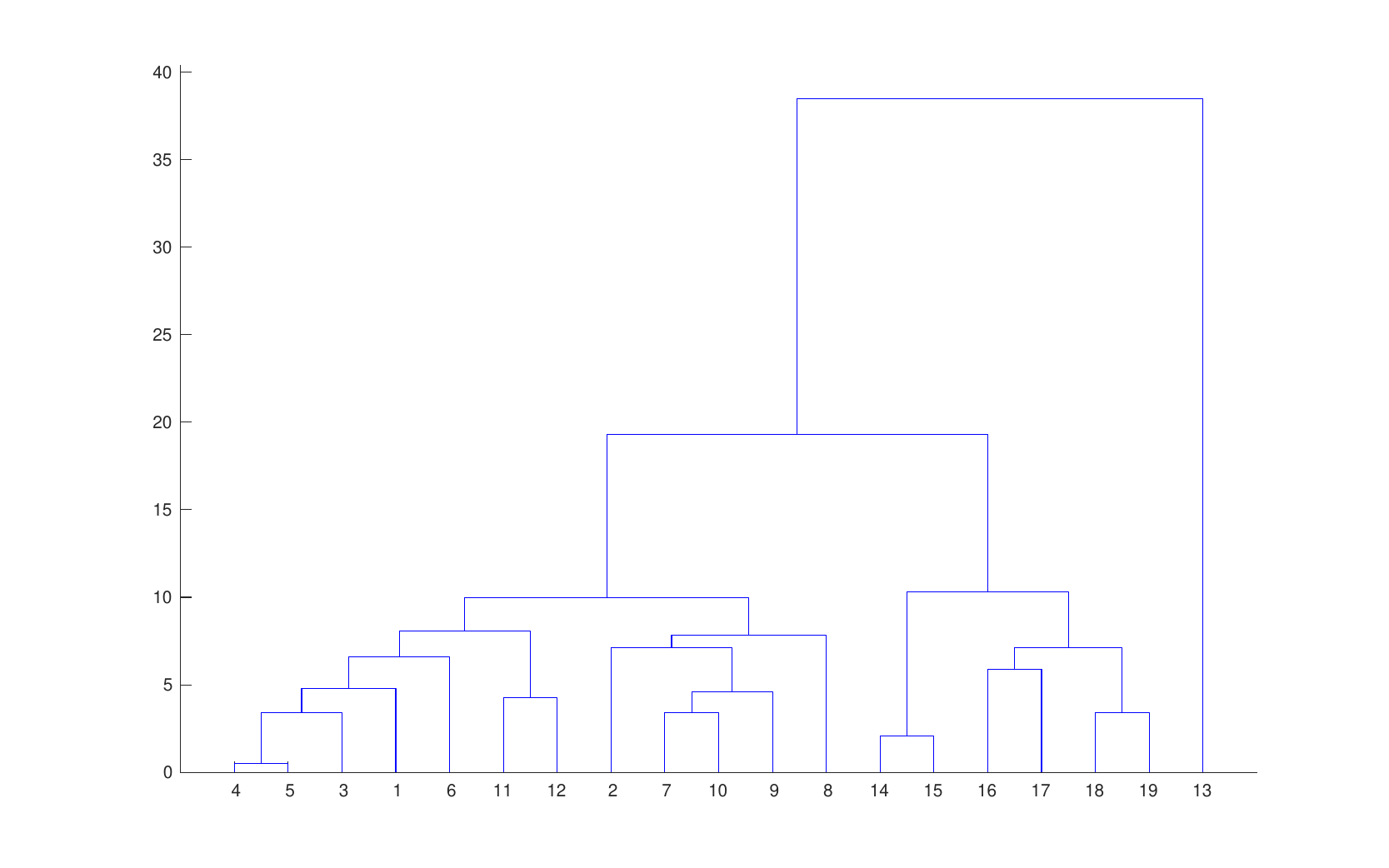}
	\caption{Set 1, Wasserstein distance calculated for whole curves, according to Formula \ref{WD}, with weights  $1/n$ according to \ref{waga_uniform}.}
	\label{Set1_cale}
\end{figure}

\begin{figure}[H]
	\centering
	\includegraphics[width=0.9\linewidth]{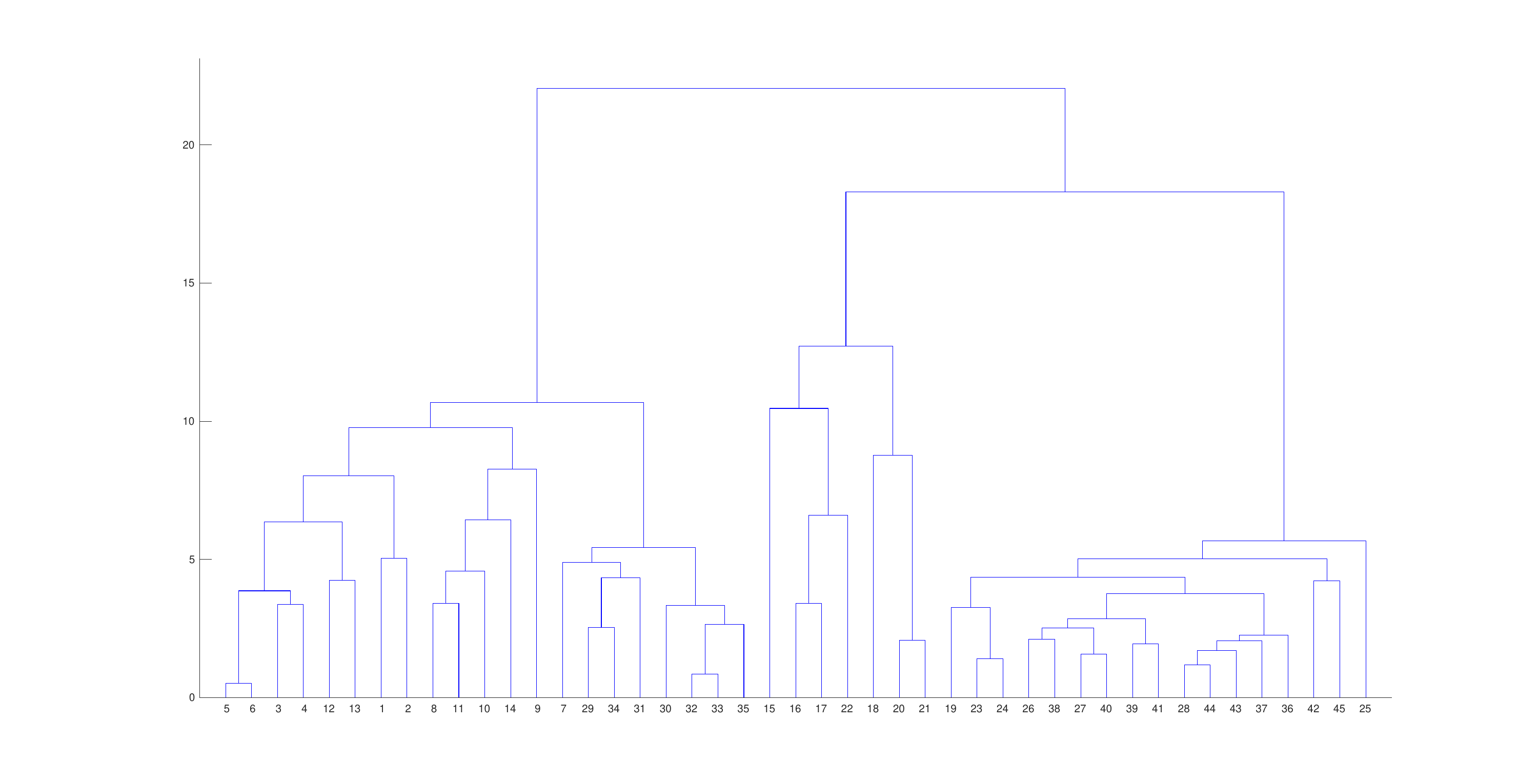}
	\caption{Set 2, Wasserstein distance calculated for whole curves, according to Formula \ref{WD}, with weights  $1/n$ according to \ref{waga_uniform}.}
	\label{Set2_cale}
\end{figure}

\begin{figure}[H]
	\centering
	\includegraphics[width=0.9\linewidth]{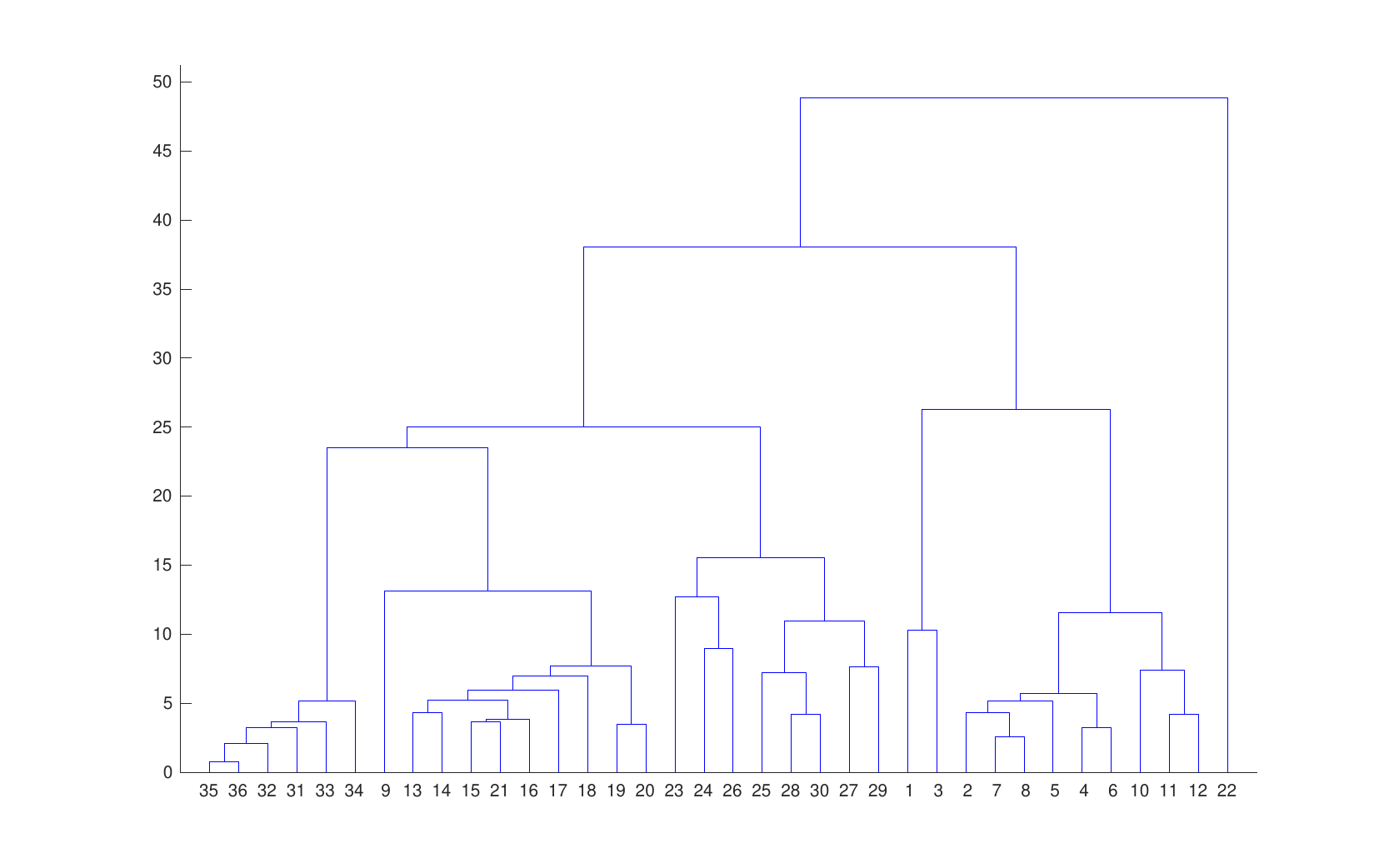}
	\caption{Set 3, Wasserstein distance calculated for whole curves, according to Formula \ref{WD}, with weights  $1/n$ according to \ref{waga_uniform}.}
	\label{Set3_cale}
\end{figure}

\begin{figure}[H]
	\centering
	\includegraphics[width=0.9\linewidth]{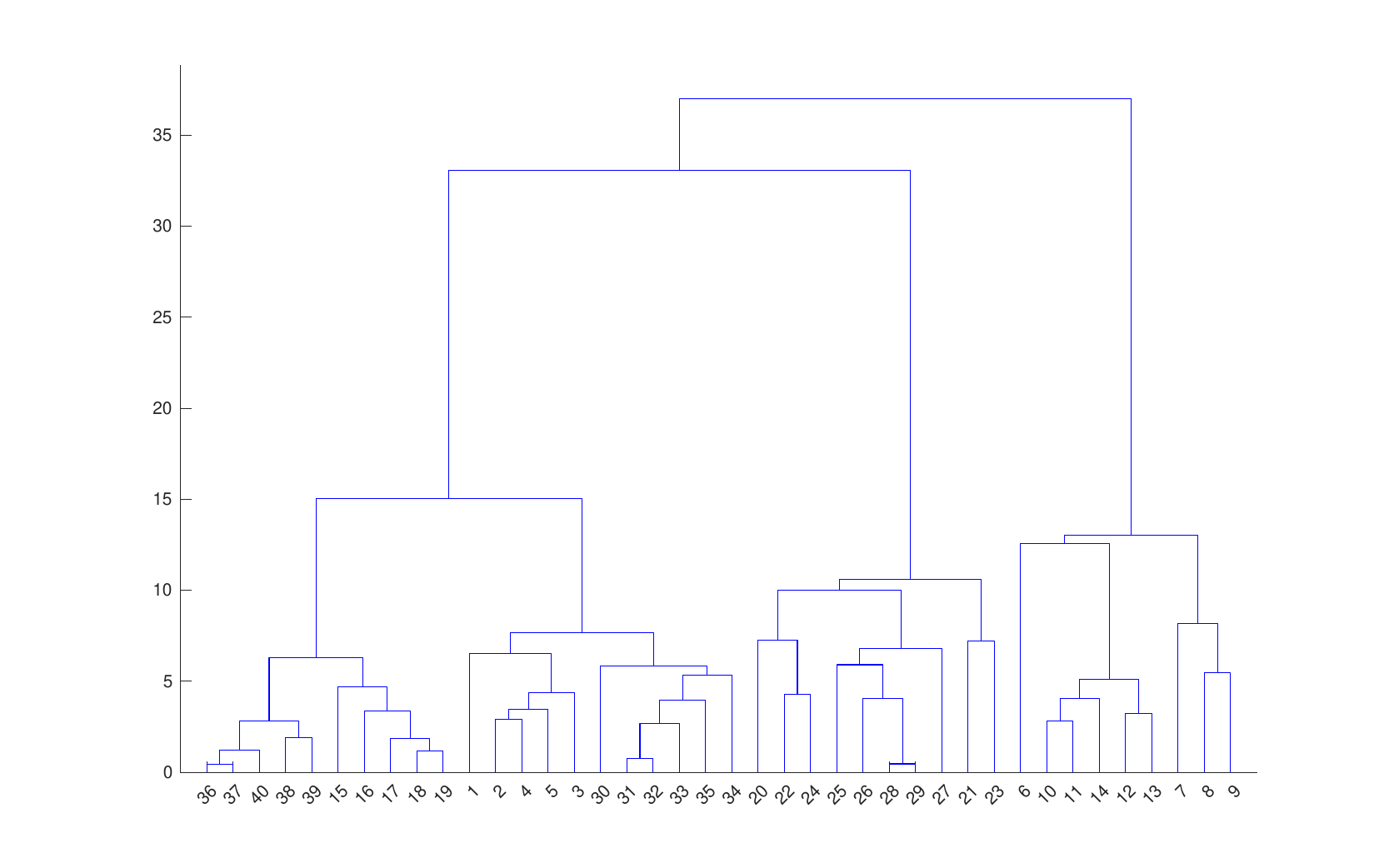}
	\caption{Set 4, Wasserstein distance calculated for whole curves, according to Formula \ref{WD}, with weights  $1/n$ according to \ref{waga_uniform}.}
	\label{Set4_cale}
\end{figure}

\begin{figure}[H]
	\centering
	\includegraphics[width=0.9\linewidth]{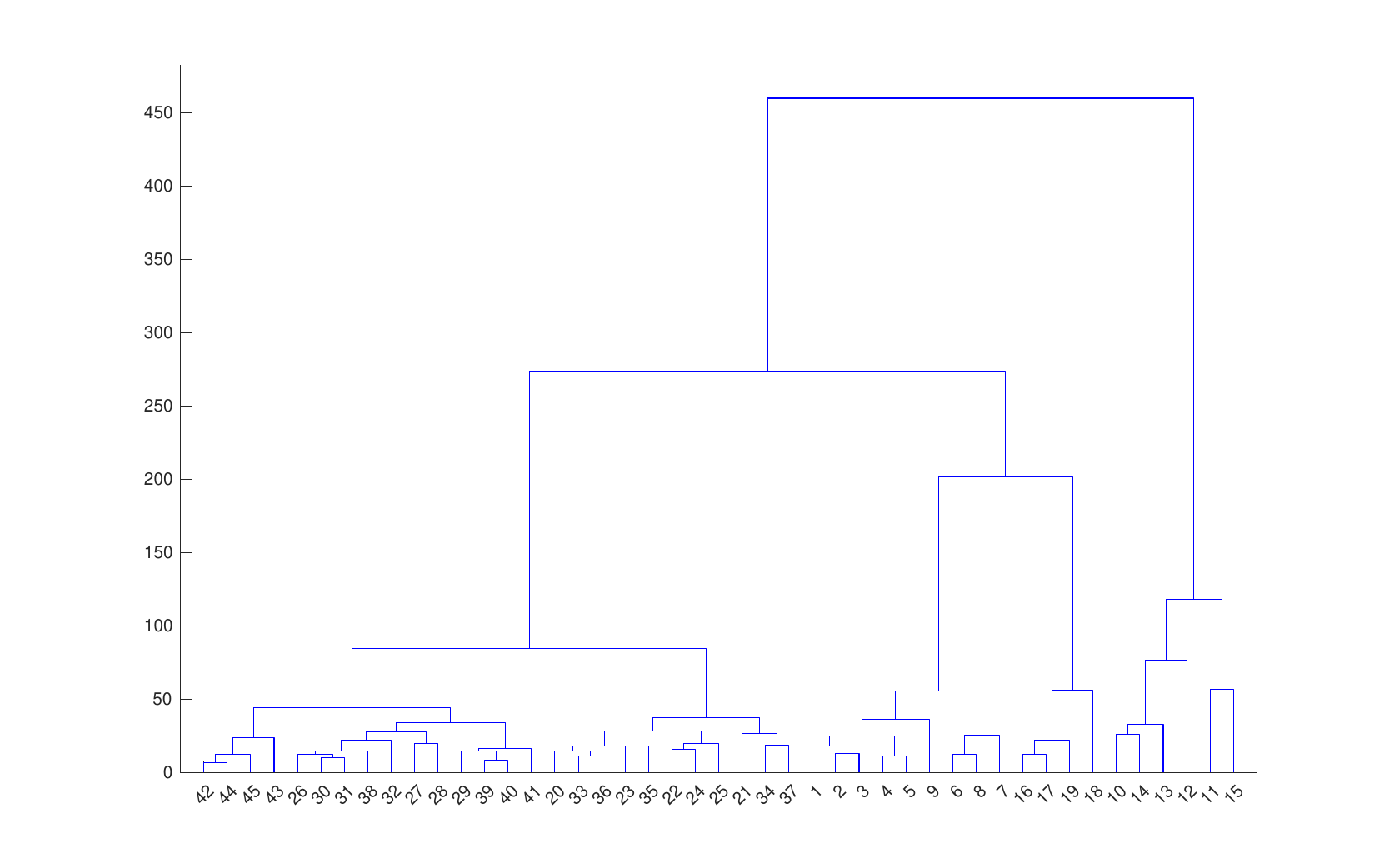}
	\caption{Set 5, Wasserstein distance calculated for whole curves, according to Formula \ref{WD}, with weights  $1/n$ according to \ref{waga_uniform}.}
	\label{Set5_cale}
\end{figure}

In order to investigate the performance of the proposed method we have conducted Experiment 1. The results are shown in Figures \ref{Set1_cale}, \ref{Set2_cale}, \ref{Set3_cale}, \ref{Set4_cale}, \ref{Set5_cale}, and  summarized in Table \ref{table_results}.

\subsection{Experiments on constrained data}
\label{Section_experiments_constrained}

As part of the conducted experiments we have attempted to use the Wasserstein scheme as described in Section \ref{Wass_dist} to constrained data. It means we limit our attention to some chosen parts of the curve.  For the purpose of this paper, we have chosen to focus on the uppermost fragments of our curves to be compared. It is important to mention here,that for computational purpose, the curves are inverted in the process, so the uppermost points become the lowermost. We have chosen a 1/6 height cut-off point as the appropriate choice of  $k_1, k_2$ and $l_1, l_2$. This is an arbitrary choice and other cut-off points can be viable as well. The proper choice of the cut off point should be a point of separate studies. It relies heavily on the expert's opinion and the data set in question. Nonetheless, the experiments presented in this section prove the utility of the Wasserstein distance in the classification of constrained data. 

A total of seven experiments were conducted for each of the five sets to test the proposed measures restricting the classification to a prescribed section of the curve and focusing the classification to it. 

\begin{figure}[H]
	\centering
	\includegraphics[width=0.9\linewidth]{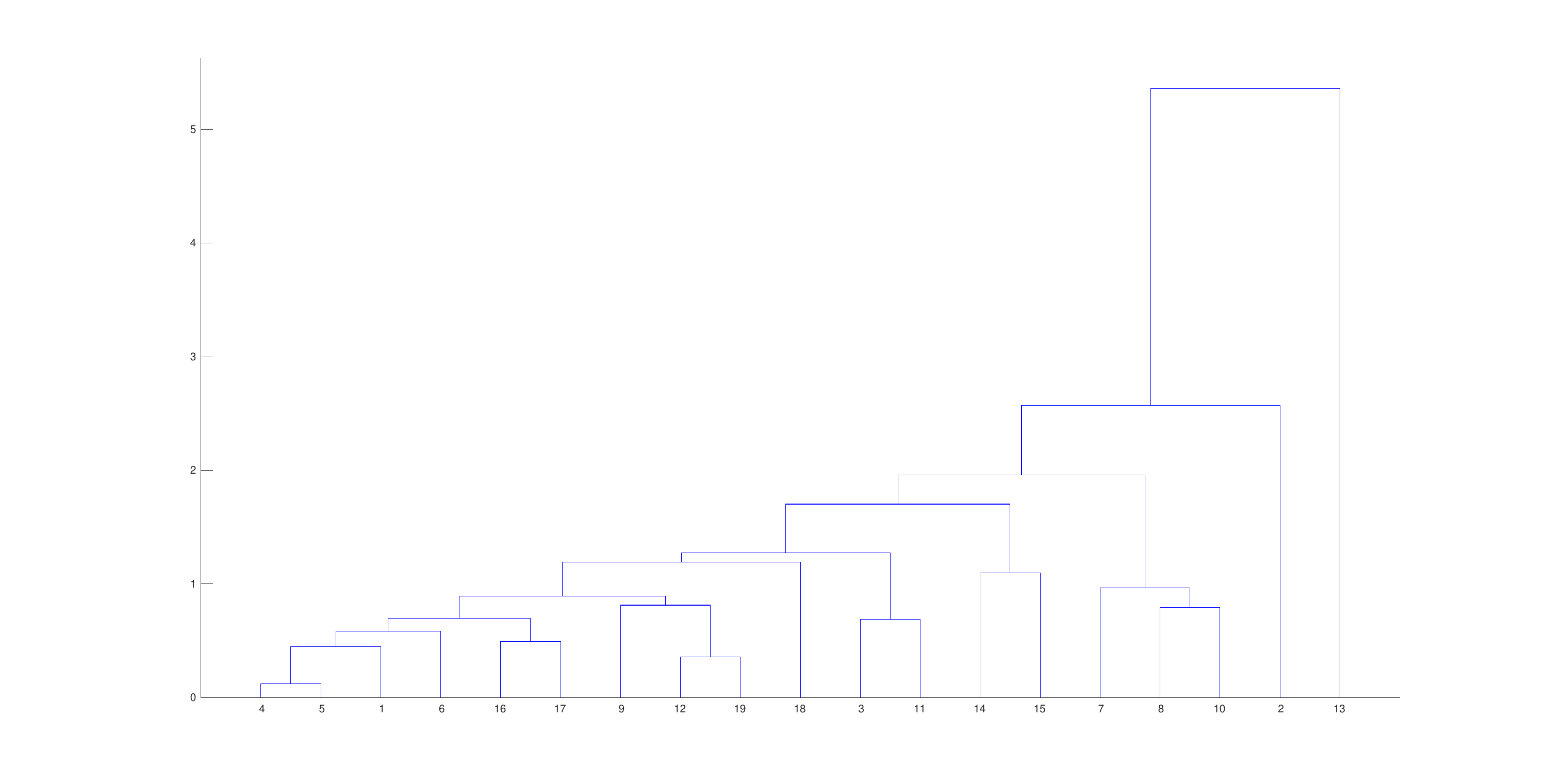}
	\caption{Set 1, Wasserstein distance for 1/6 of the curve, with weight according to \ref{waga_uniform}.}
	\label{Set1_1_6}
\end{figure}

\begin{figure}[H]
	\centering
	\includegraphics[width=0.9\linewidth]{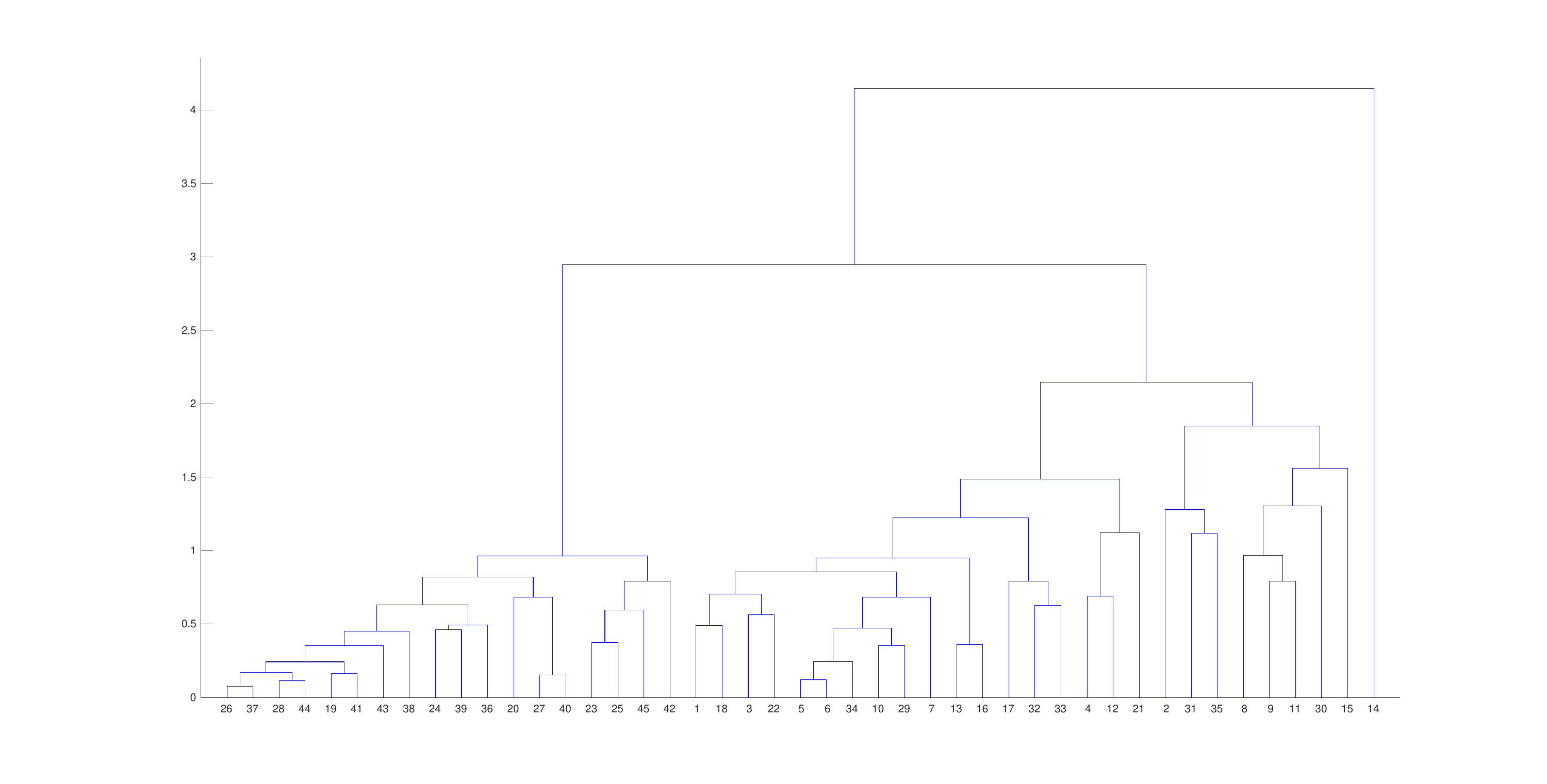}
	\caption{Set 2, Wasserstein distance for 1/6 of the curve, with weight according to \ref{waga_uniform}.}
	\label{Set2_1_6}
\end{figure}

\begin{figure}[H]
	\centering
	\includegraphics[width=0.9\linewidth]{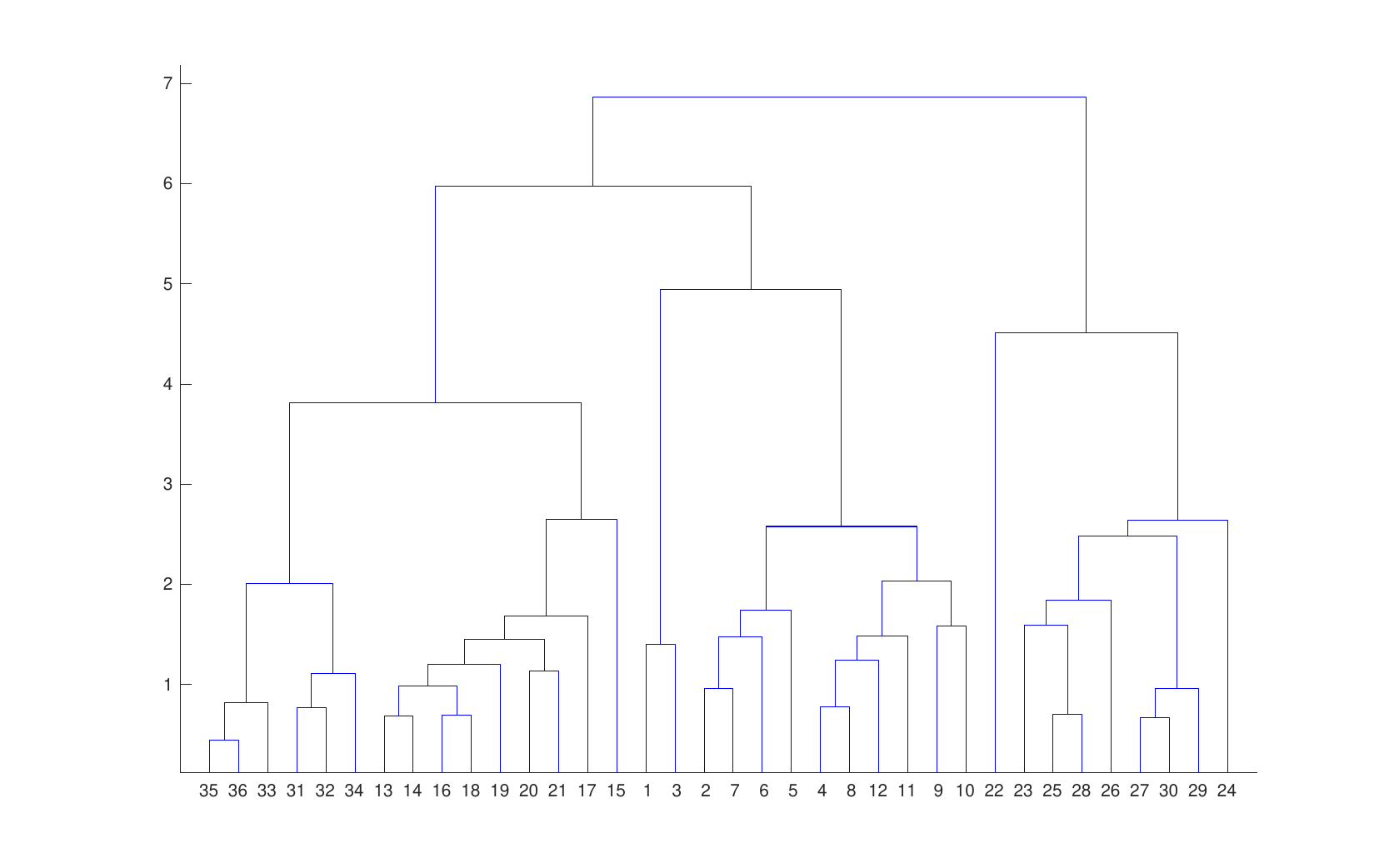}
	\caption{Set 3, Wasserstein distance for 1/6 of the curve, with weight according to \ref{waga_uniform}.}
	\label{Set3_1_6}
\end{figure}

\begin{figure}[H]
	\centering
	\includegraphics[width=0.9\linewidth]{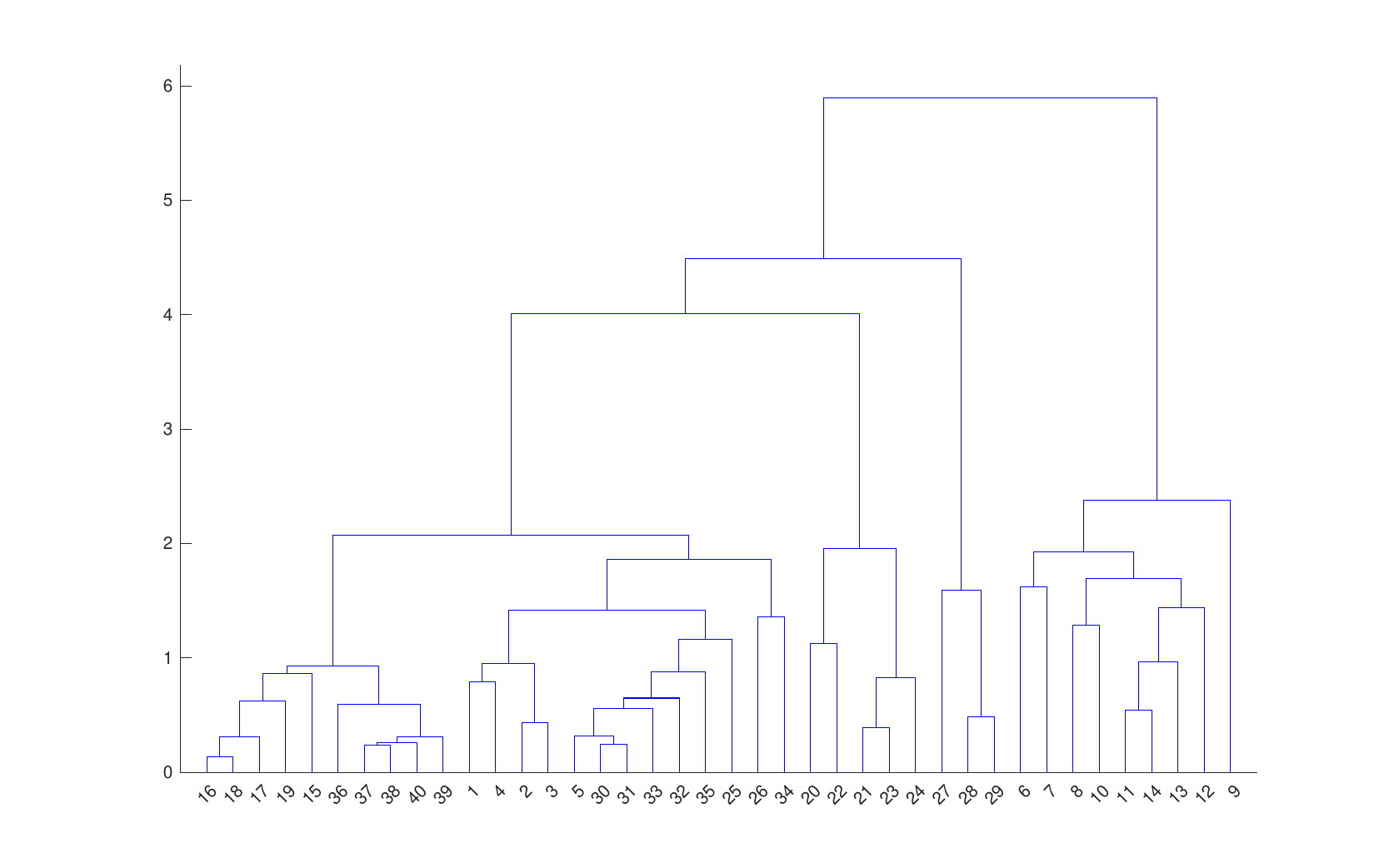}
	\caption{Set 4, Wasserstein distance for 1/6 of the curve, with weight according to \ref{waga_uniform}.}
	\label{Set4_1_6}
\end{figure}

\begin{figure}[H]
	\centering
	\includegraphics[width=0.9\linewidth]{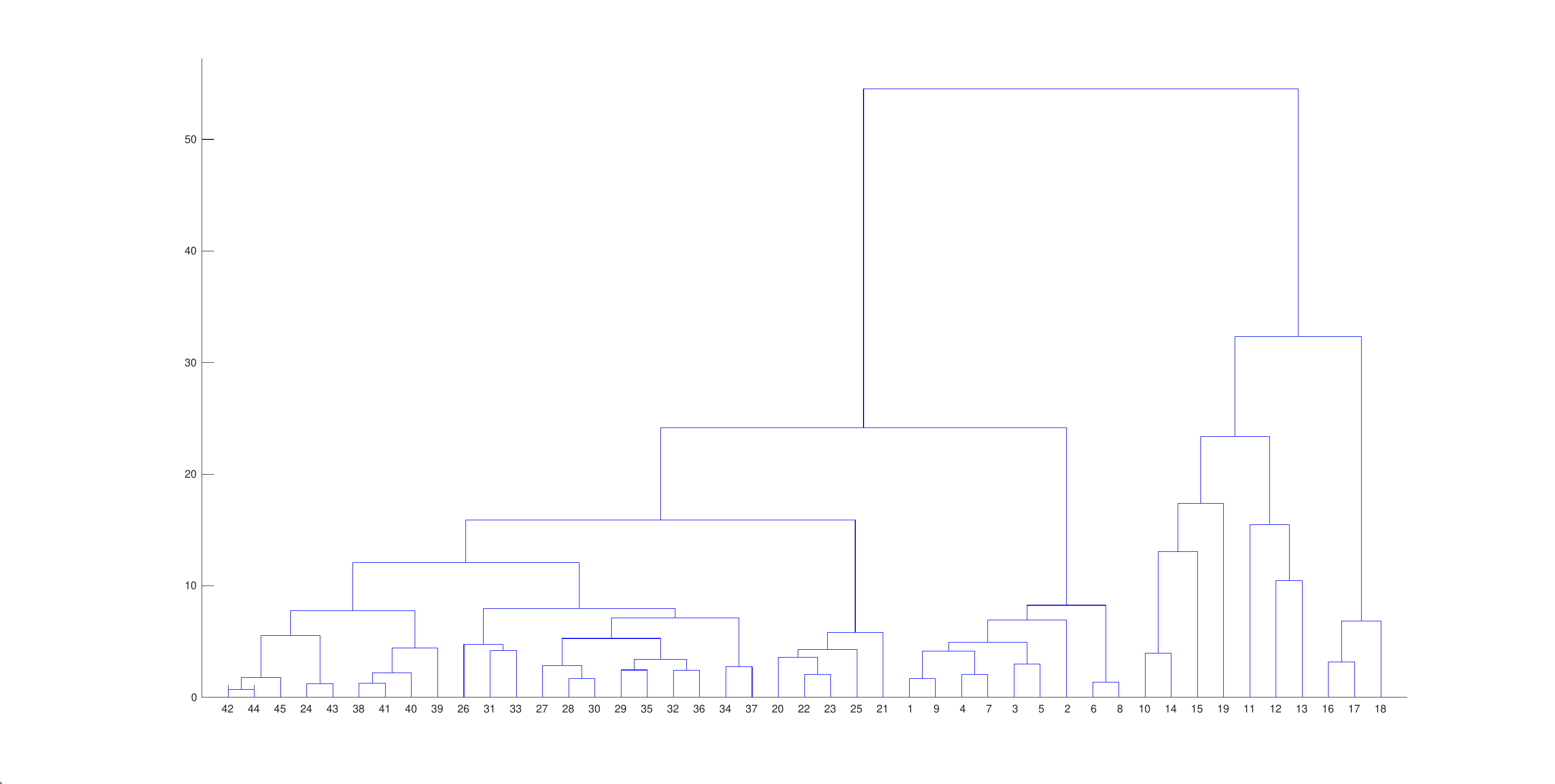}
	\caption{Set 5, Wasserstein distance for 1/6 of the curve, with weight according to \ref{waga_uniform}.}
	\label{Set5_1_6}
\end{figure}

\begin{figure}[H]
	\centering
	\includegraphics[width=0.9\linewidth]{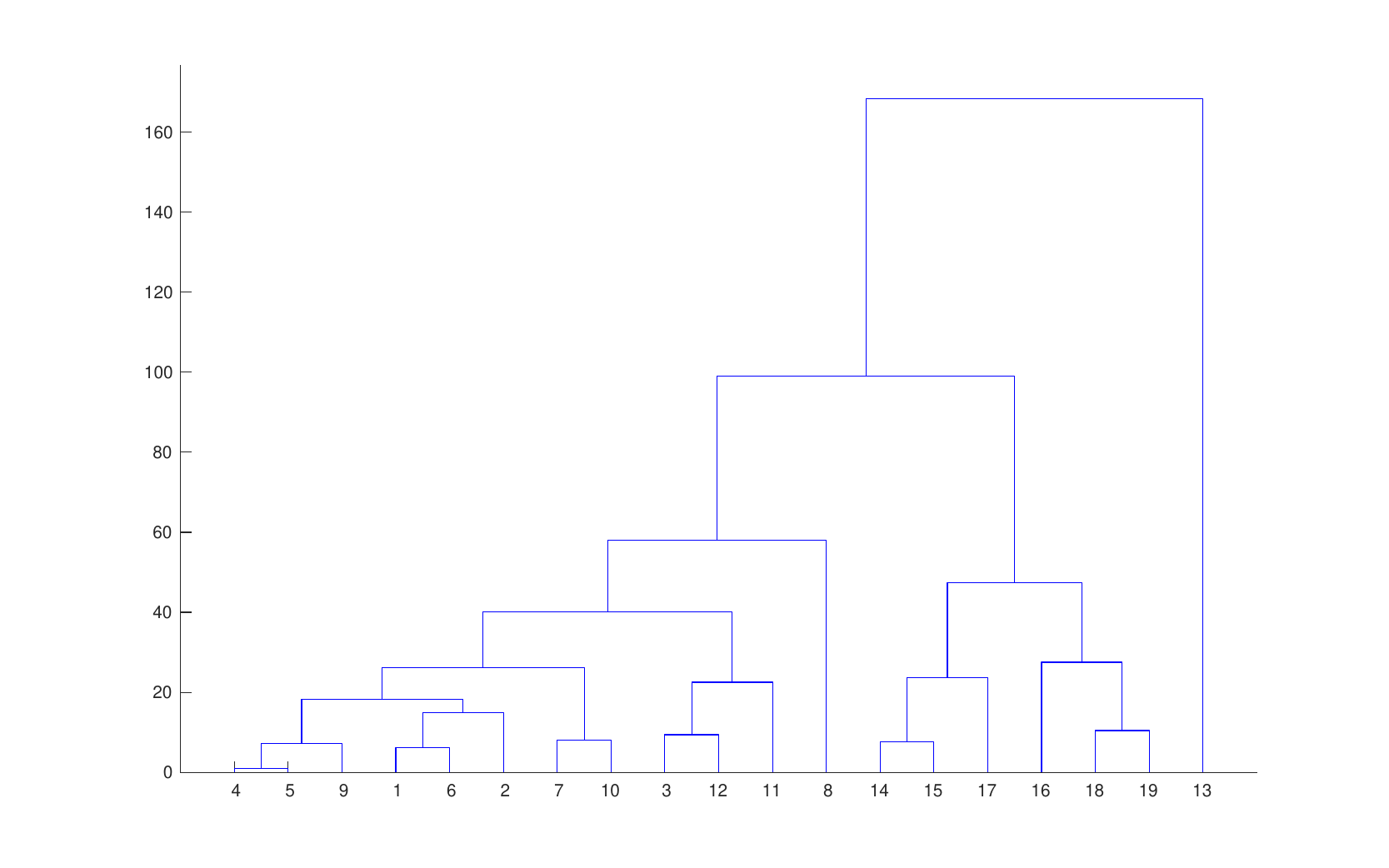}
	\caption{Set 1, Wasserstein distance for the top 1/6 of the curve, with weights according to \ref{waga_uniform}, for the remainder 5/6 of the curve weights of "0" were assigned.}
	\label{Set1_zero1/6}
\end{figure}

\begin{figure}[H]
	\centering
	\includegraphics[width=0.9\linewidth]{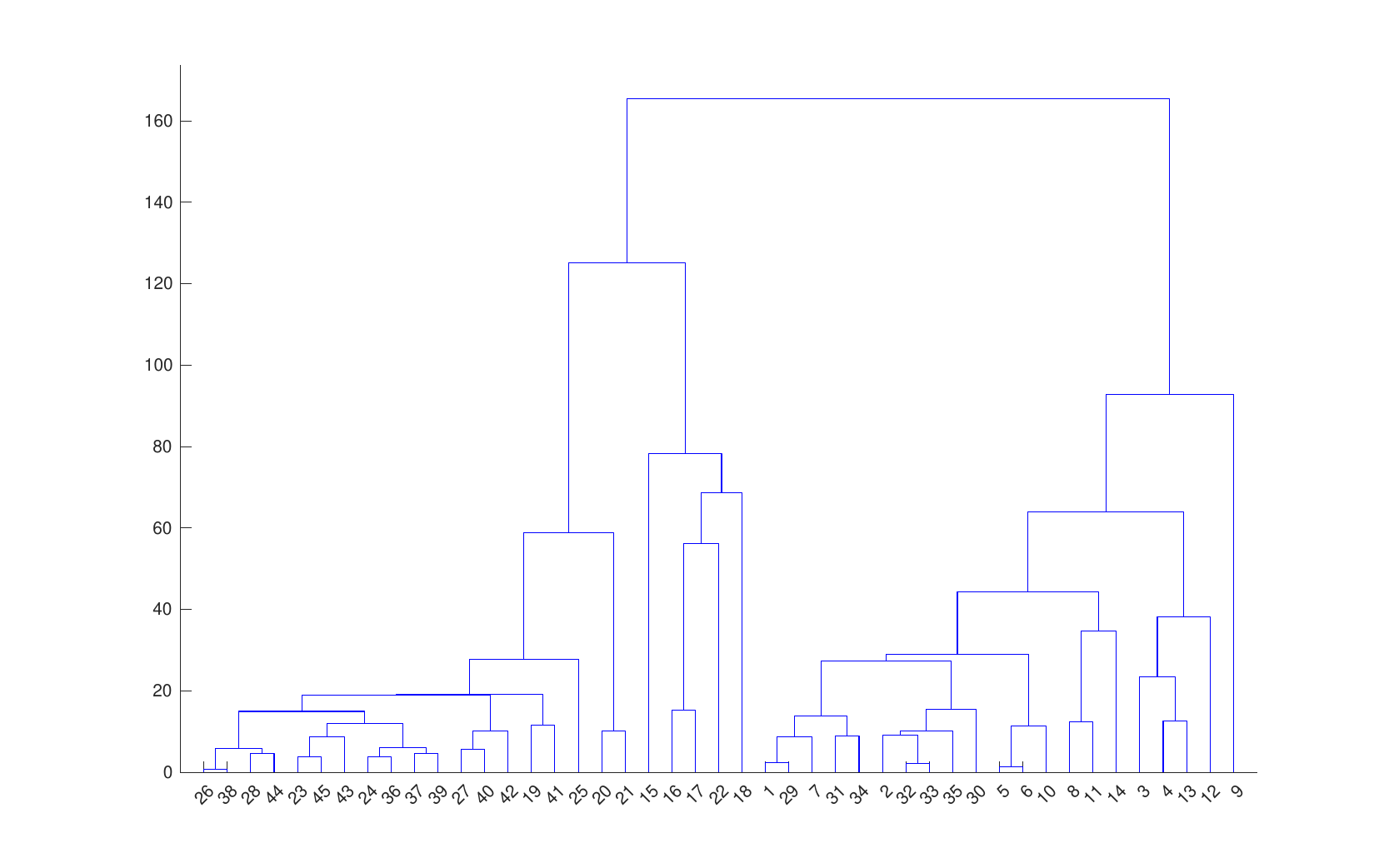}
	\caption{Set 2, Wasserstein distance for the top 1/6 of the curve, with weights according to \ref{waga_uniform}, for the remainder 5/6 of the curve weights of "0" were assigned.}
	\label{Set2_zero1/6}
\end{figure}

\begin{figure}[H]
	\centering
	\includegraphics[width=0.9\linewidth]{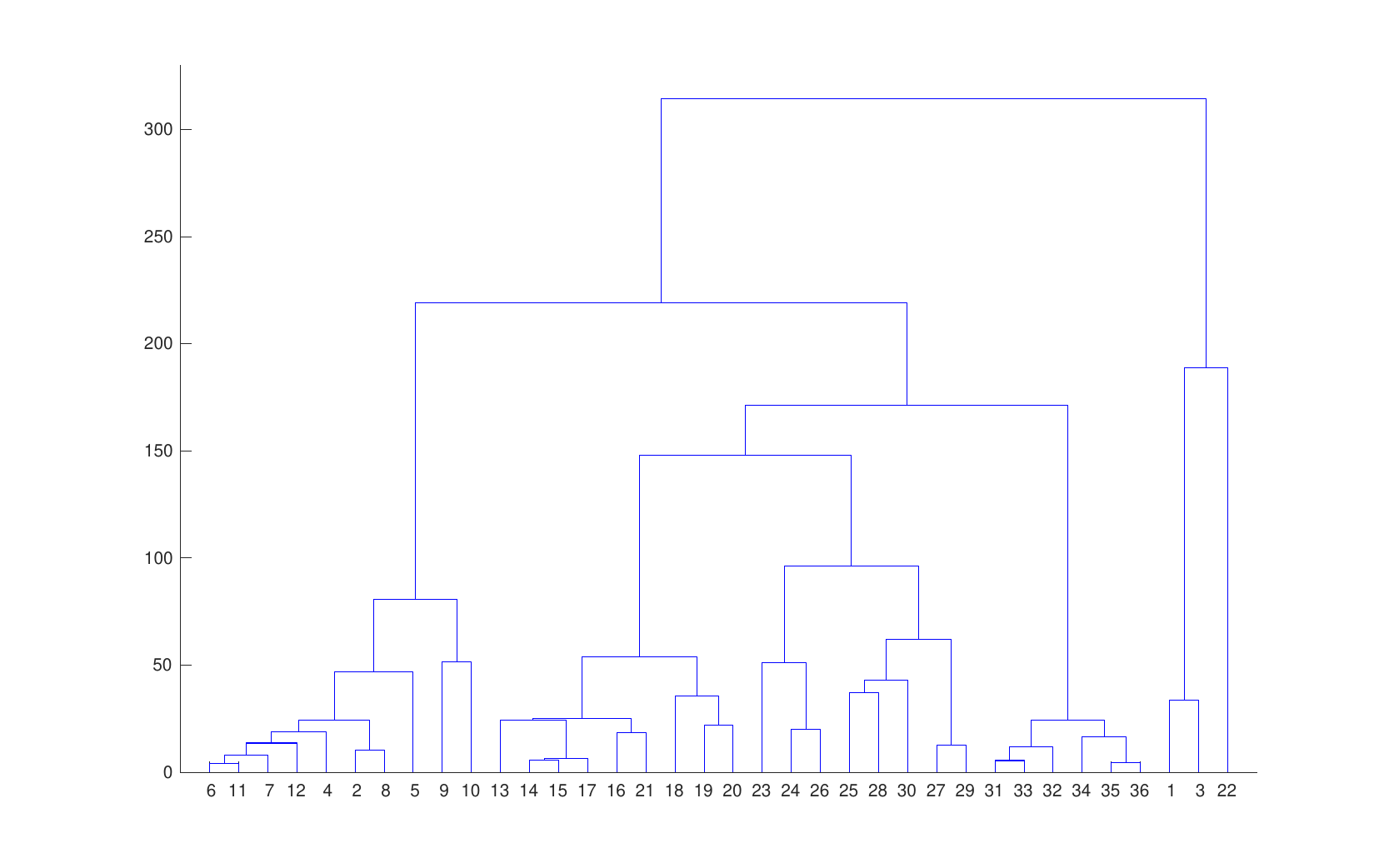}
	\caption{Set 3, Wasserstein distance for the top 1/6 of the curve, with weights according to \ref{waga_uniform}, for the remainder 5/6 of the curve weights of "0" were assigned.}
	\label{Set3_zero1/6}
\end{figure}

\begin{figure}[H]
	\centering
	\includegraphics[width=0.9\linewidth]{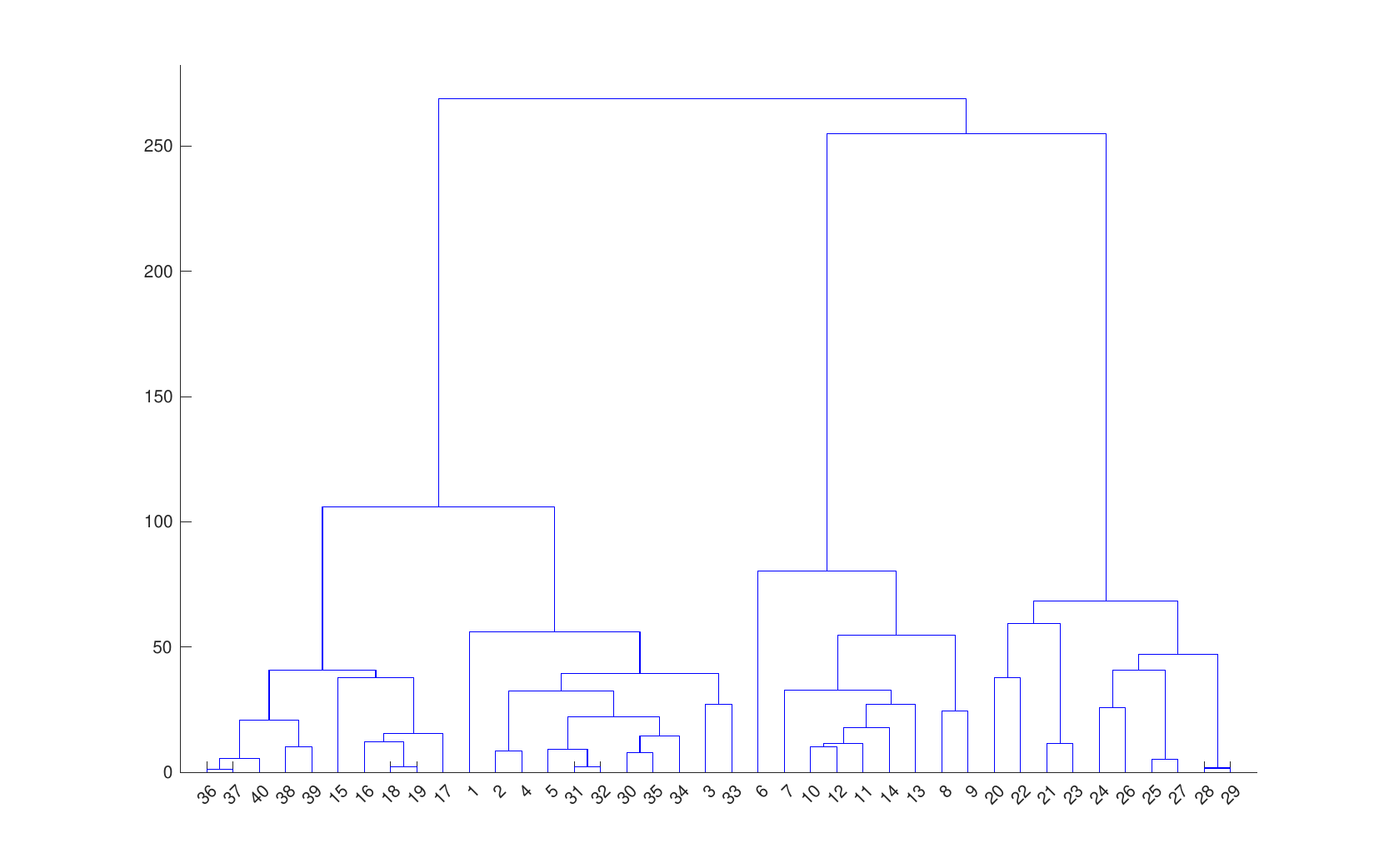}
	\caption{Set 4, Wasserstein distance for the top 1/6 of the curve, with weights according to \ref{waga_uniform}, for the remainder 5/6 of the curve weights of "0" were assigned.}
	\label{Set4_zero1/6}
\end{figure}

\begin{figure}[H]
	\centering
	\includegraphics[width=0.9\linewidth]{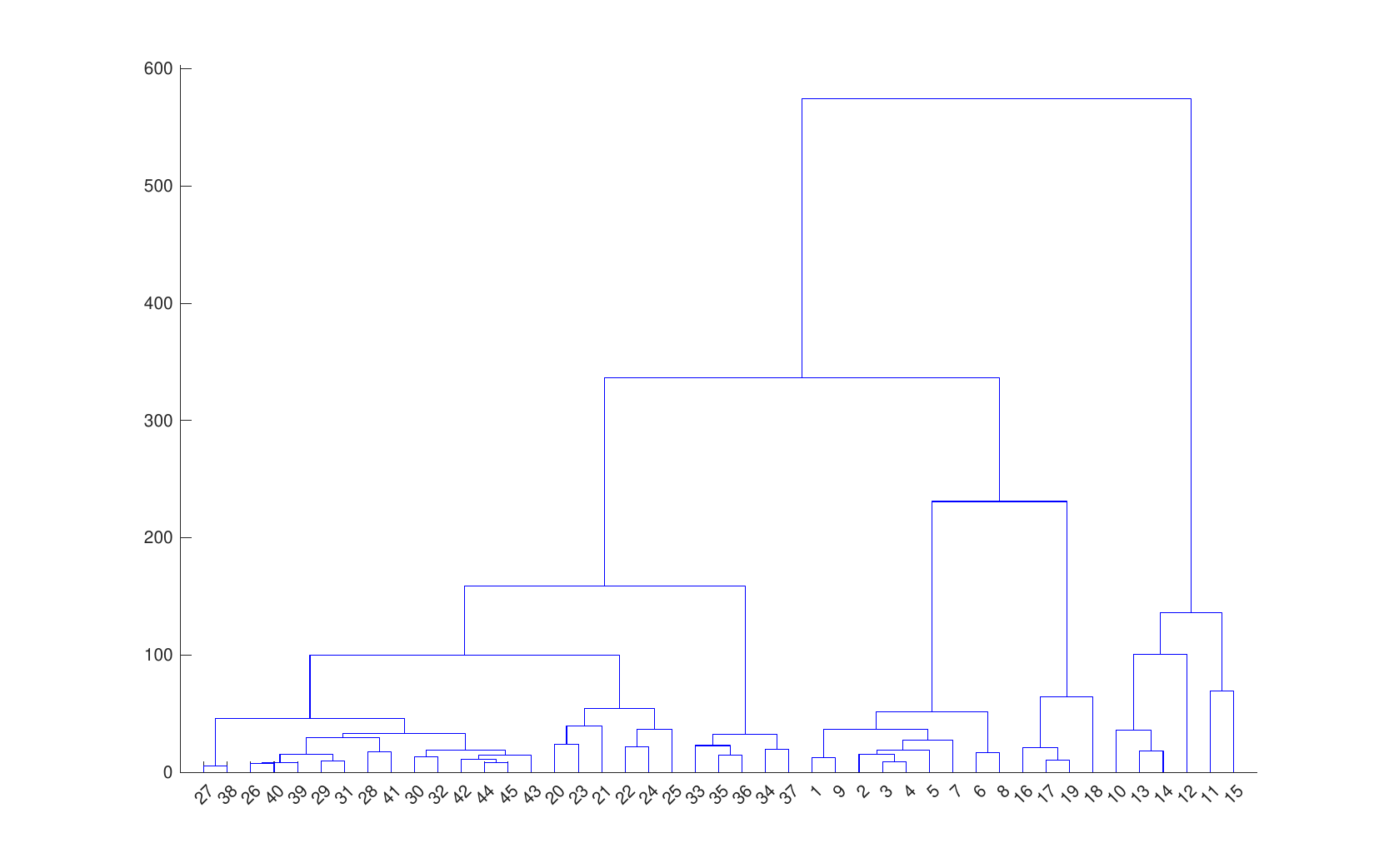}
	\caption{Set 5, Wasserstein distance for the top 1/6 of the curve, with weights according to \ref{waga_uniform}, for the remainder 5/6 of the curve weights of "0" were assigned.}
	\label{Set5_zero1/6}
\end{figure}

\begin{figure}[H]
	\centering
	\includegraphics[width=0.9\linewidth]{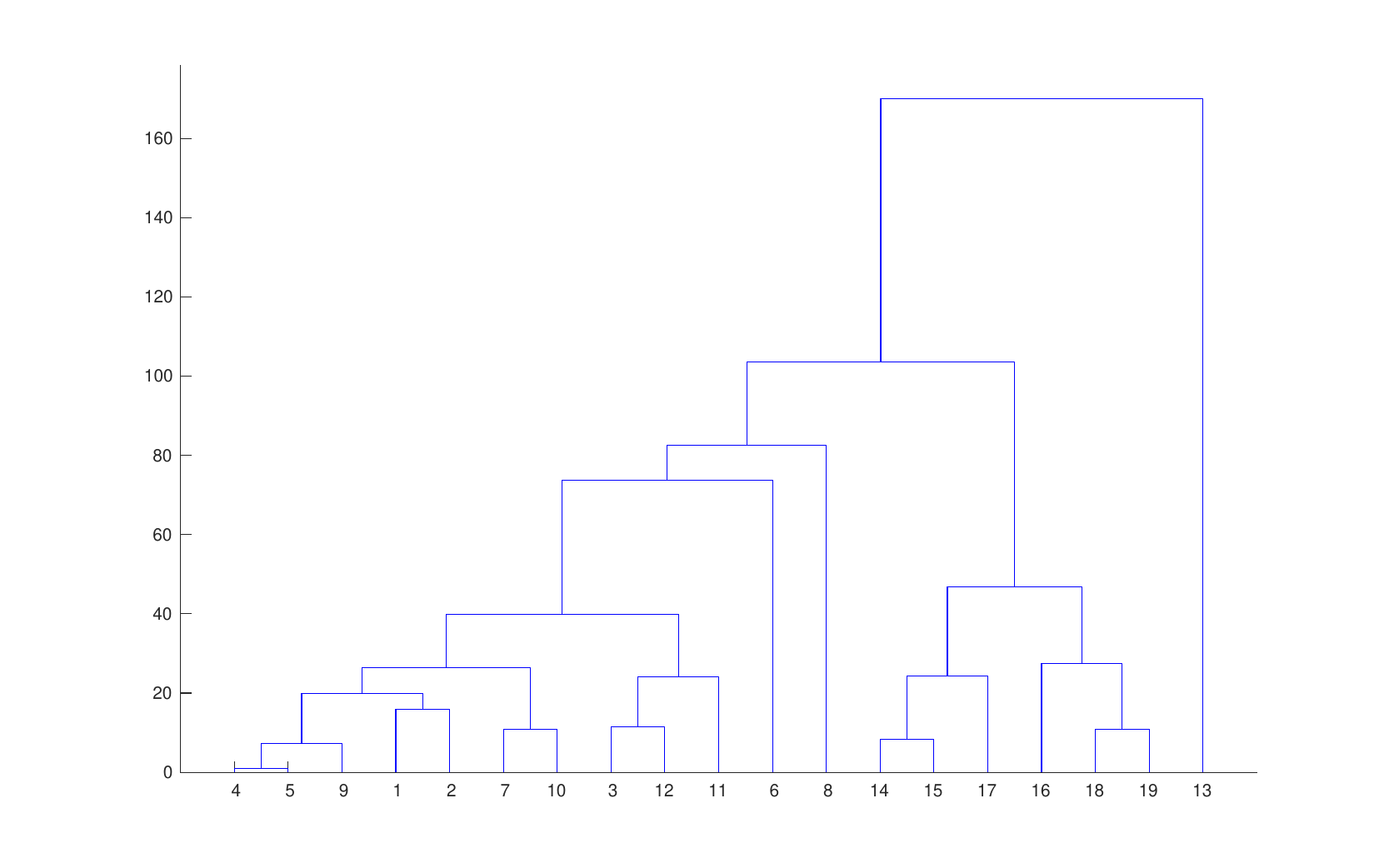}
	\caption{Set 1, Wasserstein distance calculated for the whole curve, weights assigned according to Formula \ref{waga_po_y_dlak}.}
	\label{Set1_normal_distribution_24}
\end{figure}

\begin{figure}[H]
	\centering
	\includegraphics[width=0.9\linewidth]{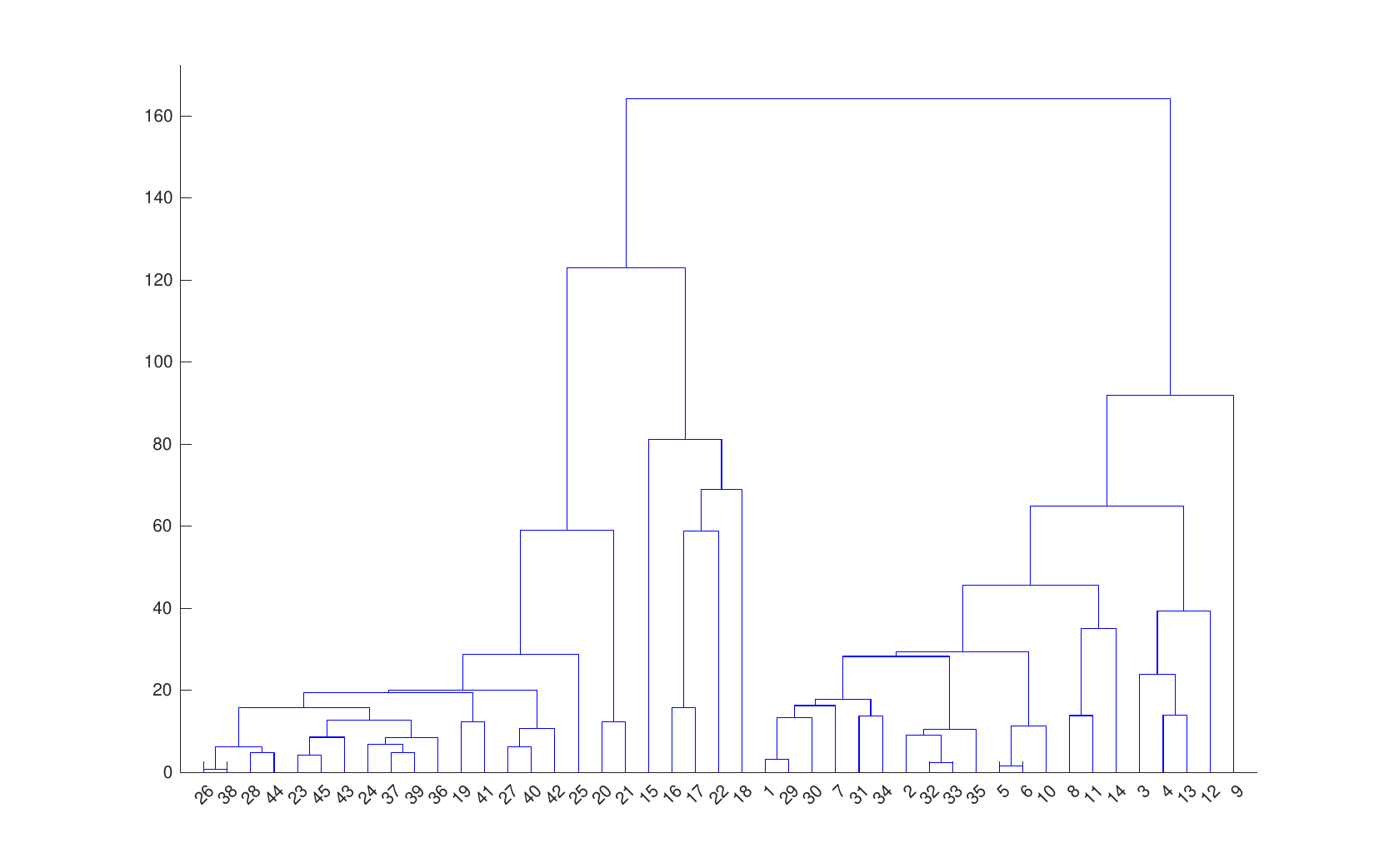}
	\caption{Set 2, Wasserstein distance calculated for the whole curve, weights assigned according to Formula \ref{waga_po_y_dlak}.}
	\label{Set2_normal_distribution_24}
\end{figure}

\begin{figure}[H]
	\centering
	\includegraphics[width=0.9\linewidth]{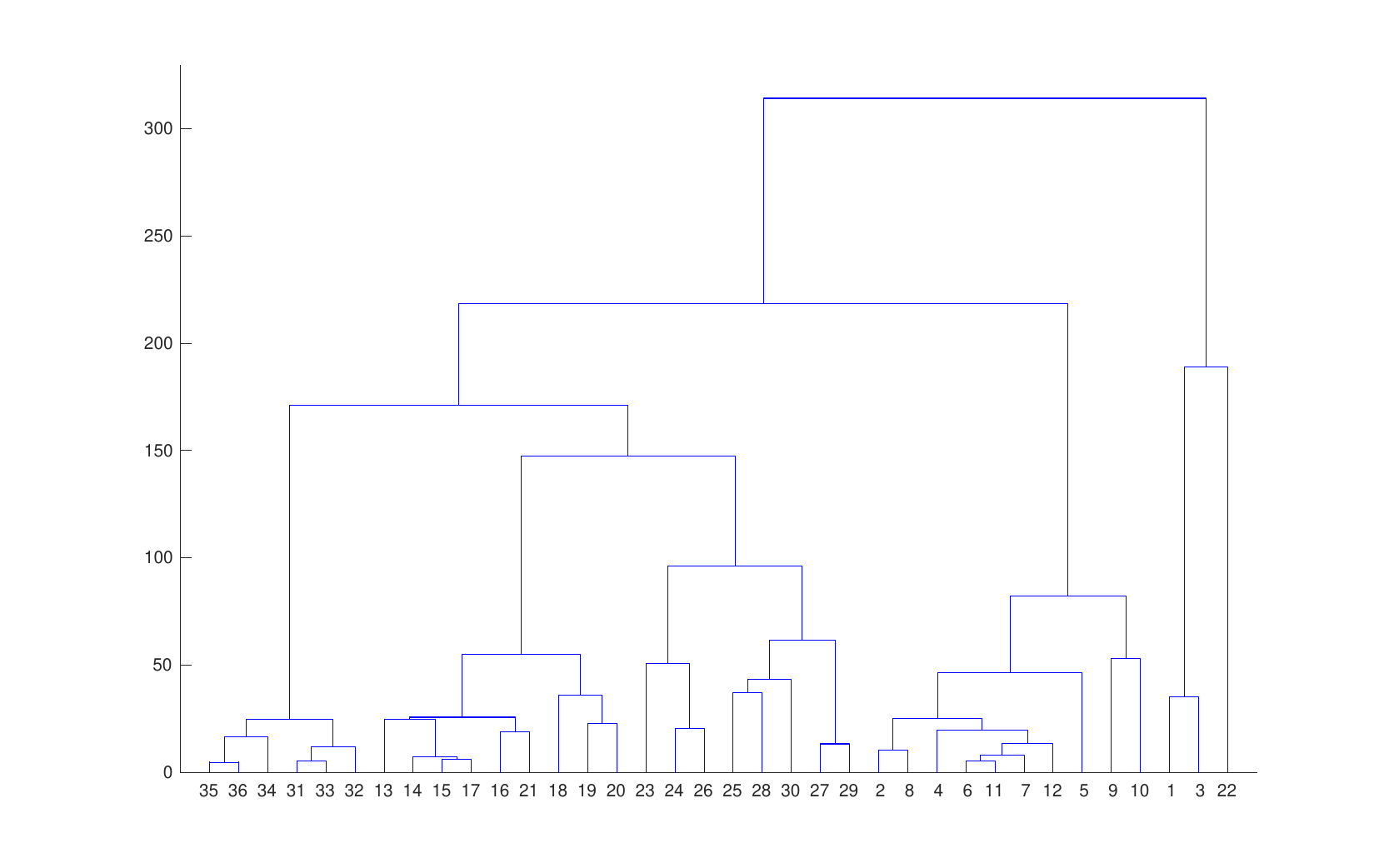}
	\caption{Set 3, Wasserstein distance calculated for the whole curve, weights assigned according to Formula \ref{waga_po_y_dlak}.}
	\label{Set3_normal_distribution_24}
\end{figure}

\begin{figure}[H]
	\centering
	\includegraphics[width=0.9\linewidth]{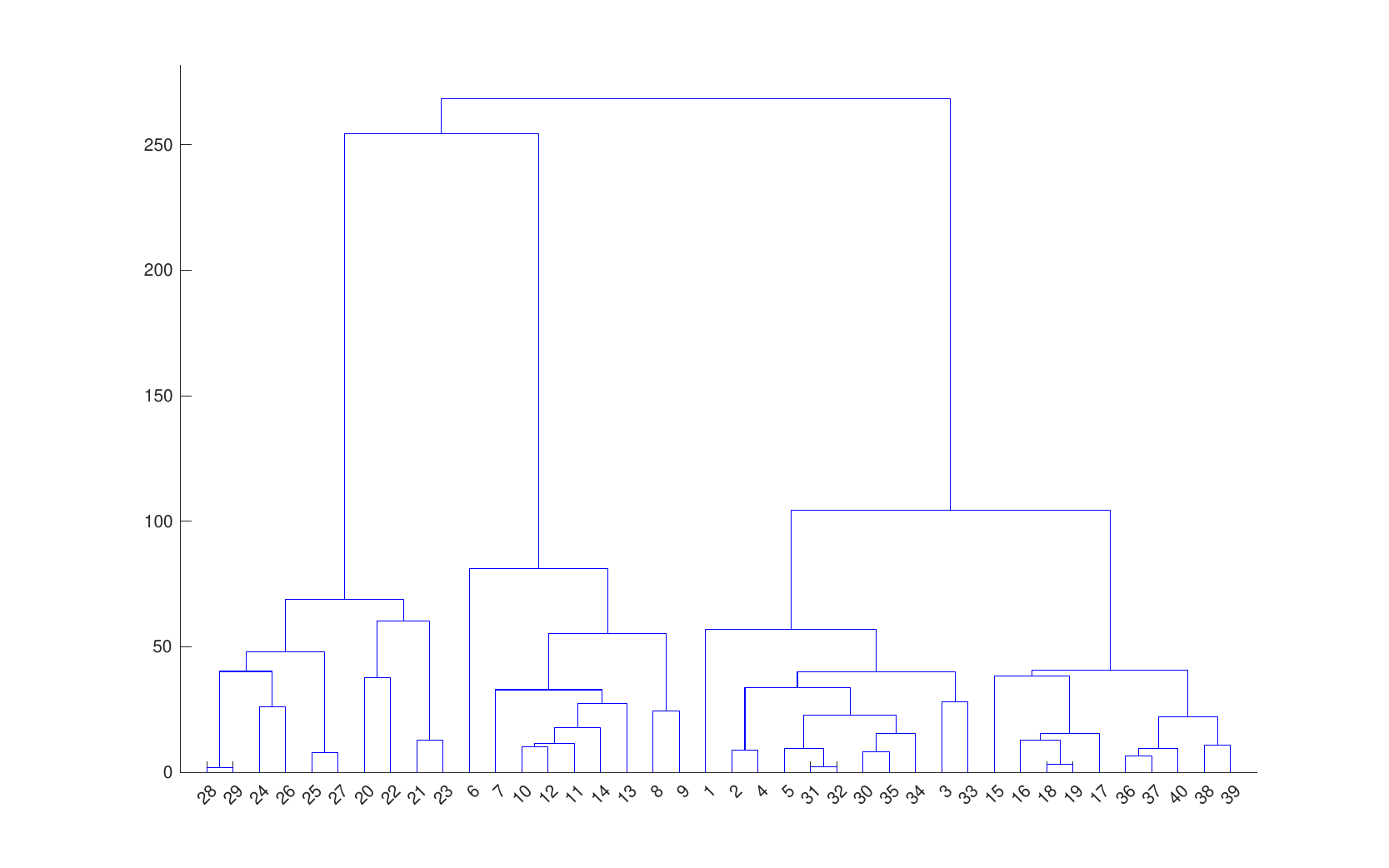}
	\caption{Set 4, Wasserstein distance calculated for the whole curve, weights assigned according to Formula \ref{waga_po_y_dlak}.}
	\label{Set4_normal_distribution_24}
\end{figure}

\begin{figure}[H]
	\centering
	\includegraphics[width=0.9\linewidth]{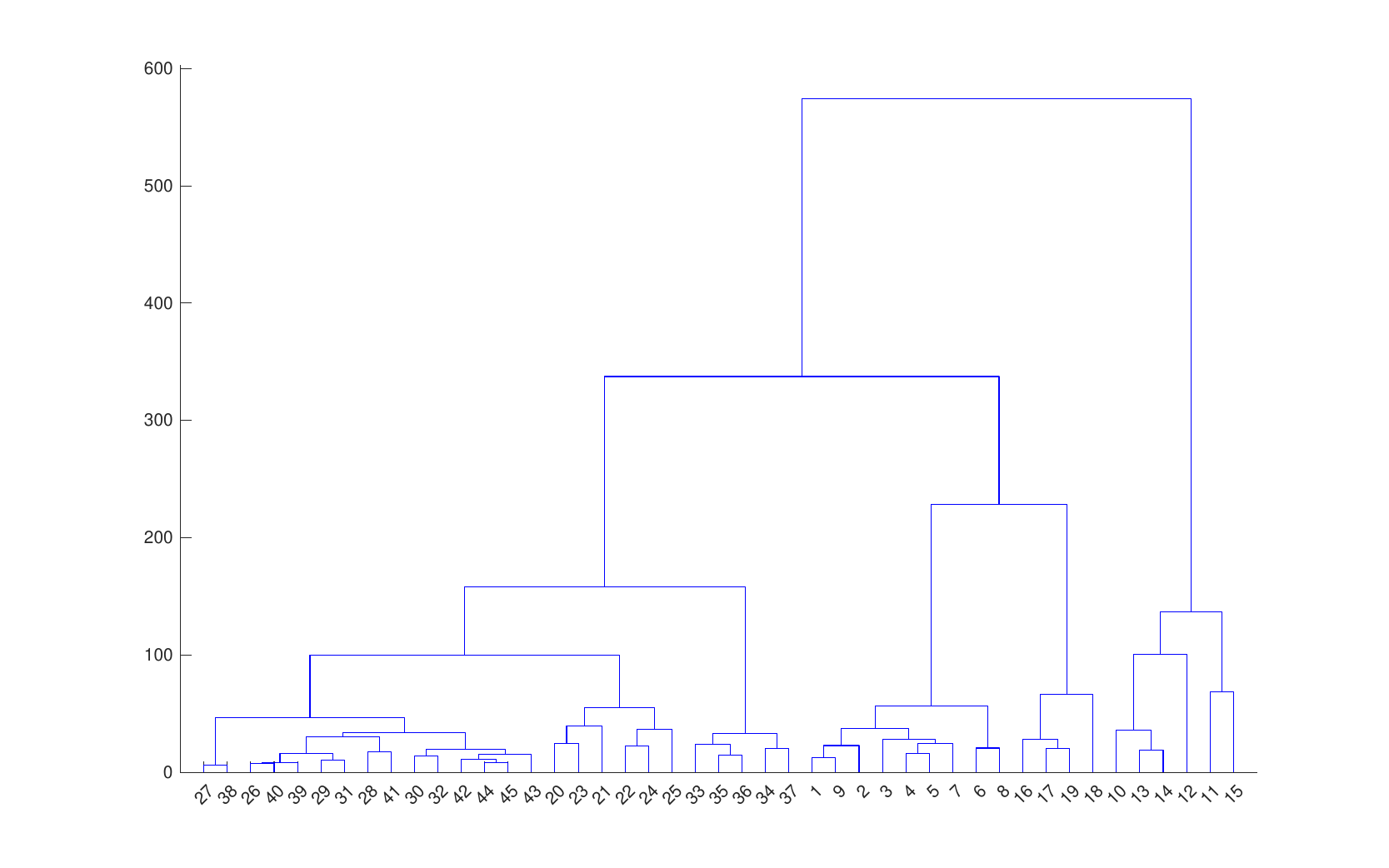}
	\caption{Set 5, Wasserstein distance calculated for the whole curve, weights assigned according to Formula \ref{waga_po_y_dlak}.}
	\label{Set5_normal_distribution_24}
\end{figure}

 In Experiment 3 we have applied uniform weights of $1/n$ to the top 1/6 of the curve, the weights for the remainder of the curve were assigned as "0". The results are presented in Figures \ref{Set1_zero1/6} -\ref{Set5_zero1/6}. This was done to contrast the results with the results of Experiment 2, where the curve was automatically cut to the section of interest, in our case the top 1/6. To this section uniform weight of $1/n$ were assigned (Formula \ref{waga_uniform}). The results are presented in Figures \ref{Set1_1_6}-\ref{Set5_1_6}. A somewhat similar result is obtained in Experiment 8, using Formula \ref{waga_po_y_dlak}. This measure allows to apply weights to a prescribed section of the curve  $k_1, k_2$ and $l_1, l_2$, simultaneously distributing the weights on the selected fragment as to focus the classification on the initial pints of the curve (the furthest the point the smaller the weight). The results are presented in Figures \ref{Set1_normal_distribution_24}-\ref{Set5_normal_distribution_24}.
 
\begin{figure}[H]
	\centering
	\includegraphics[width=0.9\linewidth]{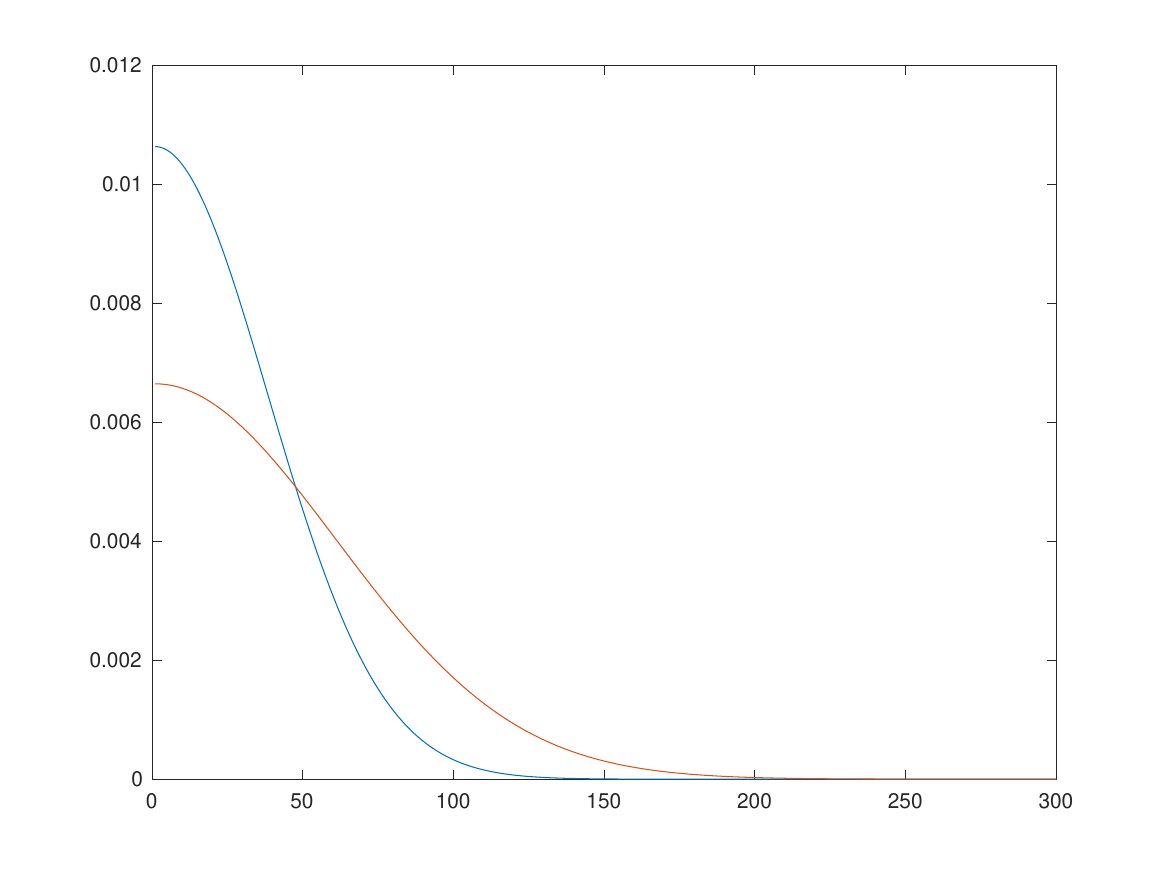}
	\caption{The binomial distribution curve, according to which weights are assigned (Formula \ref{waga_uniform}). The blue curve (a) represents the a focus on the first $1/6$ of the curve with a sharp drop of interest,  the red curve (b) represents the a focus on the first $1/3$ of the curve with a soft drop of interest.}
	\label{fig_normal_distributions}
\end{figure}

\begin{figure}[H]
	\centering
	\includegraphics[width=0.9\linewidth]{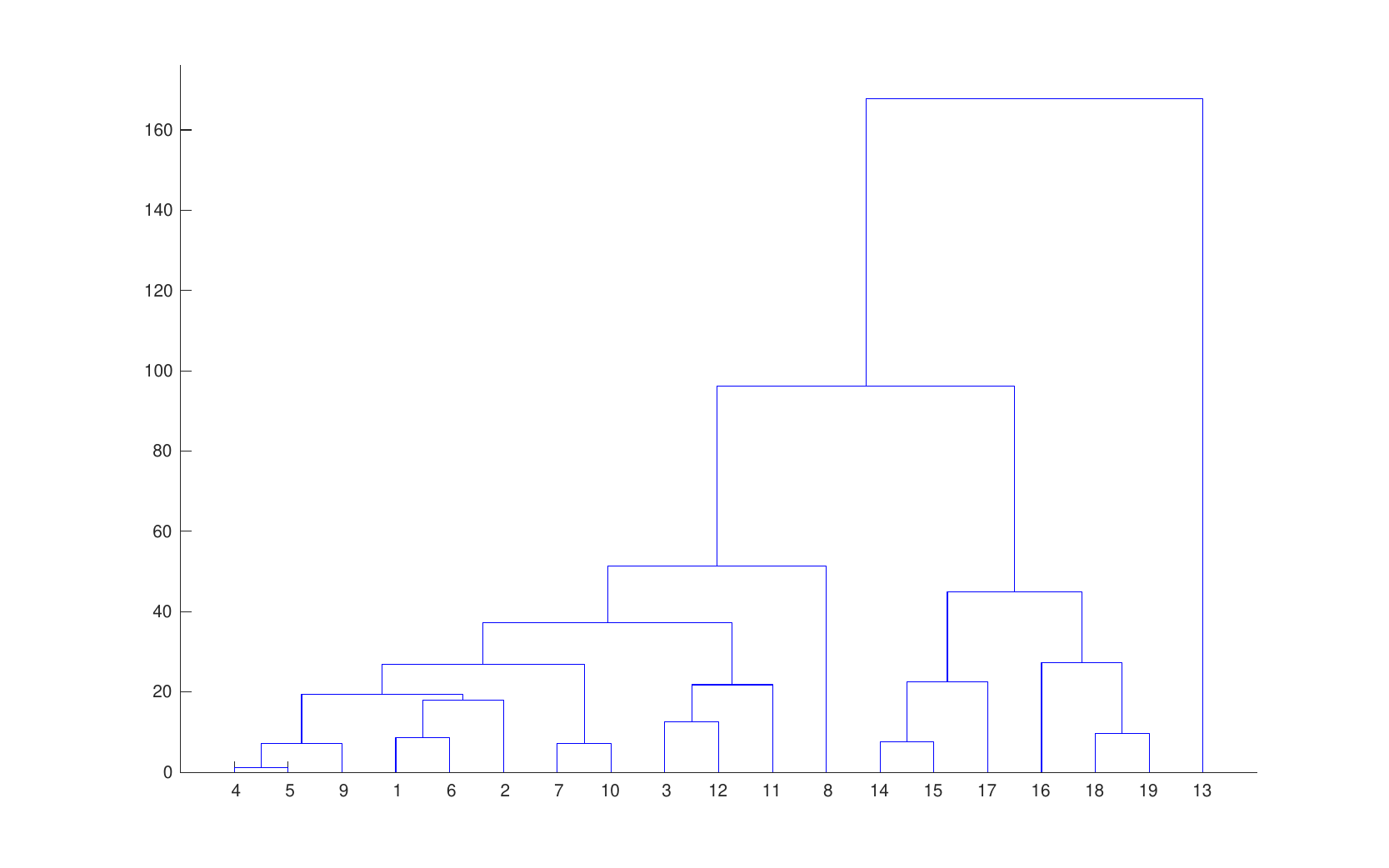}
	\caption{Set 1, Wasserstein distance calculated for the whole curve,weights assigned according to the curve (a) in Figure \ref{fig_normal_distributions}.}
	\label{Set1_normal1/6_ostra}
\end{figure}

\begin{figure}[H]
	\centering
	\includegraphics[width=0.9\linewidth]{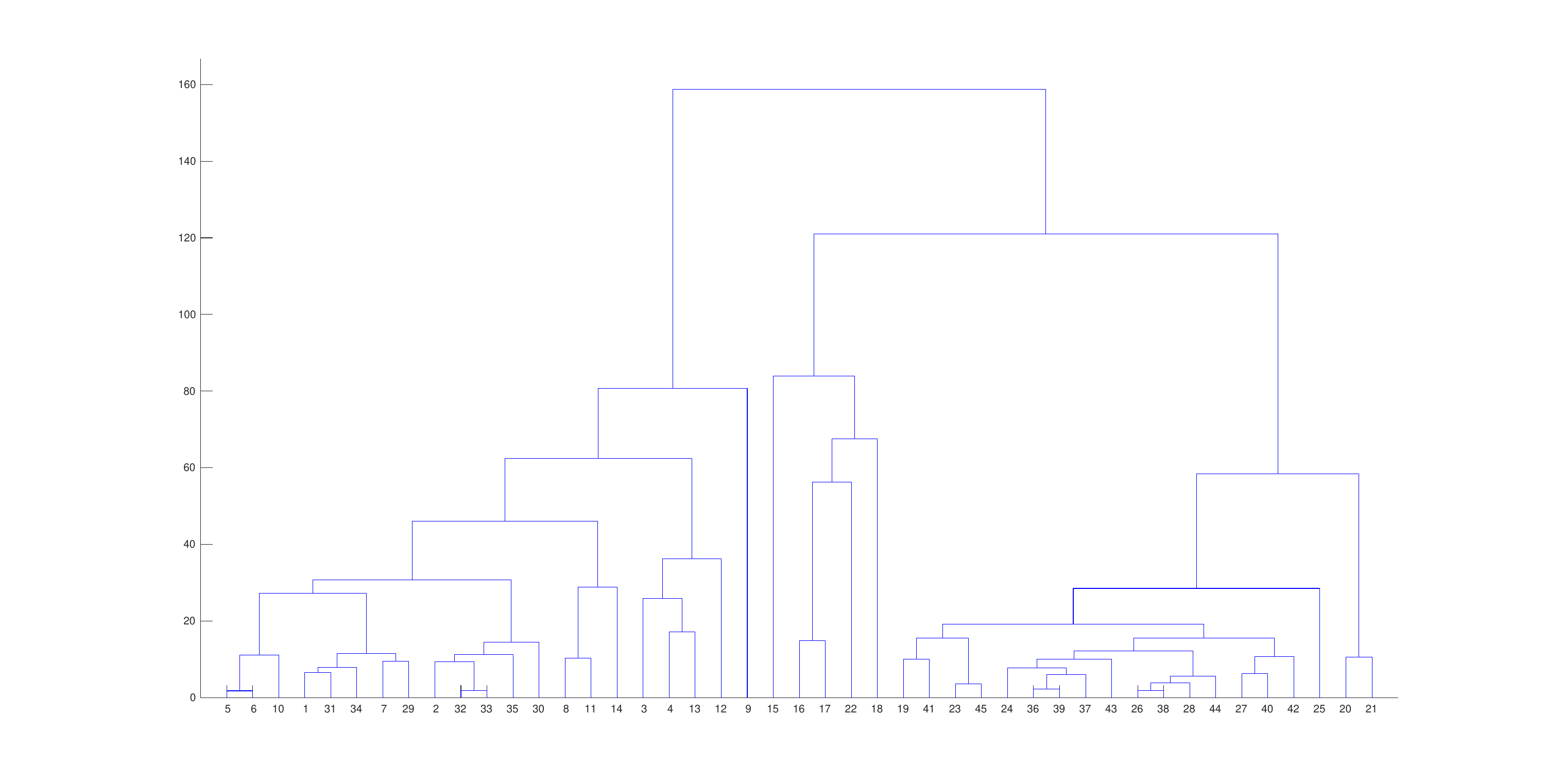}
	\caption{Set 2, Wasserstein distance calculated for the whole curve,weights assigned according to the curve (a) in Figure \ref{fig_normal_distributions}.}
	\label{Set2_normal1/6_ostra}
\end{figure}

\begin{figure}[H]
	\centering
	\includegraphics[width=0.9\linewidth]{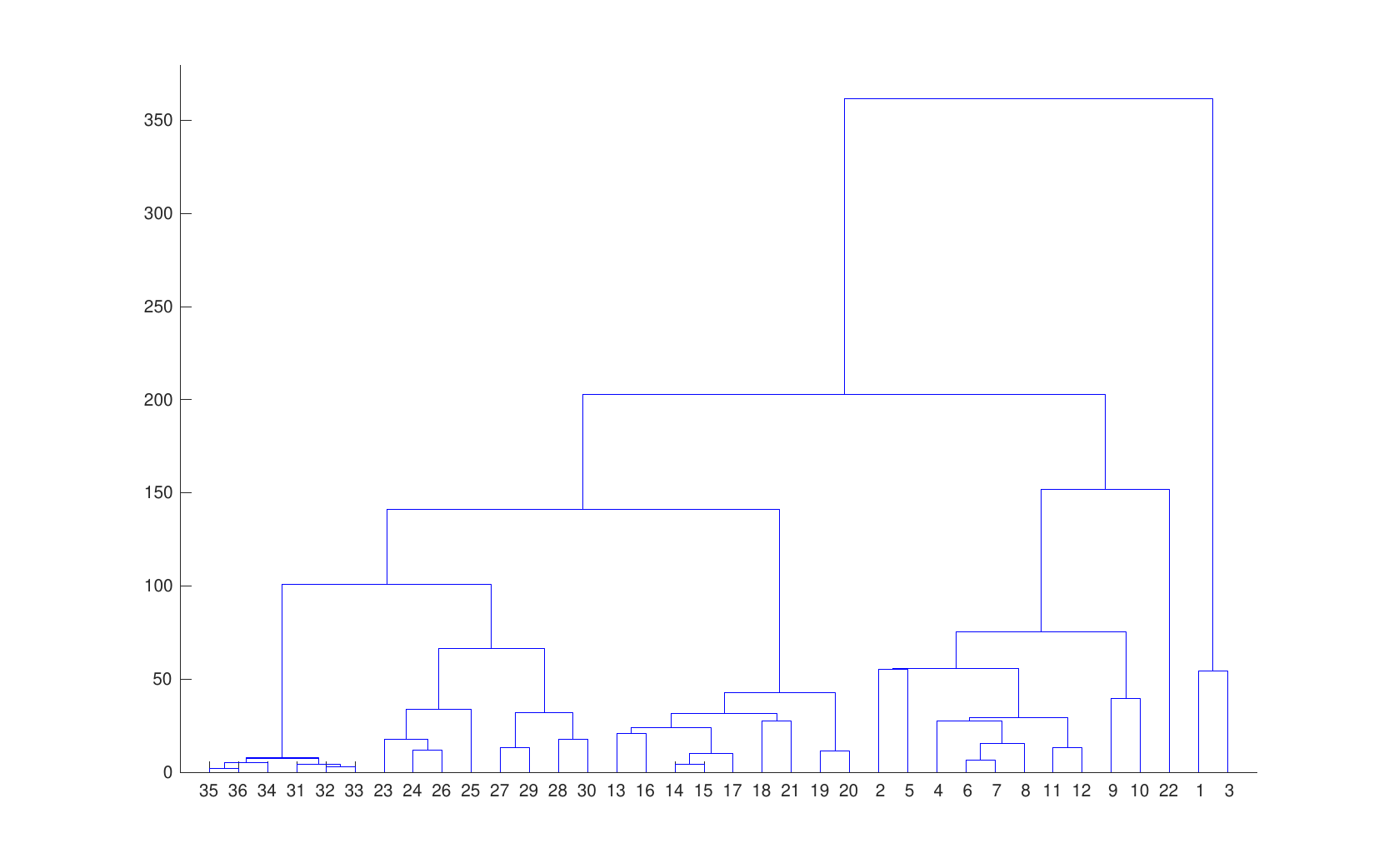}
	\caption{Set 3,  Wasserstein distance calculated for the whole curve,weights assigned according to the curve (a) in Figure \ref{fig_normal_distributions}.}
	\label{Set3_normal1/6_ostra}
\end{figure}

\begin{figure}[H]
	\centering
	\includegraphics[width=0.9\linewidth]{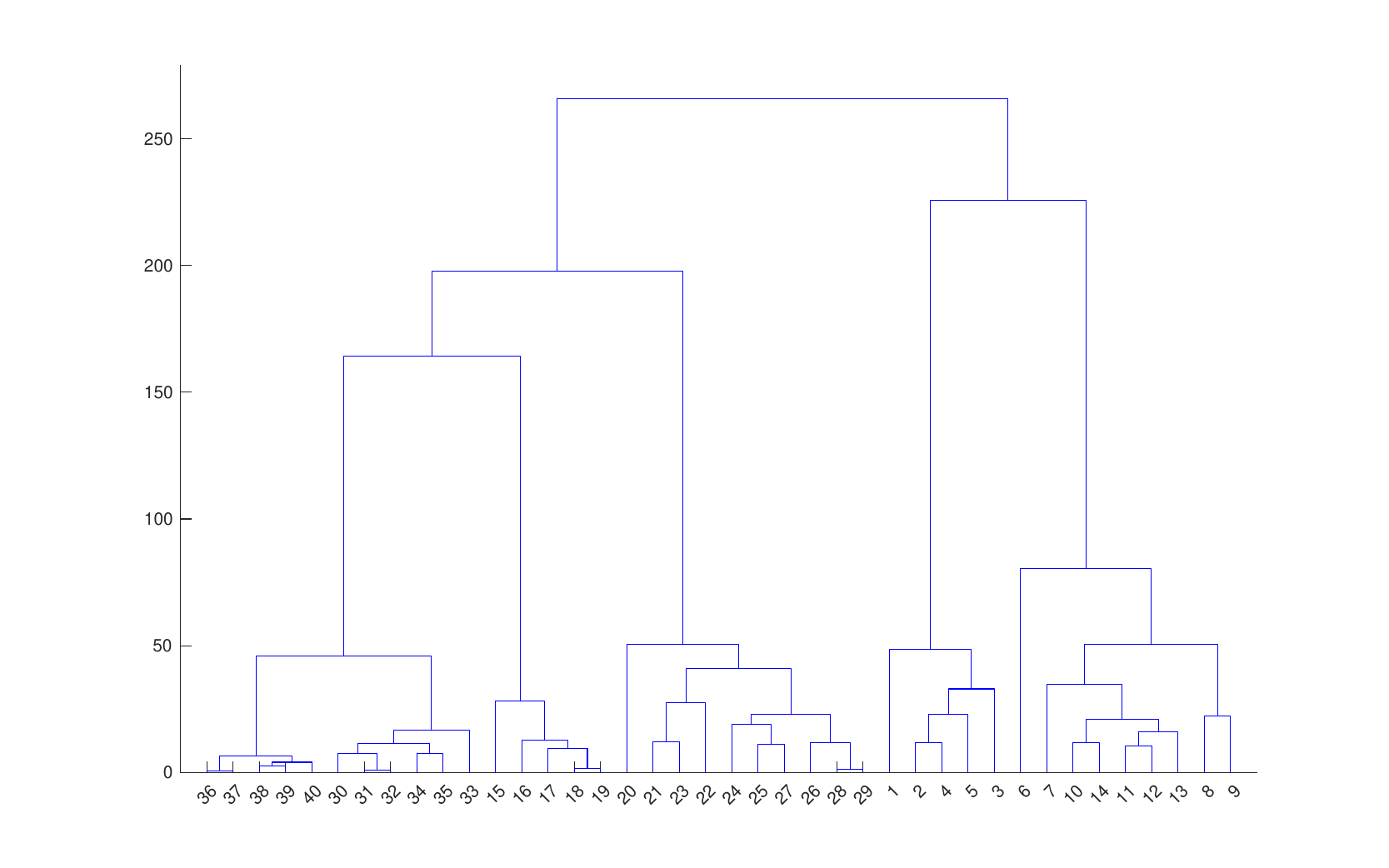}
	\caption{Set 4,  Wasserstein distance calculated for the whole curve,weights assigned according to the curve (a) in Figure \ref{fig_normal_distributions}.}
	\label{Set4_normal1/6_ostra}
\end{figure}

\begin{figure}[H]
	\centering
	\includegraphics[width=0.9\linewidth]{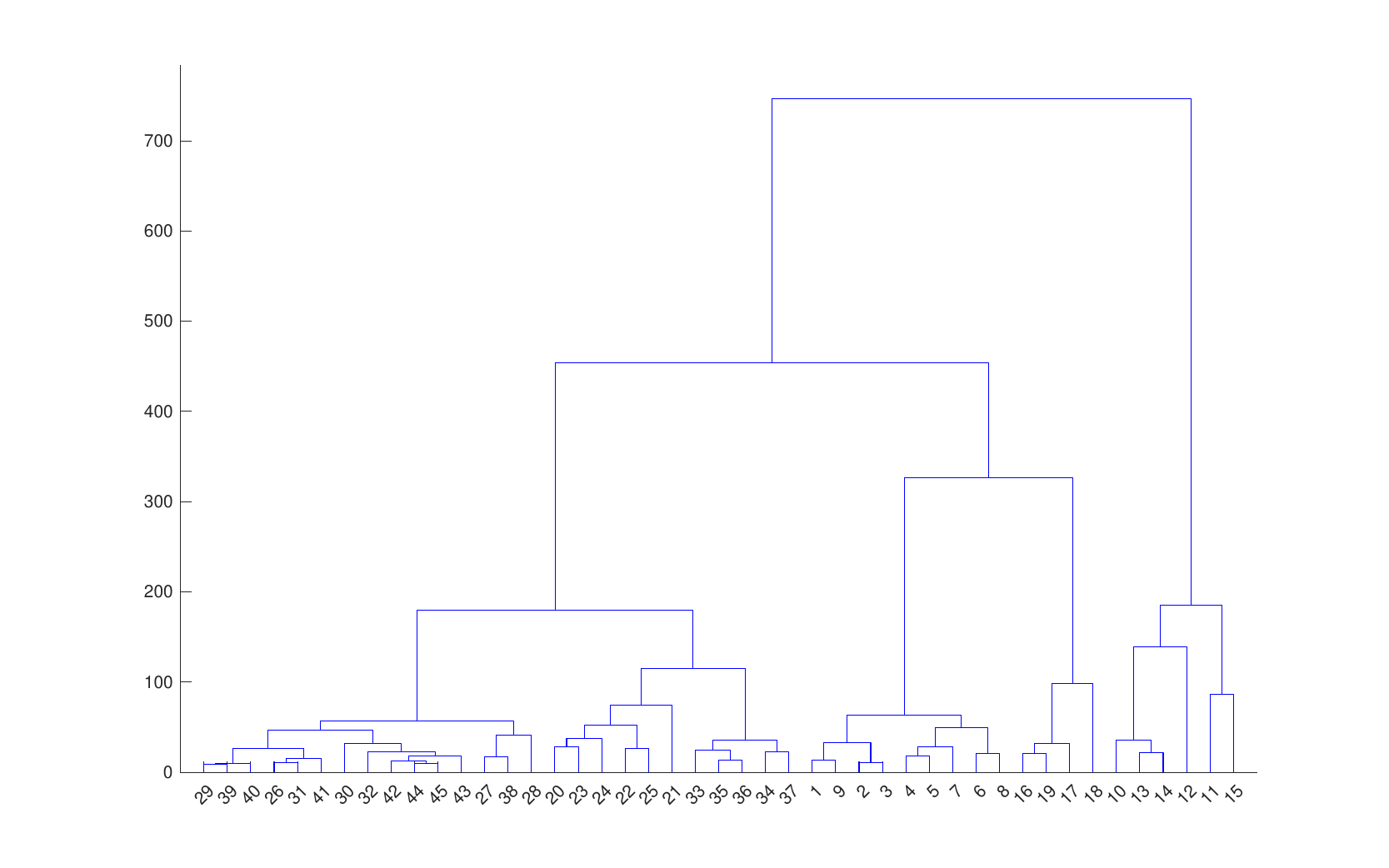}
	\caption{Set 5,  Wasserstein distance calculated for the whole curve,weights assigned according to the curve (a) in Figure \ref{fig_normal_distributions}.}
	\label{Set5_normal1/6_ostra}
\end{figure}

\begin{figure}[H]
	\centering
	\includegraphics[width=0.9\linewidth]{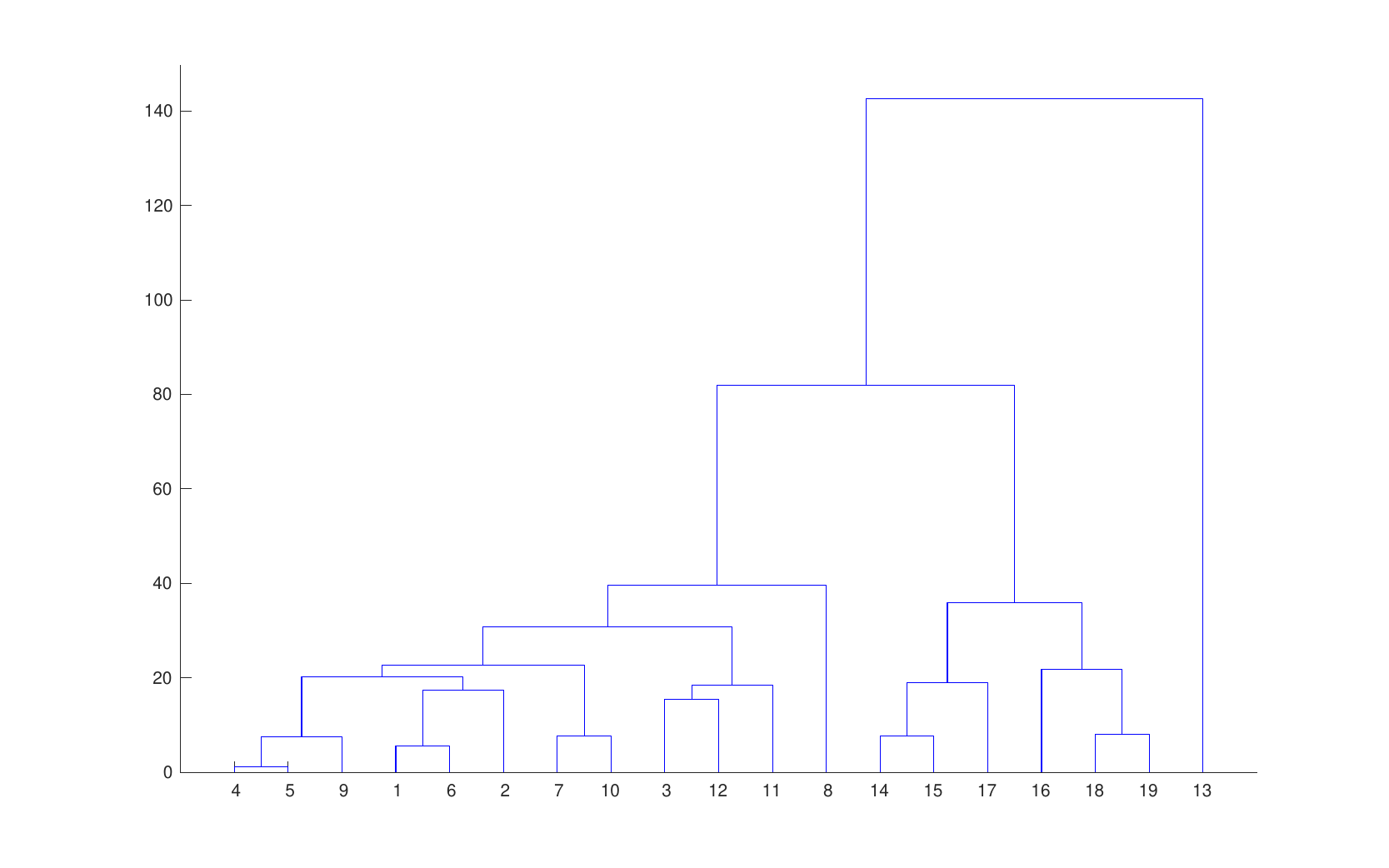}
	\caption{Set 1, Wasserstein distance calculated for the whole curve,weights assigned according to the curve (b) in Figure \ref{fig_normal_distributions}.}
	\label{Set1_normal1/6_lagodna}
\end{figure}

\begin{figure}[H]
	\centering
	\includegraphics[width=0.9\linewidth]{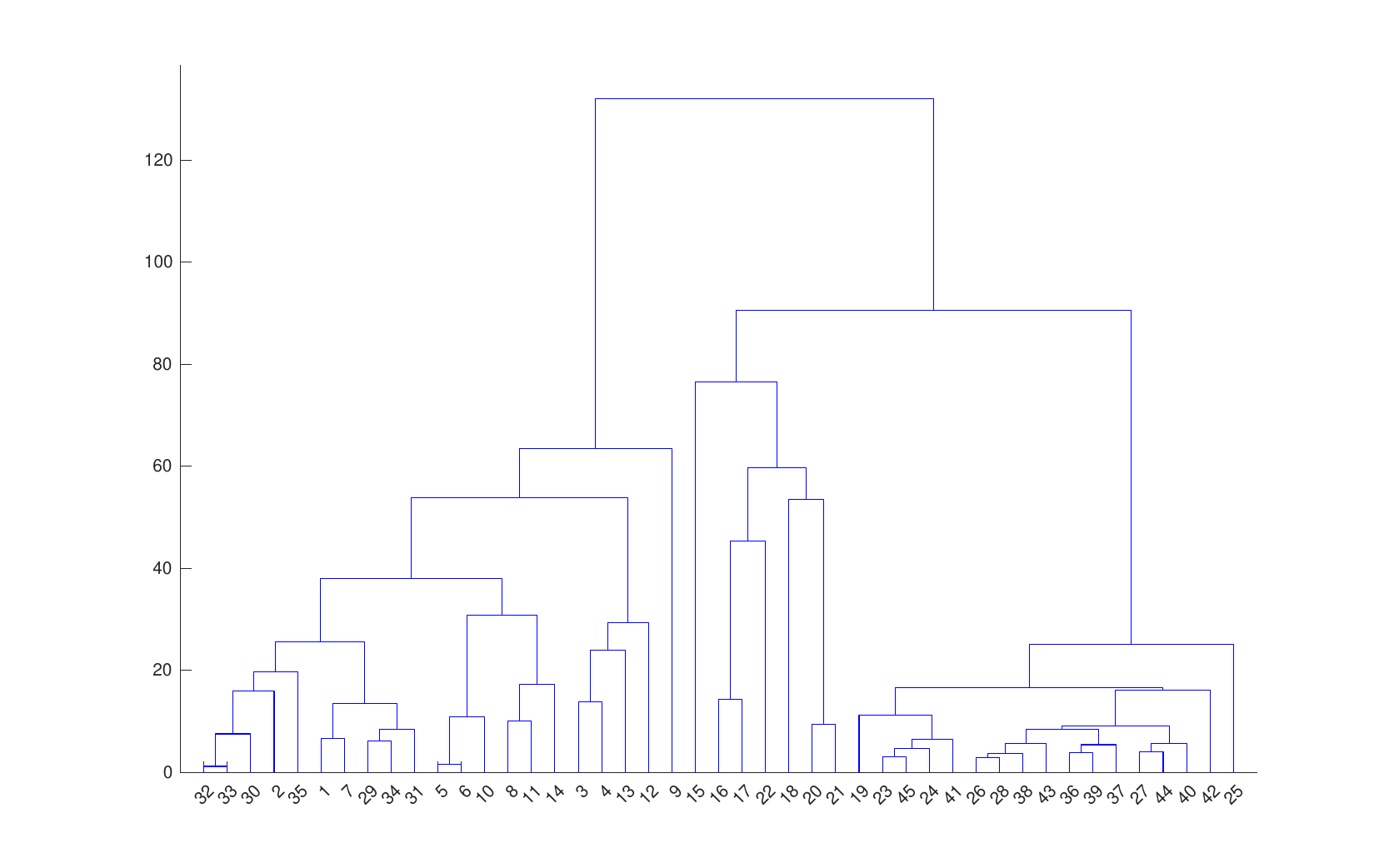}
	\caption{Set 2, Wasserstein distance calculated for the whole curve,weights assigned according to the curve (b) in Figure \ref{fig_normal_distributions}.}
	\label{Set2_normal1/6_lagodna}
\end{figure}

\begin{figure}[H]
	\centering
	\includegraphics[width=0.9\linewidth]{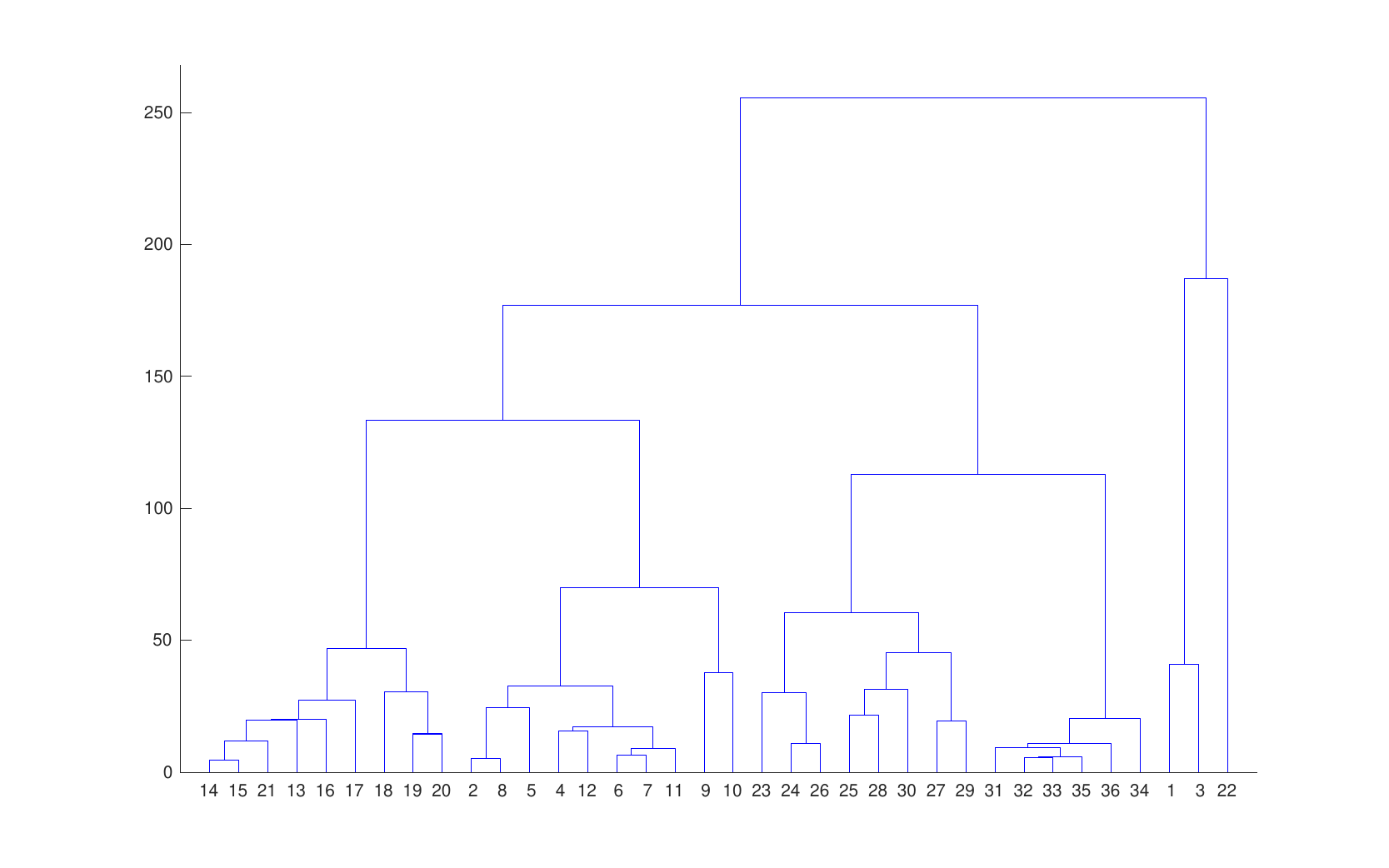}
	\caption{Set 3, Wasserstein distance calculated for the whole curve,weights assigned according to the curve (b) in Figure \ref{fig_normal_distributions}.}
	\label{Set3_normal1/6_lagodna}
\end{figure}

\begin{figure}[H]
	\centering
	\includegraphics[width=0.9\linewidth]{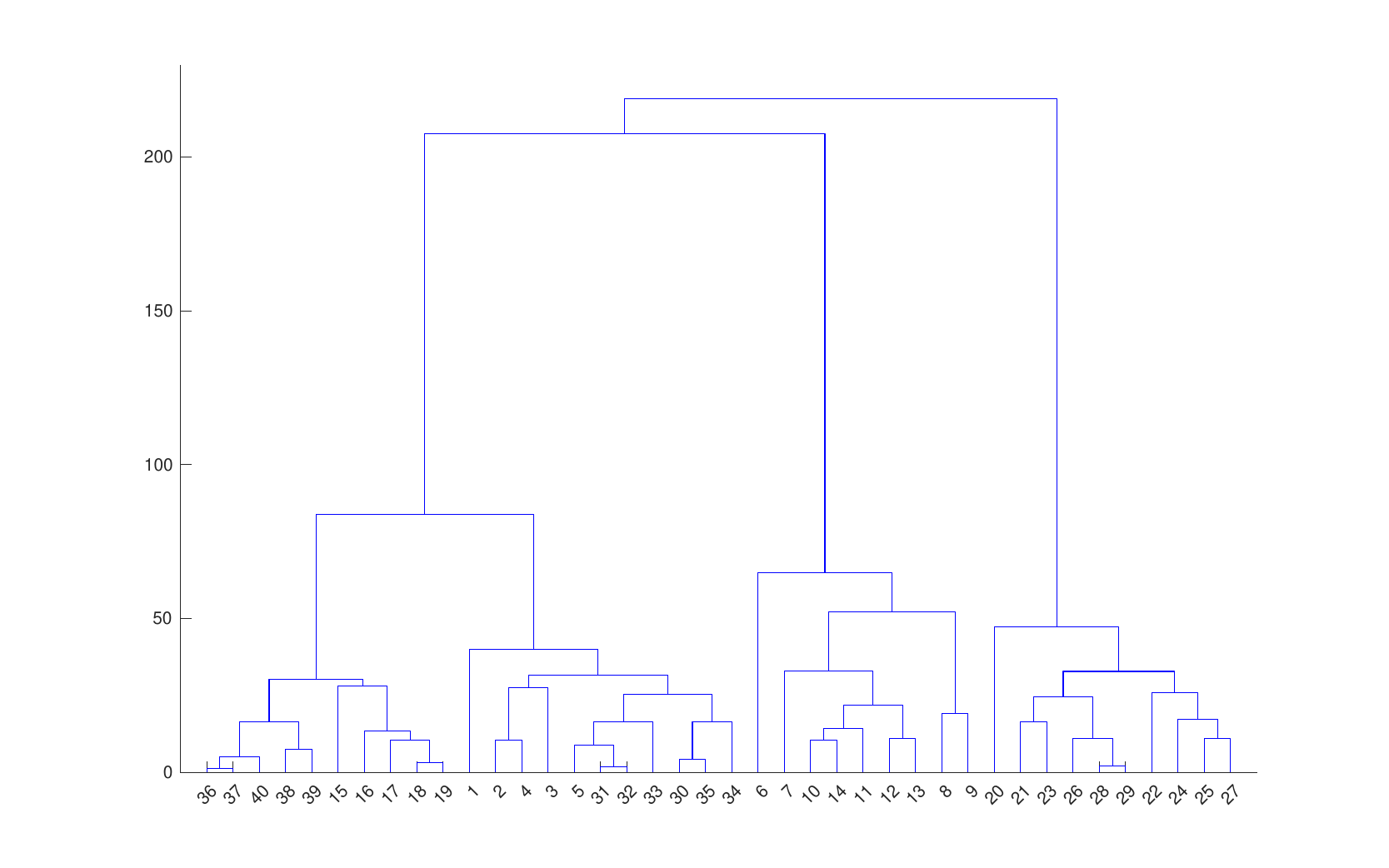}
	\caption{Set 4, Wasserstein distance calculated for the whole curve,weights assigned according to the curve (b) in Figure \ref{fig_normal_distributions}.}
	\label{Set4_normal1/6_lagodna}
\end{figure}

\begin{figure}[H]
	\centering
	\includegraphics[width=0.9\linewidth]{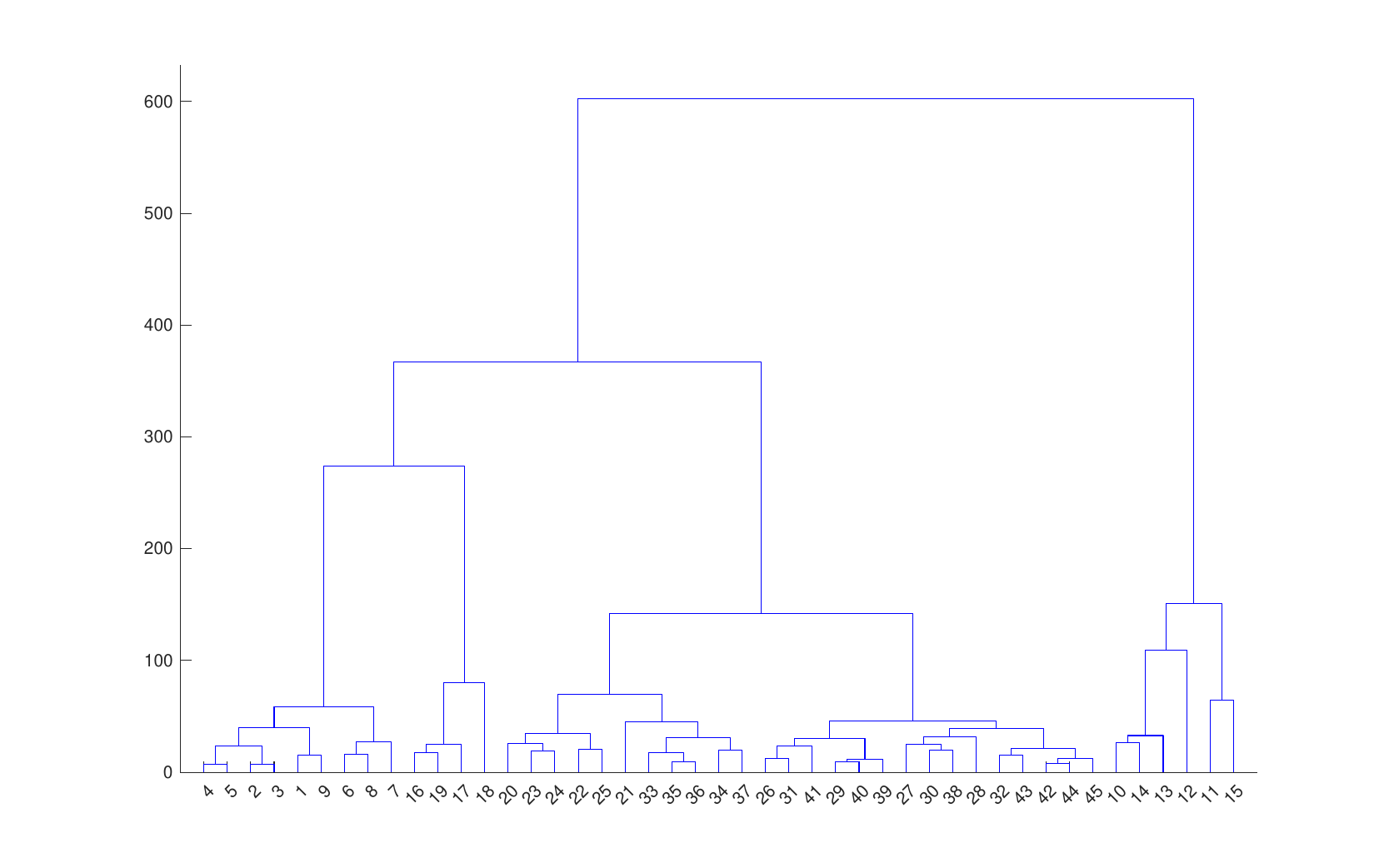}
	\caption{Set 5, Wasserstein distance calculated for the whole curve, weights assigned according to the curve (b) in Figure \ref{fig_normal_distributions}.}
	\label{Set5_normal1/6_lagodna}
\end{figure}

 For Experiments 4 and 5 we have assigned weights according to the binomial distribution. We take advantage of the specific formulation of the Wasserstein distance \ref{waga_uniform}. Here other distributions are also conceivable. We have chosen two distributions with different parameters, as demonstrated in Figure \ref{fig_normal_distributions}. This method allows flexibility in highlighting the section of interest in the curve without the need to establish a sharp cut-off point. In our opinion this method reflects best the natural way of perceiving shape similarity. The results are presented in Figures \ref{Set1_normal1/6_ostra}-\ref{Set5_normal1/6_lagodna}.
 
\begin{figure}[H]
	\centering
	\includegraphics[width=0.9\linewidth]{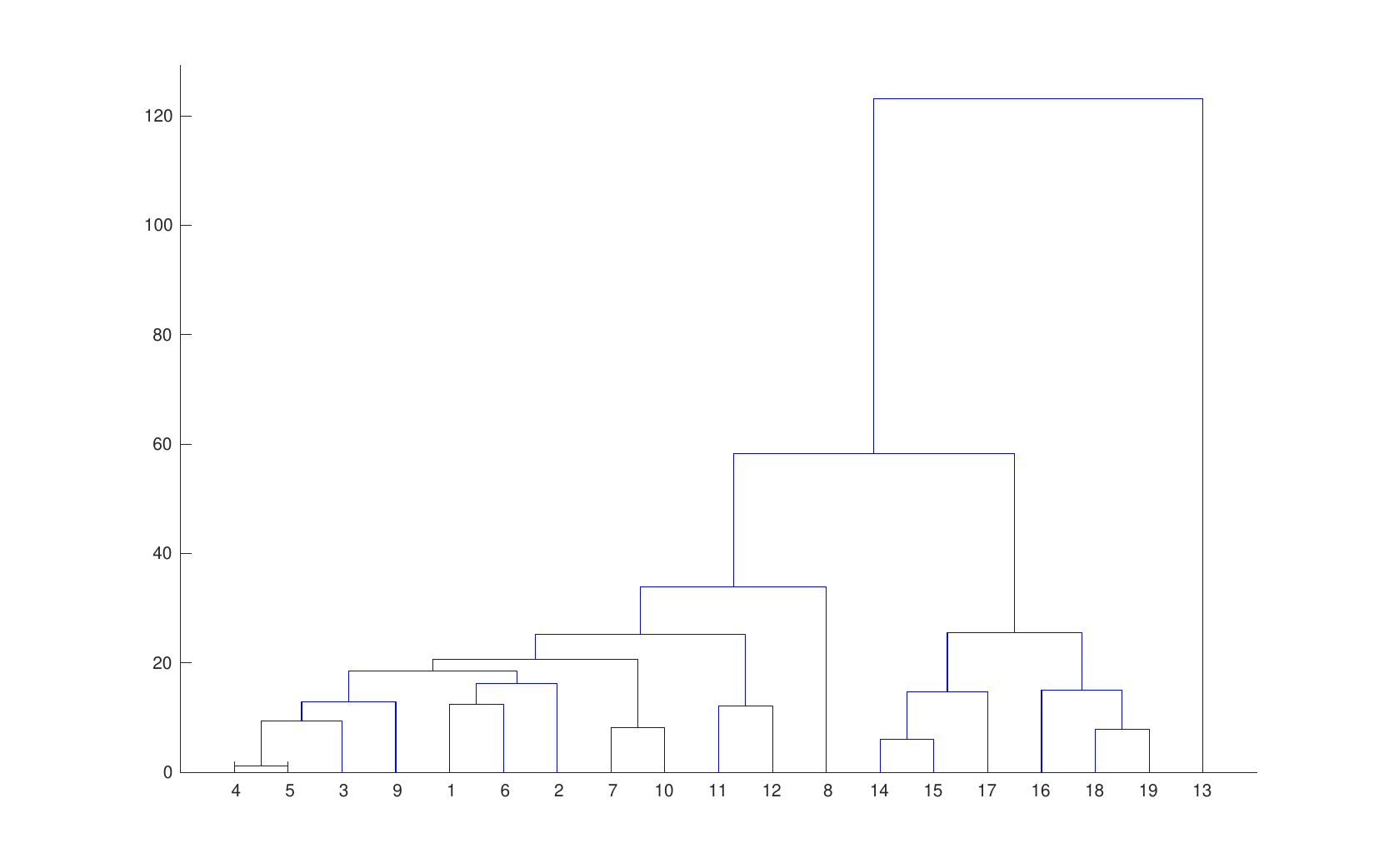}
	\caption{Set 1, Wasserstein distance calculated for the whole curve, weights assigned according to Formula \ref{waga_wspolrzedne_odwrocona}.}
	\label{Set1_normal_distribution_12}
\end{figure}

\begin{figure}[H]
	\centering
	\includegraphics[width=0.9\linewidth]{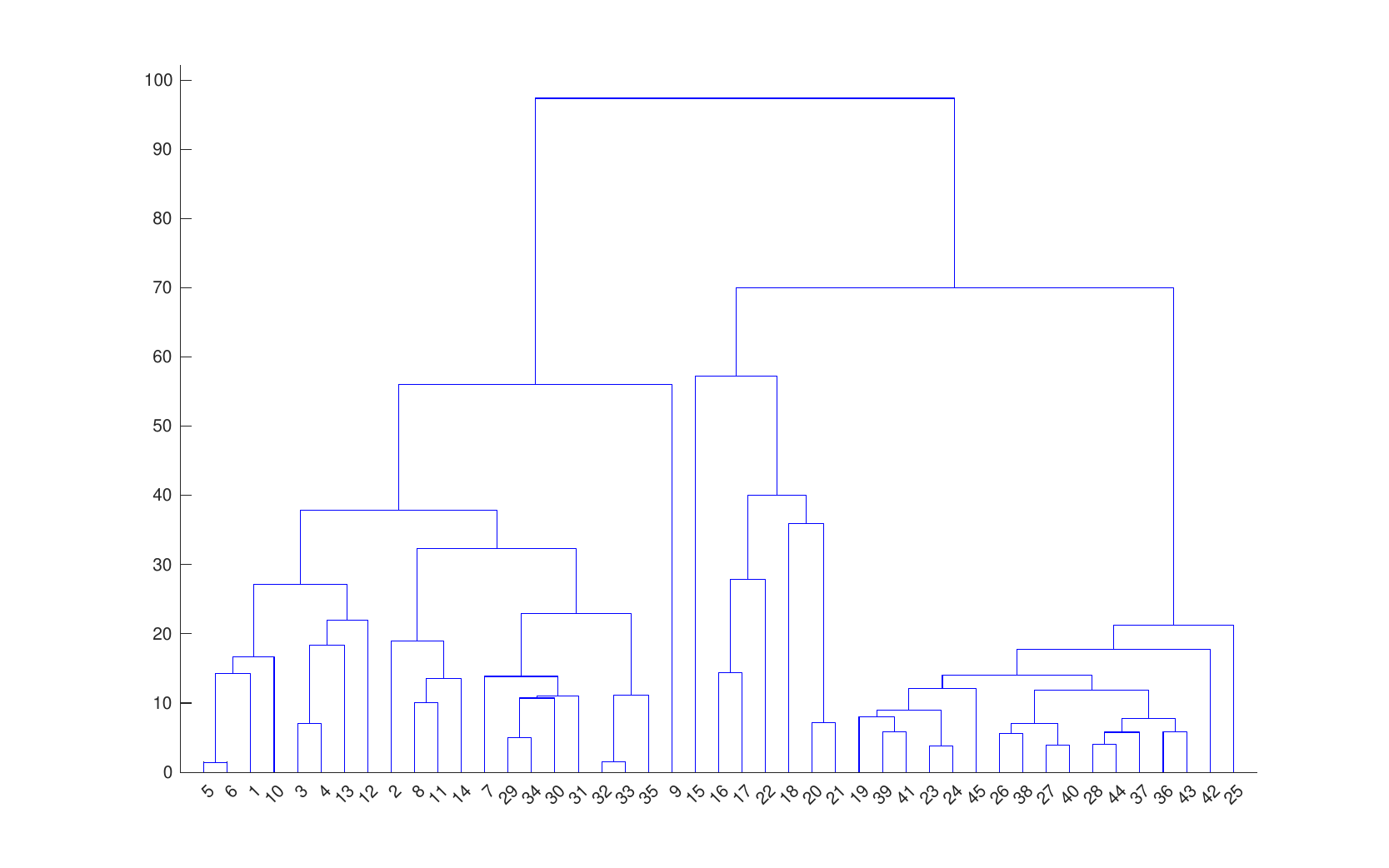}
	\caption{Set 2, Wasserstein distance calculated for the whole curve, weights assigned according to Formula \ref{waga_wspolrzedne_odwrocona}.}
	\label{SEt2_normal_distribution_12}
\end{figure}

\begin{figure}[H]
	\centering
	\includegraphics[width=0.9\linewidth]{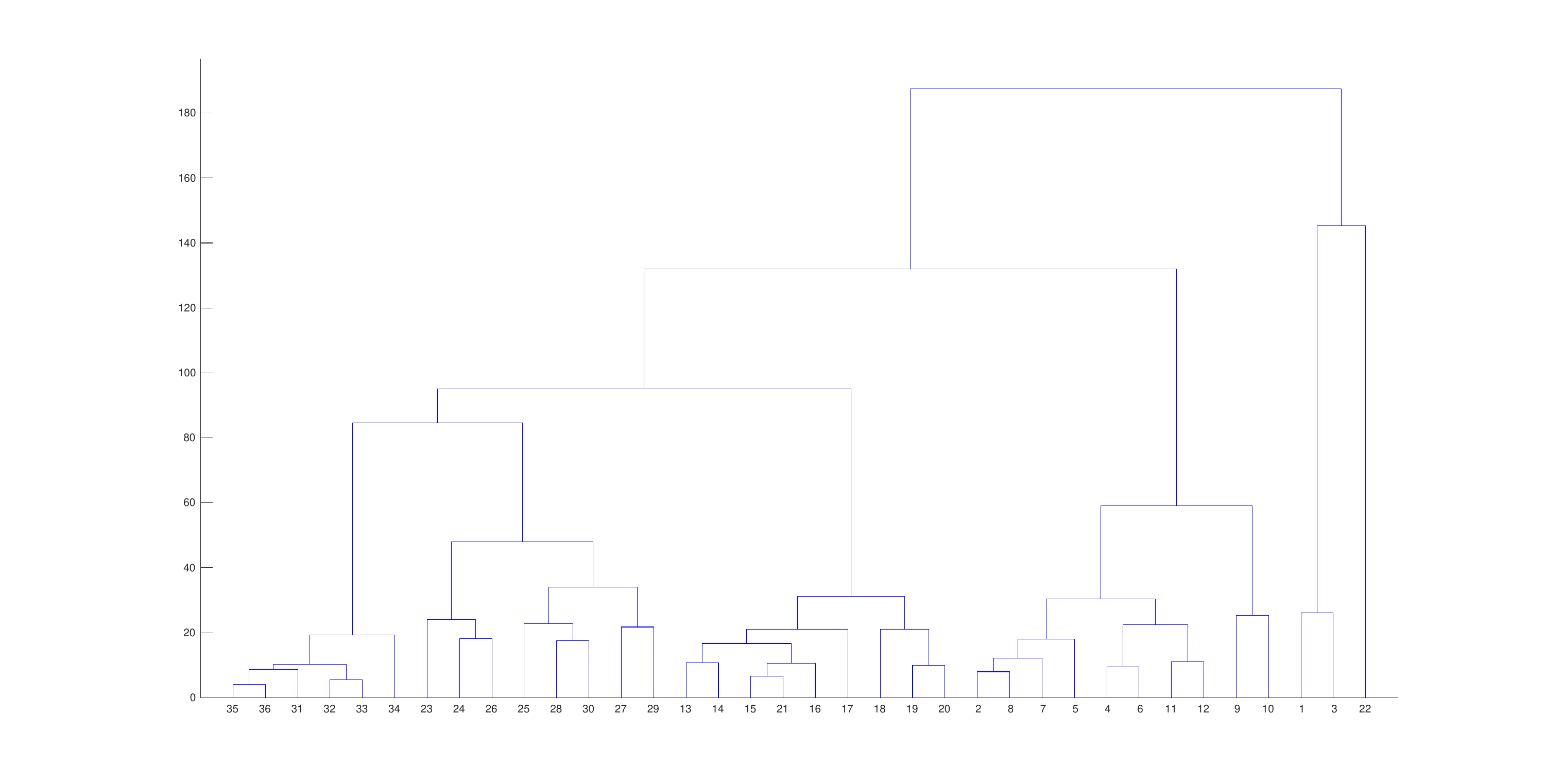}
	\caption{Set 3, Wasserstein distance calculated for the whole curve, weights assigned according to Formula \ref{waga_wspolrzedne_odwrocona}.}
	\label{Set3_normal_distribution_12}
\end{figure}

\begin{figure}[H]
	\centering
	\includegraphics[width=0.9\linewidth]{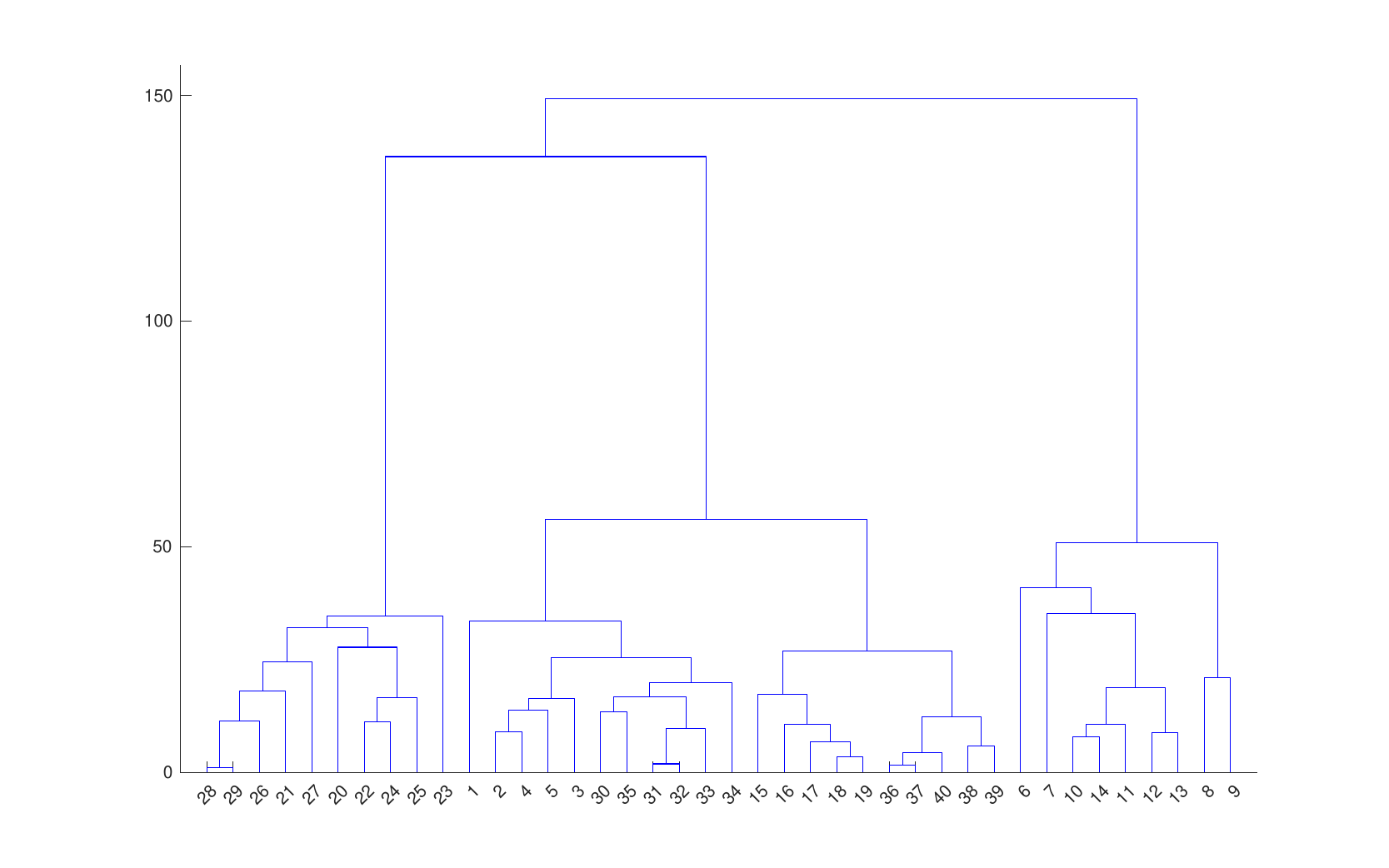}
	\caption{Set 4, Wasserstein distance calculated for the whole curve, weights assigned according to Formula \ref{waga_wspolrzedne_odwrocona}.}
	\label{Set4_normal_distribution_12}
\end{figure}

\begin{figure}[H]
	\centering
	\includegraphics[width=0.9\linewidth]{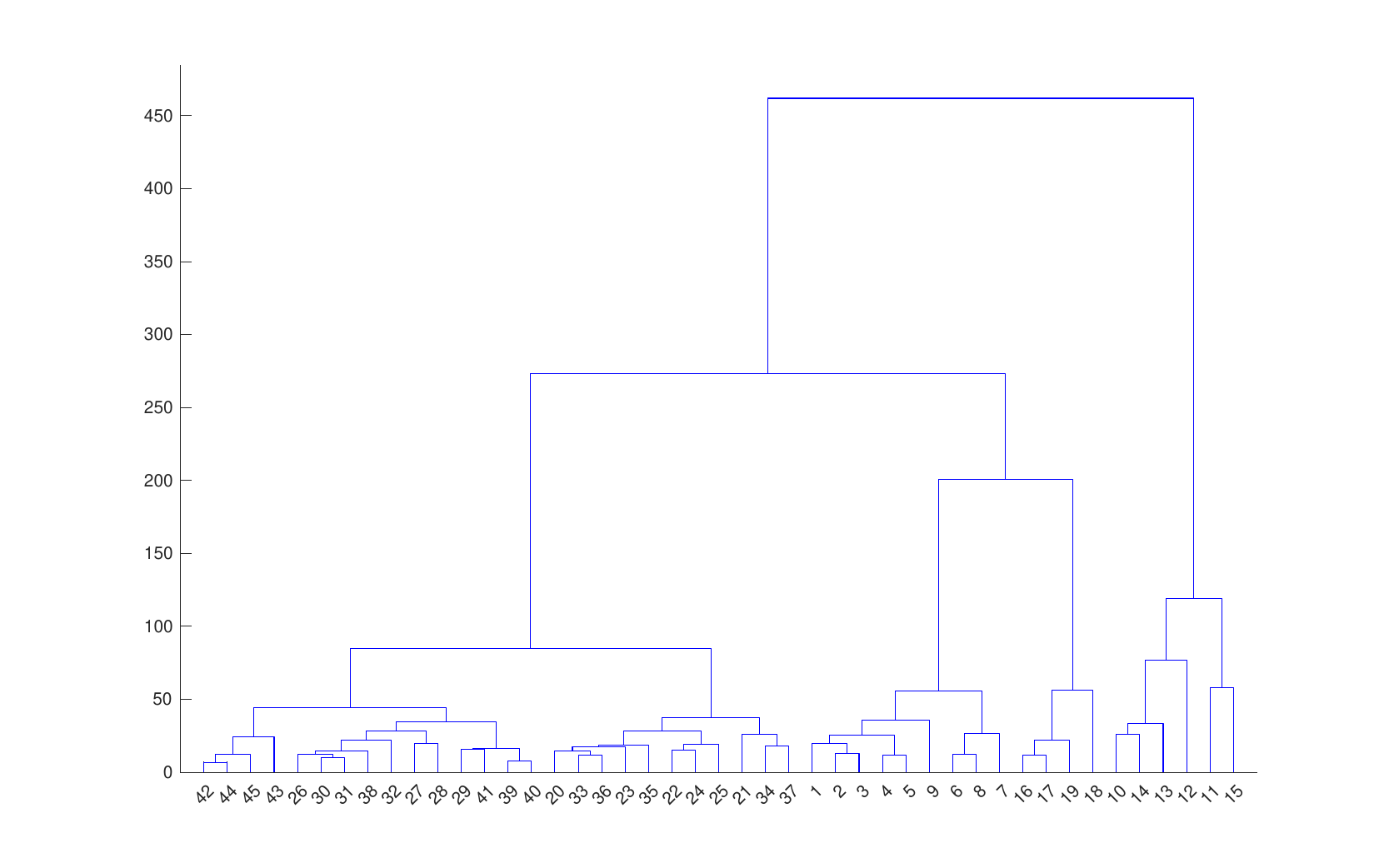}
	\caption{Set 5, Wasserstein distance calculated for the whole curve, weights assigned according to Formula \ref{waga_wspolrzedne_odwrocona}.}
	\label{Set5_normal_distribution_12}
\end{figure}

\begin{figure}[H]
	\centering
	\includegraphics[width=0.9\linewidth]{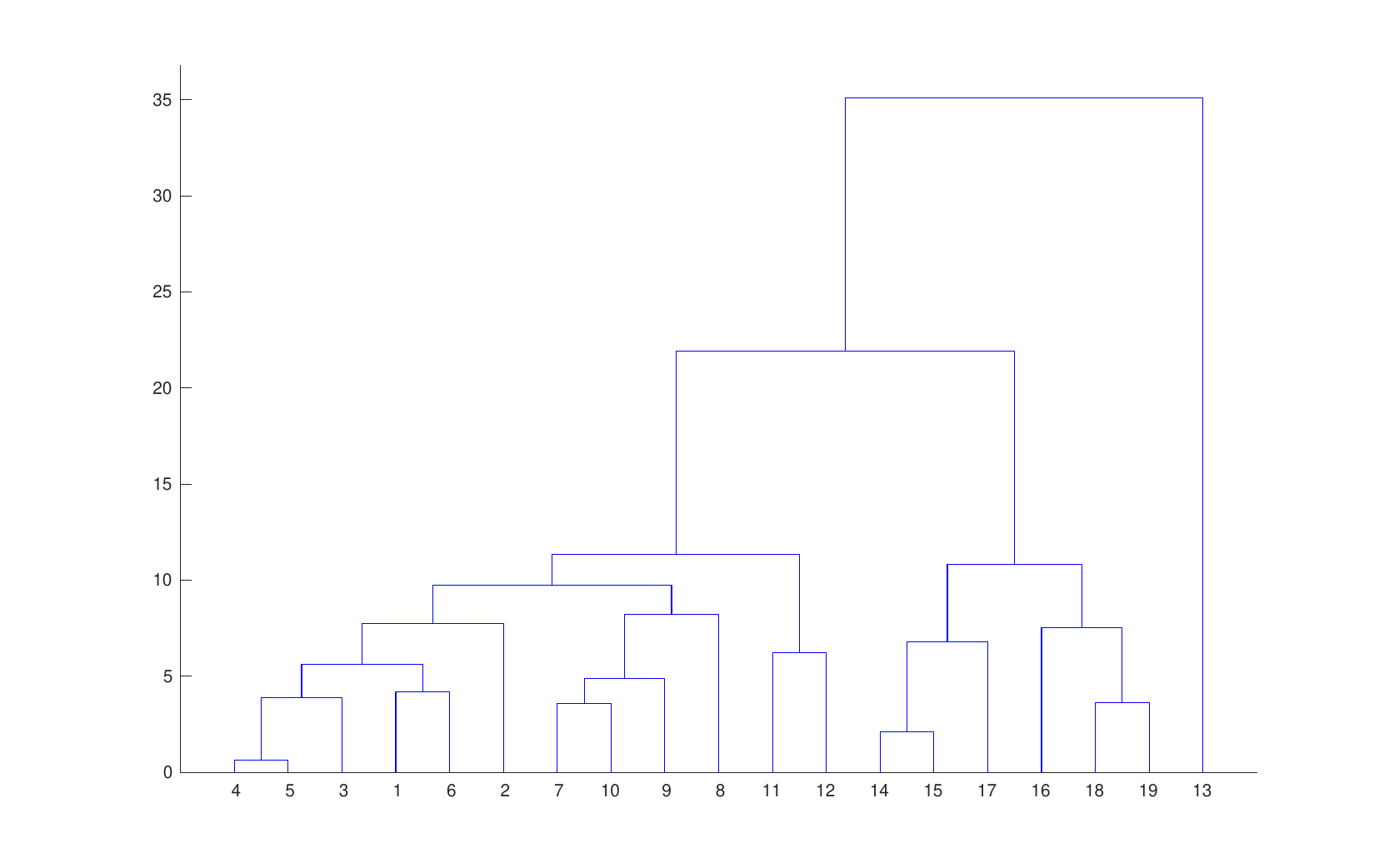}
	\caption{Set 1, Wasserstein distance calculated for the whole curve, weights assigned according to Formula \ref{waga_po_ycale_odwrocona}.}
	\label{Set1_normal_distribution_15}
\end{figure}

\begin{figure}[H]
	\centering
	\includegraphics[width=0.9\linewidth]{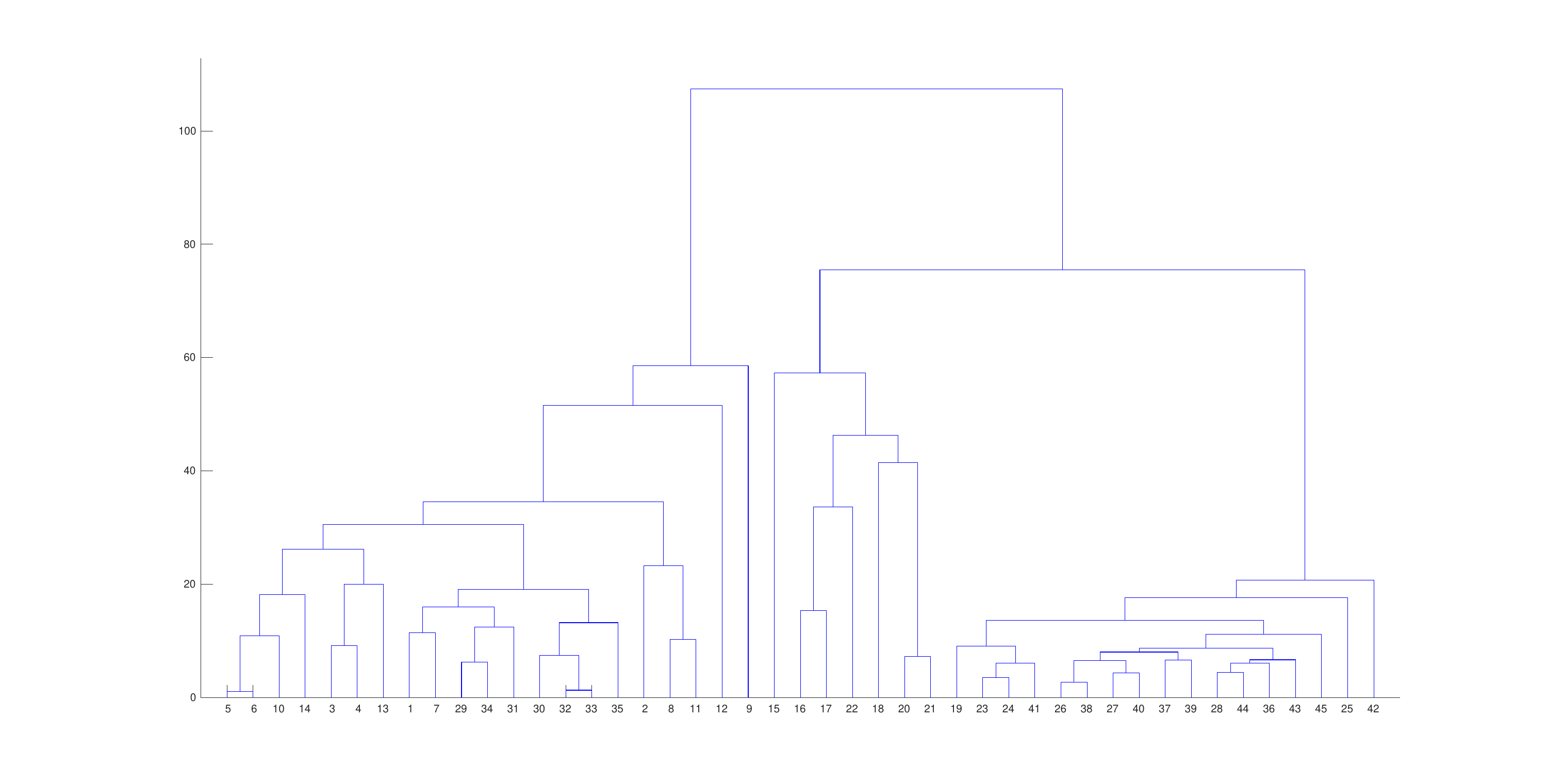}
	\caption{Set 2, Wasserstein distance calculated for the whole curve, weights assigned according to Formula \ref{waga_po_ycale_odwrocona}.}
	\label{Set2_normal_distribution_15}
\end{figure}

\begin{figure}[H]
	\centering
	\includegraphics[width=0.9\linewidth]{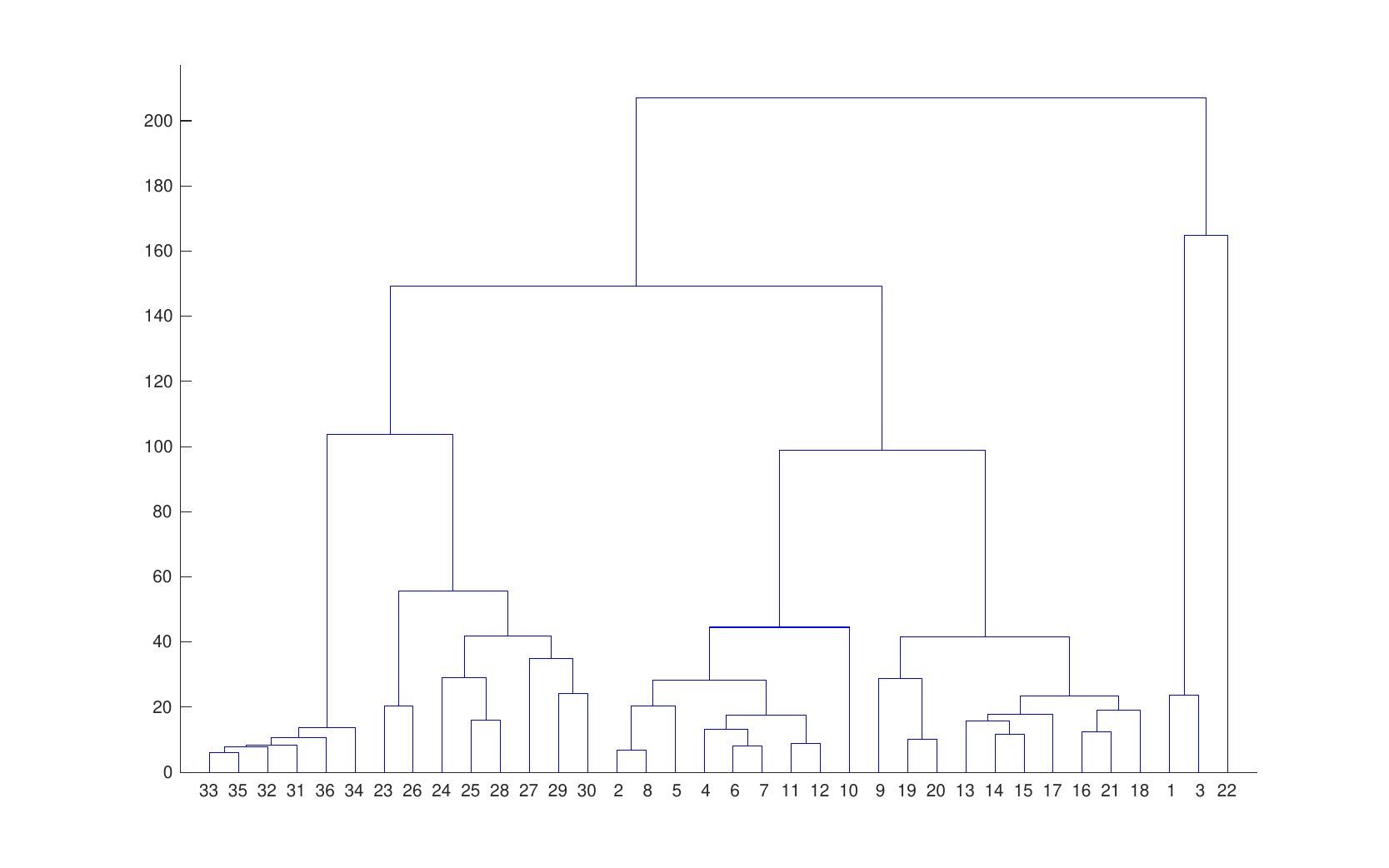}
	\caption{Set 3, Wasserstein distance calculated for the whole curve, weights assigned according to Formula \ref{waga_po_ycale_odwrocona}.}
	\label{Set3_normal_distribution_15}
\end{figure}

\begin{figure}[H]
	\centering
	\includegraphics[width=0.9\linewidth]{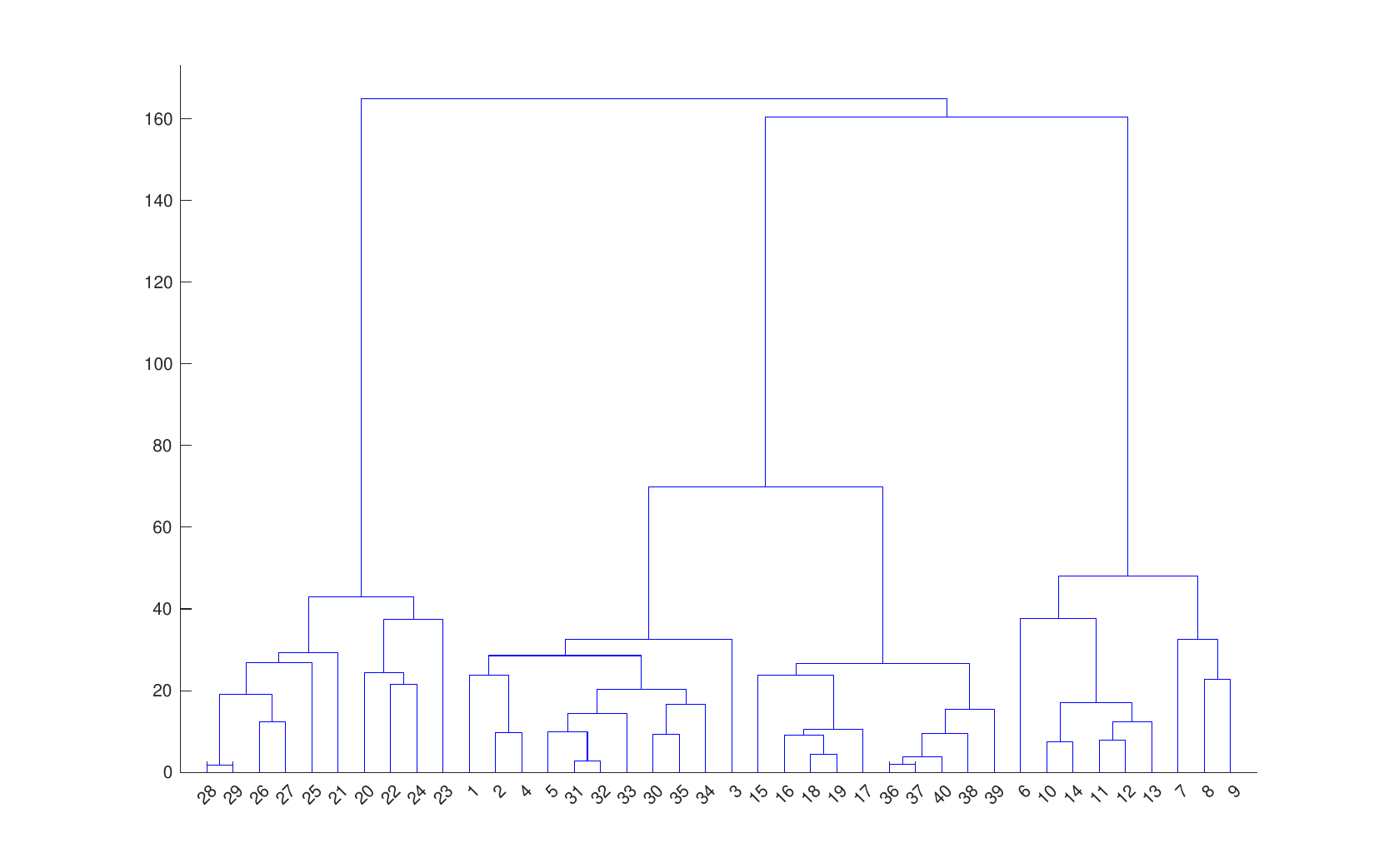}
	\caption{Set 4, Wasserstein distance calculated for the whole curve, weights assigned according to Formula \ref{waga_po_ycale_odwrocona}.}
	\label{Set4_normal_distribution_15}
\end{figure}

\begin{figure}[H]
	\centering
	\includegraphics[width=0.9\linewidth]{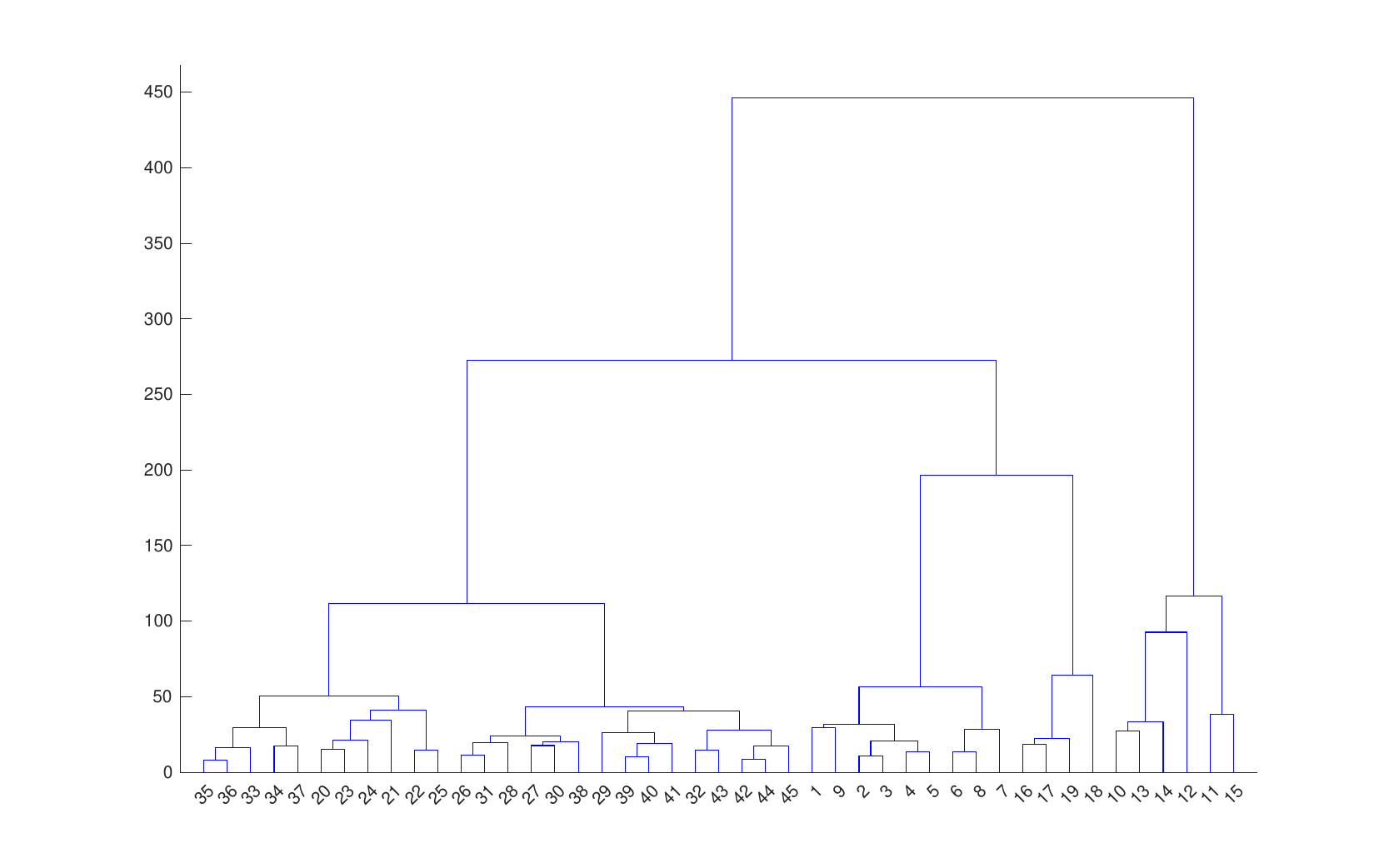}
	\caption{Set 5, Wasserstein distance calculated for the whole curve, weights assigned according to Formula \ref{waga_po_ycale_odwrocona}.}
	\label{Set5_normal_distribution_15}
\end{figure}

 In Experiment 6 and 7 we use Formulas \ref{waga_wspolrzedne_odwrocona} and \ref{waga_po_ycale_odwrocona} in the assignment of weights. Theses allow us to take advantage of the indices and the second component respectively to assign weights, linking the value of the weight to the shape property of the investigated curve. The results are presented in Figures \ref{Set1_normal_distribution_12}- \ref{Set5_normal_distribution_15}.
 
 Results of all Experiments are summarized in Table \ref{table_results}.

\subsection{Experiments: discussion of results}
\label{Section_experiments_disscusion}

The obtained results can be generally described as very good, most of them are above 85\% of correctly classified objects. The differences in the performance of any given method are mostly due to the difference in the character of the set which can render the set more or less well suited for the given method. The exact choice between the presented methods is up to the expert and the objective of the performed grouping. 

Experiment 1 is performed to establish a baseline for the performance of the classic formulation of the Wasserstein distance. 

Experiments 2 and 3 were conducted were conducted in order to test the results when the value of "0" is applied to a section of the investigated curve. The obtained results proved valuable in determining that despite the value "0" is assigned to a given section it is not equivalent to the removal of that section as in Experiment 2. When comparing the results of these two experiments we can conclude that in Experiment 3, the section of the curve with weights of "0" still impacts the result, however to a small degree. Such an approach would be useful to the classification of a set of curves were a known, prescribed section of the curve is of high interest in the study, but the information in the remainder of the curve is not completely relevant.

Experiments 4 and 5  we have employed the binomial distribution to reflect the distribution of interest in points along the curve. The parameters of the employed distribution allow great flexibility in highlighting the area of interest, reflecting the natural perception of fragments of shapes. It does not necessitate to select a speciffic cut-off point.
The right choice of the parameters of the normal distribution can result in a similar result to Experiment 3, where the weights $1/n$ are assigned to only a part of the curve and the remainder in taken as "0". This is similar to the situation in Experiment 4 and 5 where the parameters of the binomial distribution curve are such that, in our case, 1/6-th of the curve co responds to a sharp drop in the values of the weights approaching "0".

Experiments 6, 7, and 8 are based on Formulas \ref{waga_wspolrzedne_odwrocona}, \ref{waga_po_ycale_odwrocona}, and \ref{waga_po_y_dlak}. Experiments 6 and 7 are somewhat analogous to Experiments 4 and 5, but there is no flexibility in modelling the part of interest. The weights are linked to the indices of the points in the case of Experiment 6 or to the values of the second coordinate in the case of Experiment 7. The exact choice of one of those measures is linked to the characteristic of the investigated shape. A range of similar measures, suitable for different applications, are described in Section \ref{Section_Wass_for_different_distributions}.  

Experiment 8 is an extension of Experiments 2 and 3, allowing to classify a prescribed section of the curve simultaneously linking the second component to the weights, further stressing the importance of the given part of the curve. This approach has many conceivable variants, some of them are presented in Section \ref{Section_Wass_for_different_distributions}. 

\section{Conclusions}
\label{Section_conclusions}
The proposed methods were aimed at the application of the Wasserstein distance to the classification of shapes, 2D curves in particular. The Wasserstein distance is proven to be a good, robust metric for shape classification. One of the properties of the metric is the possibility to assign weights to the points of the investigated curves, which is embedded in the formulation of the distance. We take advantage of this formulation, providing a number of distributions that allow the focus of the classification on prescribed section of the curve. The Proposed distributions are tested through a series of experiments.

The encouraging results obtained above prove the validity of the using the Wasserstein distance for shape similarity detection and shape classification of 2D curves. The formulation of the Wasserstein distance makes it particularly useful for focusing the classification to a selected part of the investigated curves.  The proposed approach can be  applied to data coming form different disciplines, eg. signals.

\bibliography{wass}

\section{Figures}
\label{sec:Figures}
\begin{table*}[!t]
 \caption{Percentage of correctly classified elements as evaluated by the expert for Sets 1-5. The number of elements in each set, and the number of correctly classified elements, is given in brackets. The best result for each experiment is bolded.}
\label{table_results}
\renewcommand{\arraystretch}{1.6}
\begin{tabular}{ | m{0.2cm}| m{7em} | m{1.8cm}| m{1.8cm} | m{1.8cm} | m{1.8cm} |  m{1.8cm}|}
\hline
Nr.& Experiment
& set 1 [19]& set 2 [45] & set 3 [36]& set 4 [40]  & set 5 [45]  \\

\hline

1 & Whole curve, according to Formula \ref{WD}, with weights  $1/n$ according to \ref{waga_uniform} & 84.21 \%[16]& 86.67\%[39]& \bf{94.44\%[34]}&  92.5\%[37]  & 86.7\% [39]\\

 	\hline
2 & $1/6$ of the curve, with weights  $1/n$ according to \ref{waga_uniform} 
 & 89.47\%[17]& 93.33\%[42] & \bf{100\%[36]}&  90\%[36] & 91\% [41]\\
	
	\hline
  
3& The top $1/6$ of the curve with with weights  $1/n$ according to \ref{waga_uniform}, for the remainder 5/6 of the curve weights of "0" were assigned	 &  89.47\%[17]& 82.22\%[37] & \bf{100\%[36]} &  90\%[36] & 
95.6 \%[43]\\
	
		\hline
4& Whole curve, weights assigned according to the binomial distribution curve (a) Fig. \ref{fig_normal_distributions}  
	& 84.21 \%[16]& 91.11\%[41]&  \bf{91.67\%[33]}&  90\%[36] & 91\% [41] \\
	
			\hline
5& Whole curve, weights assigned according to the binomial distribution curve (b) Fig. \ref{fig_normal_distributions}  
	& 84.21 \%[16]& \bf{100\%[45]}&  94.44\%[34]&  87,5\%[35] & 86.7\%[39]\\

		\hline
6& Whole curve, according to Formula \ref{waga_wspolrzedne_odwrocona}
	 &  89.47\%[17]& \bf{97.78\%[44]}& 94.44\%[34]& 87.5\%[35] & 86.7\%[39]\\

		\hline
7& Whole curve, according to Formula \ref{waga_po_ycale_odwrocona}
	 & 94.74\%[18]& 91.11\%[41]& 89.67\%[33]& \bf{95\%[38]} & 93.3\%[42]\\

		\hline
8& Whole curve, according to Formula \ref{waga_po_y_dlak}
	 & 84.21\%[16]& 84.4\%[38]& \bf{100\%[36]}& 90\%[36] & 91\%[41]\\

		\hline
\end{tabular}
\end{table*}

\subsection{Data sets}

\begin{figure}[H]
	\centering
	\includegraphics[width=1\linewidth]{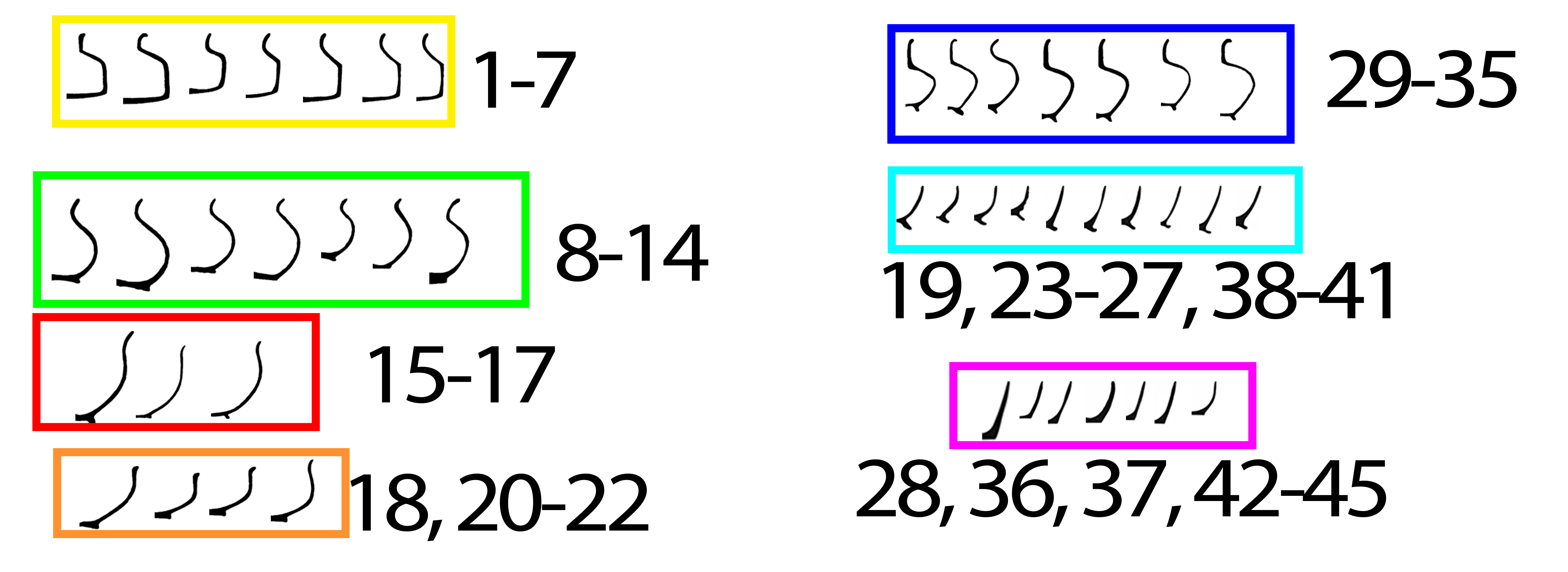}
	\caption{Set 2.}
	\label{Set2}
\end{figure}

\begin{figure}[H]
	\centering
	\includegraphics[width=1\linewidth]{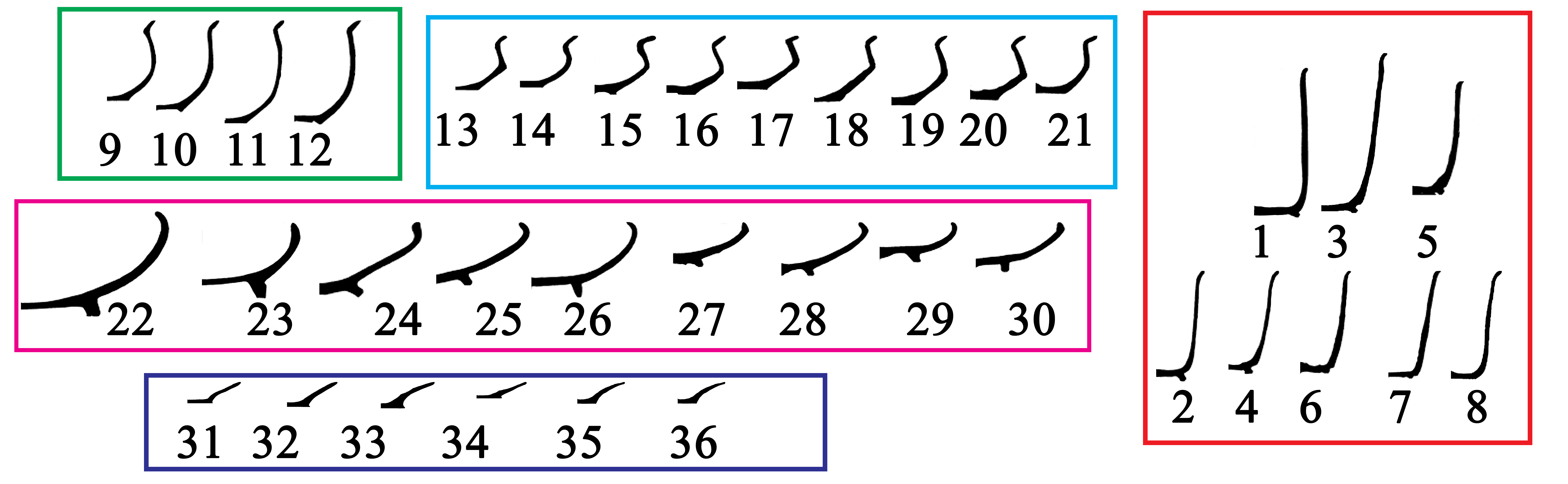}
	\caption{Set 3.}
	\label{Set3}
\end{figure}

\begin{figure}[H]
	\centering
	\includegraphics[width=1\linewidth]{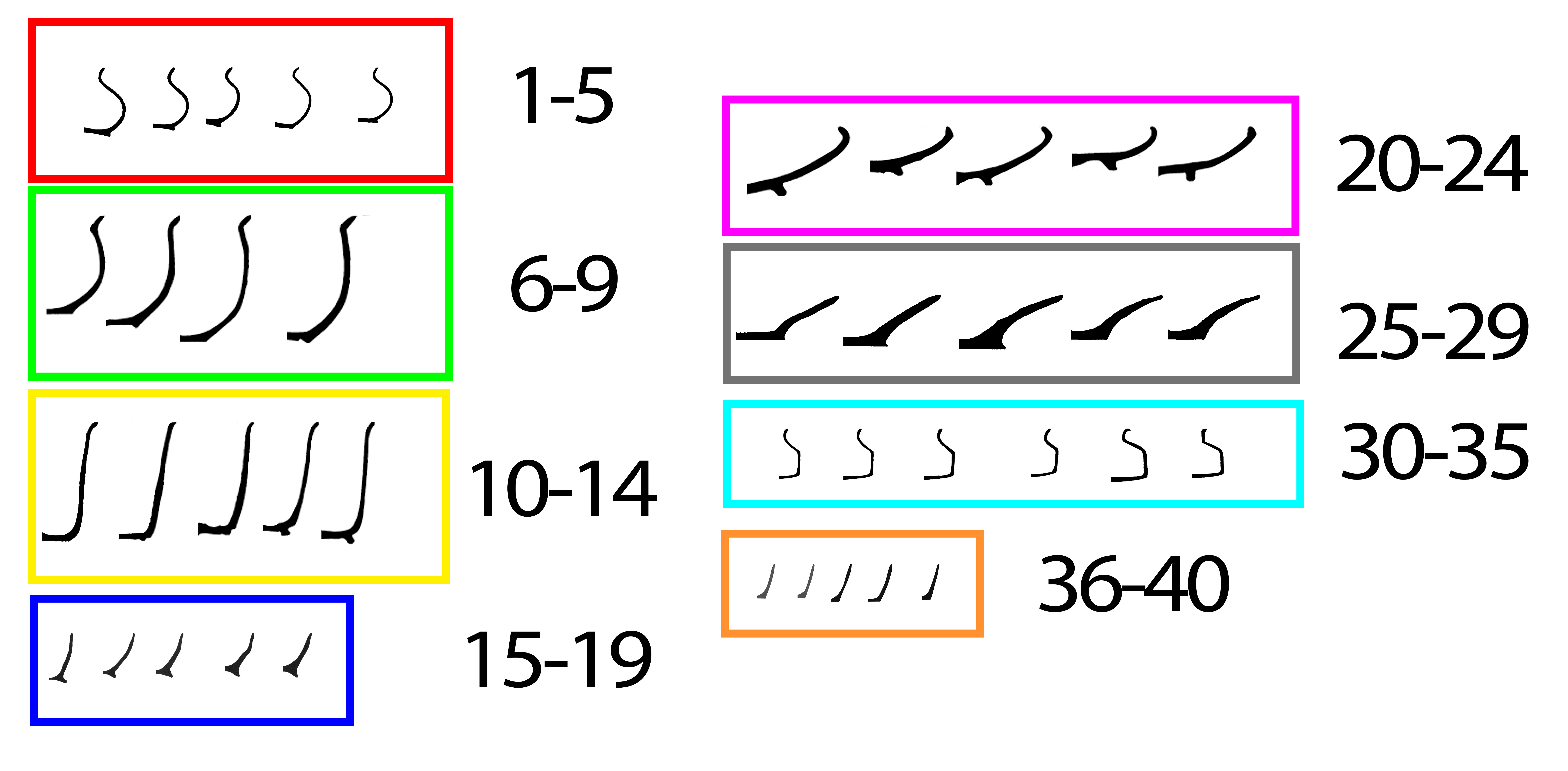}
	\caption{Set 4.}
	\label{Set4}
\end{figure}

\begin{figure}[H]
	\centering
	\includegraphics[width=0.7\linewidth]{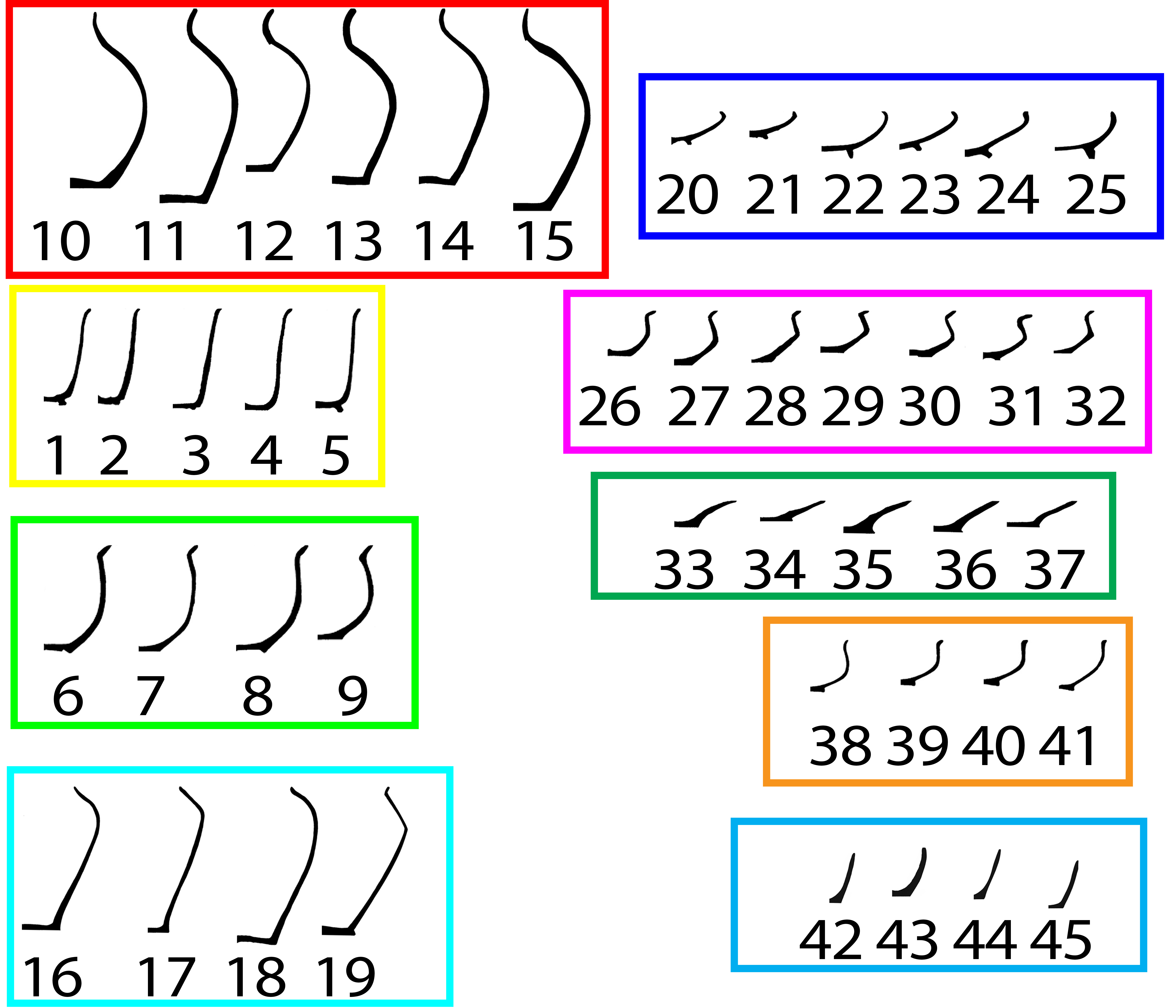}
	\caption{Set 5.}
	\label{Set5}
\end{figure}

\subsection{Experiment 1}
\label{Figures_experiment1}

\subsection{Experiment 2}

\subsection{Experiment 3}

\subsection{Experiment 4 and 5}



\subsection{Experiment 6}

\subsection{Experiment 7}

\subsection{Experiment 8}

\end{document}